\documentclass[10pt,twocolumn,letterpaper]{article}

\usepackage{cvpr}
\usepackage{times}
\usepackage{epsfig}
\usepackage{graphicx}
\usepackage{amsmath}
\usepackage{amssymb}

\usepackage{epstopdf}
\usepackage{threeparttable}
\usepackage{booktabs}
\usepackage{pifont}
\newcommand{\cmark}{\ding{51}}%
\newcommand{\xmark}{\ding{55}}%
\usepackage{makecell}
\usepackage{multirow}
\usepackage{enumerate}

\newcommand{\PreserveBackslash}[1]{\let\temp=\\#1\let\\=\temp}
\newcolumntype{C}[1]{>{\PreserveBackslash\centering}p{#1}}
\newcolumntype{R}[1]{>{\PreserveBackslash\raggedleft}p{#1}}
\newcolumntype{L}[1]{>{\PreserveBackslash\raggedright}p{#1}}

\newcommand{\tabincell}[2]{\begin{tabular}{@{}#1@{}}#2\end{tabular}}

\usepackage[pagebackref=true,breaklinks=true,letterpaper=true,colorlinks,bookmarks=false]{hyperref}

\cvprfinalcopy 

\def\cvprPaperID{2421} 

\begin{document}

\title{LaSOT: A High-quality Benchmark for Large-scale Single Object Tracking}

\author{Heng Fan$^{1}$\thanks{Authors make equal contributions to this work.} \;\;\;  Liting Lin$^{2*}$   \;\; Fan Yang$^{1*}$  \;\;   Peng Chu$^{1*}$ \;\;
	Ge Deng$^{1}$ \;\; Sijia Yu$^{1}$ \;\;  Hexin Bai$^{1}$ \\ Yong Xu$^{2}$ \;\; Chunyuan Liao$^{3}$ \;\; Haibin Ling$^{1}$\thanks{Corresponding author.} \\
	{\normalsize $^{1}$Department of Computer and Information Sciences, Temple University, Philadelphia, USA}\\
	{\normalsize $^{2}$School of Computer Science \& Engineering, South China Univ. of Tech., Guangzhou,} \\{\normalsize Peng Cheng Laboratory, Shenzhen, China} \\
	{\normalsize $^{3}$Meitu HiScene Lab, HiScene Information Technologies, Shanghai, China}\\
	{\tt \normalsize \url{https://cis.temple.edu/lasot/}}
}

\maketitle

\begin{abstract}
	In this paper, we present {\bf LaSOT}, a high-quality benchmark for {\bf La}rge-scale {\bf S}ingle {\bf O}bject {\bf T}racking. LaSOT consists of 1,400 sequences with more than 3.5M frames in total. Each frame in these sequences is carefully and manually annotated with a bounding box, making LaSOT the largest, to the best of our knowledge, densely annotated tracking benchmark. The average video length of LaSOT is more than 2,500 frames, and each sequence comprises various challenges deriving from the wild where target objects may disappear and re-appear again in the view. By releasing LaSOT, we expect to provide the community with a large-scale dedicated benchmark with high quality for both the training of deep trackers and the veritable evaluation of tracking algorithms. Moreover, considering the close connections of visual appearance and natural language, we enrich LaSOT by providing additional language specification, aiming at encouraging the exploration of natural linguistic feature for tracking. A thorough experimental evaluation of 35 tracking algorithms on LaSOT is presented with detailed analysis, and the results demonstrate that there is still a big room for improvements.
\end{abstract}

\section{Introduction}
\label{intro}

\begin{figure}[!t]
	\centering
	\includegraphics[width=1\linewidth]{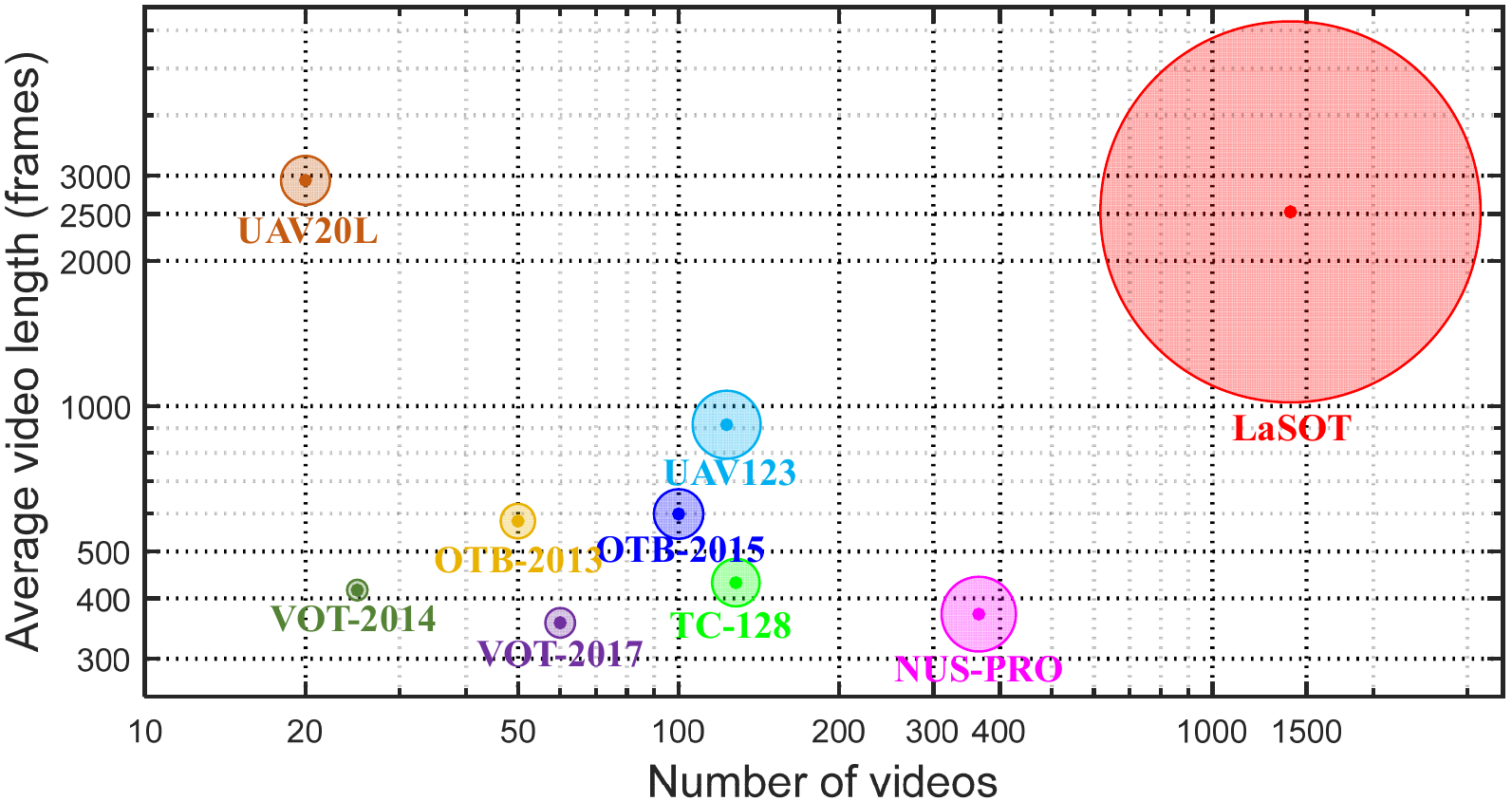}\\
	\caption{Summaries of existing tracking benchmarks with high-quality dense (per frame) annotations, including OTB-2013~\cite{wu2013online}, OTB-2015~\cite{wu2015object}, TC-128~\cite{liang2015encoding}, NUS-PRO~\cite{li2016nus}, UAV123~\cite{mueller2016benchmark}, UAV20L~\cite{mueller2016benchmark}, VOT-2014~\cite{kristan2014visual}, VOT-2017~\cite{kristan2017visual} and LaSOT. The circle diameter is in proportion to the number of frames of a benchmark. The proposed LaSOT is {\it larger} than all other benchmarks, and focused on {\it long-term} tracking. Best viewed in color.}
	\label{fig:benchmark-scale}
\end{figure}

Visual tracking, aiming to locate an arbitrary target in a video with an initial bounding box in the first frame, has been one of the most important problems in computer vision with many applications such as video surveillance, robotics, human-computer interaction and so forth~\cite{li2013survey,smeulders2014visual,yilmaz2006object}. With considerable progresses in the tracking community, numerous algorithms have been proposed. In this process, tracking benchmarks have played a vital role in objectively evaluating and comparing different trackers. Nevertheless, further development and assessment of tracking algorithms are restricted by existing benchmarks with several issues:

\renewcommand\arraystretch{1.05}
\begin{table*}[!t]\scriptsize
	\centering
	\caption{Comparison of LaSOT with the most popular dense benchmarks in the literatures.}
	\begin{tabular}{r||C{0.5cm}C{0.6cm}C{0.6cm}cccccccccc}
		\hline
		{Benchmark} & \tabincell{c}{ Videos} & \tabincell{c}{ Min \\ frames} & \tabincell{c}{Mean \\ frames} & \tabincell{c}{Median \\ frames} & \tabincell{c}{Max \\ frames} & \tabincell{c}{Total \\ frames}  & \tabincell{c}{Total \\ duration} &\tabincell{c}{frame \\ rate} & \tabincell{c}{Absent \\ labels} & \tabincell{c}{Object \\ classes}  & \tabincell{c}{Class \\ balance} & \tabincell{c}{Num. of \\ attributes} & \tabincell{c}{Lingual \\ feature} \\
		\hline \hline
		{\bf OTB-2013}~\cite{wu2013online} & 51    & 71    & 578   & 392   & 3,872 & 29$\mathbf{K}$   & 16.4 $\mathbf{min}$ & 30 fps    &  \xmark     & 10     &   \xmark    & 11    & \xmark \\
		{\bf OTB-2015}~\cite{wu2015object} & 100   & 71    & 590   & 393   & 3,872 & 59$\mathbf{K}$   & 32.8 $\mathbf{min}$ & 30 fps   &  \xmark      & 16    &  \xmark     & 11    & \xmark \\
		{\bf TC-128}~\cite{liang2015encoding} & 128   & 71    & 429   & 365   & 3,872 & 55$\mathbf{K}$   & 30.7 $\mathbf{min}$ & 30 fps   &   \xmark     & 27    &   \xmark    & 11    & \xmark \\
		{\bf VOT-2014}~\cite{kristan2014visual} & 25    & 164   & 409   & 307   & 1,210 & 10$\mathbf{K}$   & 5.7 $\mathbf{min}$ & 30 fps   &   \xmark     & 11     &   \xmark    & n/a     & \xmark \\
		{\bf VOT-2017}~\cite{kristan2017visual} & 60    & 41    & 356   & 293   & 1,500 & 21$\mathbf{K}$   & 11.9 $\mathbf{min}$ & 30 fps   &  \xmark      & 24    &  \xmark     & n/a     & \xmark \\
		{\bf NUS-PRO}~\cite{li2016nus} & 365   & 146   & 371   & 300   & 5,040 & 135$\mathbf{K}$  & 75.2 $\mathbf{min}$ & 30 fps   &   \xmark     & 8     &  \xmark     & n/a     & \xmark \\
		{\bf UAV123}~\cite{mueller2016benchmark} & 123   & 109   & 915   & 882   & 3,085 & 113$\mathbf{K}$  & 62.5 $\mathbf{min}$ & 30 fps   &   \xmark     & 9     &  \xmark     & 12    & \xmark \\
		{\bf UAV20L}~\cite{mueller2016benchmark} & 20    & 1,717 & 2,934 & 2,626 & 5,527 & 59$\mathbf{K}$   & 32.6 $\mathbf{min}$ & 30 fps   &   \xmark     & 5     &  \xmark     & 12    & \xmark \\
		{\bf NfS}~\cite{galoogahi2017need}   & 100   & 169   & 3,830 & 2,448 & 20,665 & 383$\mathbf{K}$  & 26.6 $\mathbf{min}$ & 240 fps  &  \xmark      & 17     &  \xmark     & 9     & \xmark \\
		{\bf GOT-10k}~\cite{huang2018got}   & 10,000   & -   & - & - & - & 1.5$\mathbf{M}$  & - & 10 fps  &  \cmark      & 563     &  \xmark     & 6     & \xmark \\
		\hline \hline
		{\bf LaSOT} & 1,400 & 1,000 & 2,506 & 2,053 & 11,397 & 3.52$\mathbf{M}$ & 32.5 $\mathbf{hours}$ & 30 fps   &  \cmark      & 70    &   \cmark    & 14    & \cmark \\
		\hline
	\end{tabular}%
	\vspace{-2mm}	\label{tab:DMA_comparison}%
\end{table*}%


\vspace{.1em}\noindent{\bf{Small-scale.}} Deep representations have been popularly applied to modern object tracking algorithms, and demonstrated state-of-the-art performances. However, it is difficult to train a deep tracker using \emph{tracking-specific} videos due to the scarcity of {\em large-scale} tracking datasets. As shown in Fig.~\ref{fig:benchmark-scale}, existing datasets seldom have more than 400 sequences. As a result, researchers are restricted to leverage either the pre-trained models (\eg,~\cite{simonyan2015very} and~\cite{he2016deep}) from image classification for deep feature extraction or the sequences from video object detection (\eg,~\cite{russakovsky2015imagenet} and~\cite{real2017youtube}) for deep feature learning, which may result in suboptimal tracking performance because of the intrinsic differences among different tasks~\cite{yosinski2014transferable}. Moreover, large scale benchmarks are desired for more reliable evaluation results.

\vspace{.1em}\noindent{\bf{Lack of high-quality dense annotations.}} For tracking, dense (\ie, per frame) annotations with high precision are of importance for several reasons. (\romannumeral1) They ensure more accurate and reliable evaluations; (\romannumeral2) they offer desired training samples for the training of tracking algorithms; and (\romannumeral3) they provide rich temporal contexts among consecutive frames that are important for tracking tasks. It is worth noting that there are recently proposed benchmarks toward large-scale and long-term tracking, such as \cite{muller2018trackingnet} and~\cite{valmadre2018long}, their annotations are however either semi-automatic (\eg, generated by a tracking algorithm) or sparse (\eg, labeled every 30 frames), limiting their usabilities.

\vspace{.1em}\noindent{\bf{Short-term tracking.}} A desired tracker is expected to be capable of locating the target in a relative long period, in which the target may disappear and re-enter the view. However, most existing benchmarks have been focused on {\em short-term} tracking where the average sequence length is less than 600 frames (\ie, 20 seconds for 30 fps, see again Fig.~\ref{fig:benchmark-scale}) and the target almost always appears in the video frame. The evaluations on such {\em short-term} benchmarks may not reflect the real performance of a tracker in real-world applications, and thus restrain the deployment in practice.

\vspace{.1em}\noindent{\bf{Category bias.}} A robust tracking system should exhibit stable performance insensitive to the category the target belongs to, which signifies that the {\em category bias} (or {\em class imbalance}) should be inhibited in both training and evaluating tracking algorithms. However, existing benchmarks usually comprise only a few categories (see Tab.~\ref{tab:DMA_comparison}) with unbalanced numbers of videos.


In the literature, many datasets have been proposed to deal with the issues above: \eg,~\cite{mueller2016benchmark,valmadre2018long} for long-term tracking, \cite{muller2018trackingnet} for large-scale, \cite{wu2013online,liang2015encoding,kristan2016novel} for precise dense annotations. Nevertheless, none of them addresses all the issues, which motivates the proposal of LaSOT.

\subsection{Contribution}

With the above motivations, we provide the community a novel benchmark for {\bf La}rge-scale {\bf S}ingle {\bf O}bject {\bf T}racking (LaSOT) with multi-fold contributions:

\vspace{-.95em}
\begin{enumerate}[1)]
	\setlength{\itemsep}{0pt}
	\setlength{\parsep}{1pt}
	\setlength{\parskip}{1pt}
	
	\item LaSOT consists of 1,400 videos with average 2,512 frames per sequence. Each frame is carefully inspected and manually labeled, and the result visually double-checked and corrected when needed. This way, we generate around 3.52 million high-quality bounding box annotations. Moreover, LaSOT contains 70 categories with each consisting of twenty sequences. To our knowledge, LaSOT is the largest benchmark with high-quality manual dense annotations for object tracking to date. By releasing LaSOT, we aim to offer a dedicated platform for the development and assessment of tracking algorithms.
	
	\item Different from existing datasets, LaSOT provides both visual bounding box annotations and rich natural language specification, which has recently been proven to be beneficial for various vision tasks (\eg, \cite{hu2016natural,li2017person}) including visual tracking~\cite{li2017tracking}. By doing so, we aim to encourage and facilitate explorations of integrating visual and lingual features for robust tracking performance.
	
	\item To assess existing trackers and provide extensive baselines for future comparisons on LaSOT, we evaluate 35 representative trackers under different protocols, and analyze their performances using different metrics.
	
\end{enumerate}

\section{Related Work}
\label{related_work}

With considerable progresses in the tracking community, many trackers and benchmarks have been proposed in recent decades. In this section, we mainly focus on the tracking benchmarks that are relevant to our work, and refer the readers to surveys~\cite{li2013survey,smeulders2014visual,yilmaz2006object,li2018deep} for tracking algorithms.

For a systematic review, we intentionally classify tracking benchmarks into two types: one with dense manual annotations (referred to as {\em dense benchmark} for short) and the other one with sparse and/or (semi-)automatic annotations. In the following, we review each of these two categories.

\subsection{Dense Benchmarks}

Dense tracking benchmark provides dense bounding box annotations for each video sequence. To ensure high quality, the bounding boxes are usually manually labeled with careful inspection. For the visual tracking task, these highly precise annotations are desired for both training and assessing trackers. Currently, the popular dense benchmarks contain OTB~\cite{wu2013online,wu2015object}, TC-128~\cite{liang2015encoding}, VOT~\cite{kristan2016novel}, NUS-PRO~\cite{li2016nus}, UAV~\cite{mueller2016benchmark},  NfS~\cite{galoogahi2017need} and GOT-10k~\cite{huang2018got}.

\vspace{0.1em}
\noindent {\bf OTB.} OTB-2013~\cite{wu2013online} firstly contributes a testing dataset by collecting 51 videos with manually annotated bounding box in each frame. The sequences are labeled with 11 attributes for further analysis of tracking performance. Later, OTB-2013 is extended to the larger OTB-2015~\cite{wu2015object} by introducing extra 50 sequences.

\vspace{0.1em}
\noindent {\bf TC-128.} TC-128~\cite{liang2015encoding} comprises 128 videos that are specifically designated to evaluate color-enhanced trackers. The videos are labeled with 11 similar attributes as in OTB~\cite{wu2013online}.

\vspace{0.1em}
\noindent {\bf VOT.} VOT~\cite{kristan2016novel} introduces a series of tracking competitions with up to 60 sequences in each of them, aiming to evaluate the performance of a tracker in a relative short duration. Each frame in the VOT datasets is annotated with a rotated bounding box with several attributes.

\vspace{0.1em}
\noindent {\bf NUS-PRO.} NUS-PRO~\cite{li2016nus} contains 365 sequences with a focus on human and rigid object tracking.  Each sequence in NUS-PRO is annotated with both target location and occlusion level for evaluation.

\vspace{0.1em}
\noindent {\bf UAV.}  UAV123 and UAV20L~\cite{mueller2016benchmark} are utilized for unmanned aerial vehicle (UAV) tracking, comprising 123 short and 20 long sequences, respectively. Both UAV123 and UAV20L are labeled with 12 attributes.

\vspace{0.1em}
\noindent {\bf NfS.} NfS~\cite{galoogahi2017need} provides 100 sequences with a high framerate of 240 fps, aiming to analyze the effects of appearance variations on tracking performance.

\vspace{0.1em}
\noindent {\bf GOT-10k.} GOT-10k~\cite{huang2018got} consists of 10,000 videos, aiming to provide rich motion trajectories for developing and evaluating trackers.

LaSOT belongs to the category of dense tracking dataset. Compared to others, LaSOT is the {\em largest} with 3.52 million frames and an average sequence length of 2,512 frames. In addition, LaSOT provides extra lingual description for each video while others do not. Tab.~\ref{tab:DMA_comparison} provides a detailed comparison of LaSOT with existing dense benchmarks.

\subsection{Other Benchmarks}

In addition to the dense tracking benchmarks, there exist other benchmarks which may not provide high-quality annotations for each frame. Instead, these benchmarks are either annotated sparsely (\eg, every 30 frames) or labeled (semi-)automatically by tracking algorithms. Representatives of this type of benchmarks include ALOV~\cite{smeulders2014visual}, TrackingNet~\cite{muller2018trackingnet} and OxUvA~\cite{valmadre2018long}.
{\bf ALOV}~\cite{smeulders2014visual} consists of 314 sequences labeled in 14 attributes. Instead of densely annotating each frame, ALOV provides annotations every 5 frames. {\bf TrackingNet}~\cite{muller2018trackingnet} is a subset of the video object detection benchmark YT-BB~\cite{real2017youtube} by selecting 30K videos, each of which is annotated by a tracker. Though the tracker used for annotation is proven to be reliable in a short period (\ie, 1 second) on OTB~2015~\cite{wu2015object}, it is difficult to guarantee the same performance on a harder benchmark. Besides, the average sequence length of TrackingNet does not exceed 500 frames, which may not demonstrate the performance of a tracker in long-term scenarios. {\bf OxUvA}~\cite{valmadre2018long} also comes from YT-BB~\cite{real2017youtube}. Unlike TrackingNet, OxUvA is focused on long-term tracking. It contains 366 videos with an average length of around 4,200 frames. However, a problem with OxUvA is that it does not provide dense annotations in consecutive frames. Each video in OxUvA is annotated every 30 frames, ignoring rich temporal context between consecutive frames when developing a tracking algorithm.

Despite reduction of annotation cost, the evaluations on these benchmarks may not faithfully reflect the true performances of tracking algorithms. Moreover, it may cause problems for some trackers that need to learn temporal models from annotations, since the temporal context in these benchmarks may be either {\em lost} due to sparse annotation or {\em inaccurate} due to potentially unreliable annotation. By contrast, LaSOT provides a large set of sequences with high-quality dense bounding box annotations, which makes it more suitable for developing deep trackers as well as evaluating long-term tracking in practical application.

\section{The Proposed LaSOT Benchmark}
\label{lasot}

\subsection{Design Principle}

LaSOT aims to offer the community a dedicated dataset for training and assessing trackers. To such purpose, we follow five principles in constructing LaSOT, including {\em large-scale}, {\em high-quality dense annotations}, {\em long-term tracking}, {\em category balance} and {\em comprehensive labeling}.

\vspace{-.95em}
\begin{enumerate}[1)]
	\setlength{\itemsep}{0pt}
	\setlength{\parsep}{1pt}
	\setlength{\parskip}{1pt}
	
	\item  {\bf Large-scale.} One of the key motivations of LaSOT is to provide a dataset for training data-hungry deep trackers, which require a large set of annotated sequences. Accordingly, we expect such a dataset to contain at least a thousand videos with at least a million frames.
	
	\item {\bf High-quality dense annotations.} As mentioned before, a tracking dataset is desired to have high-quality dense bounding box annotations, which are crucial for training robust trackers as well as for faithful  evaluation. For this purpose, each sequence in LaSOT is manually annotated with additional careful inspection and fine-tuning.
	
	\item {\bf Long-term tracking.} In comparison with short-term tracking, long-term tracking can reflect more practical performance of a tracker in the wild. We ensure that each sequence comprises {\em at least} 1,000 frames, and the average sequence length in LaSOT is around 2,500 frames.
	
	\item {\bf Category balance.} A robust tracker is expected to perform consistently regardless of the category the target object belongs to. For this purpose, in LaSOT we include a diverse set of objects from 70 classes and each class contains equal number of videos.
	
	\item {\bf Comprehensive labeling.} As a complex task, tracking has recently seen improvements from natural language specification. To stimulate more explorations, a principle of LaSOT is to provide comprehensive labeling for videos, including both visual and lingual annotations.
	
\end{enumerate}
\vspace{-0.5em}

\subsection{Data Collection}

Our benchmark covers a wide range of object categories in diverse contexts. Specifically, LaSOT consists of 70 object categories. Most of the categories are selected from the 1,000 classes from ImageNet~\cite{deng2009imagenet}, with a few exceptions (\eg {\em, drone}) that are carefully chosen for popular tracking applications. Different from existing dense benchmarks that have less than 30 categories and typically are unevenly distributed, LaSOT provides the same number of sequences for each category to alleviate potential category bias. Details of the dataset can be found in the {\bf supplementary material}.

After determining the 70 object categories in LaSOT, we have searched for the videos of each class from YouTube. Initially, we collect over 5,000 videos. With a joint consideration of the quality of videos for tracking and the design principles of LaSOT, we pick out 1,400 videos. However, these 1,400 sequences are not immediately available for the tracking task because of a large amount of irrelevant contents. For example, for the video of {\em person} category (\eg, a sporter), it often contains some introduction content of each sporter in the beginning, which is undesirable for tracking. Therefore, we carefully filter out these unrelated contents in each video and retain an usable clip for tracking. In addition, each category in LaSOT consists of 20 targets, reflecting the category balance and varieties of natural scenes.

Eventually, we have compiled a large-scale dataset by gathering 1,400 sequences with 3.52 million frames from YouTube under Creative Commons licence. The average video length of LaSOT is 2,512 frames (\ie, 84 seconds for 30 fps). The shortest video contains 1,000 frames (\ie, 33 seconds), while the longest one consists of 11,397 frames (\ie, 378 seconds).

\subsection{Annotation}

\begin{figure}[!tb]
	\centering
	\includegraphics[width=2cm]{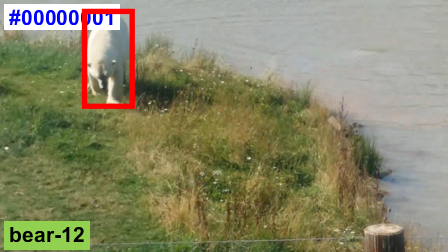}
	\includegraphics[width=2cm]{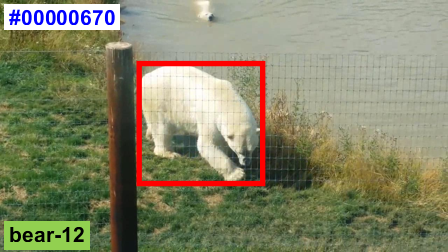}
	\includegraphics[width=2cm]{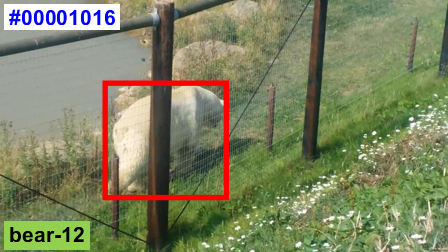} \includegraphics[width=2cm]{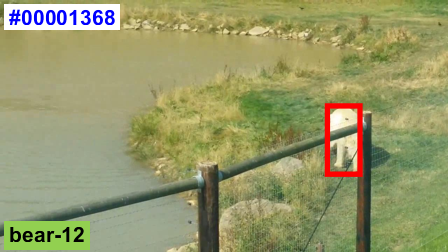} \\
	\scriptsize{{\em Bear-12}: ``white bear walking  on grass around  the river bank''} \\
	\includegraphics[width=2cm]{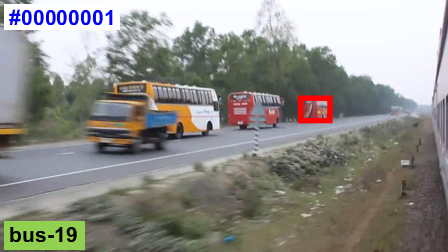}
	\includegraphics[width=2cm]{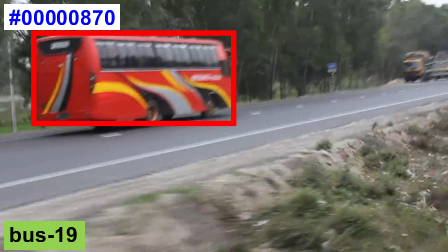}
	\includegraphics[width=2cm]{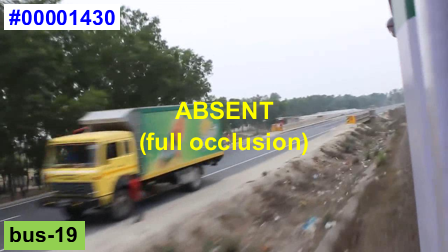} \includegraphics[width=2cm]{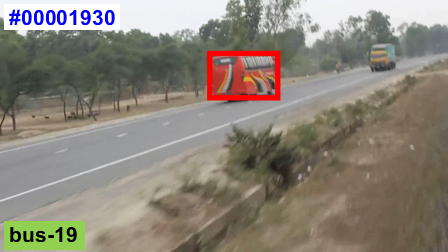} \\
	\scriptsize{{\em Bus-19}: ``red bus running on the highway''} \\
	\includegraphics[width=2cm]{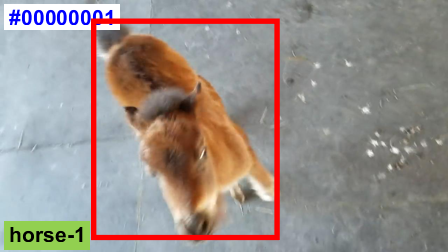}
	\includegraphics[width=2cm]{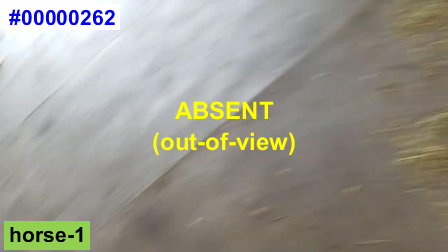}
	\includegraphics[width=2cm]{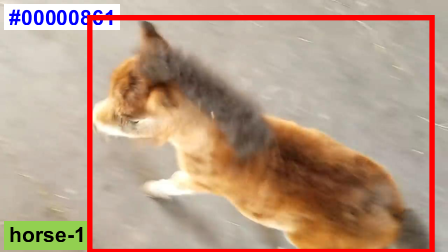} \includegraphics[width=2cm]{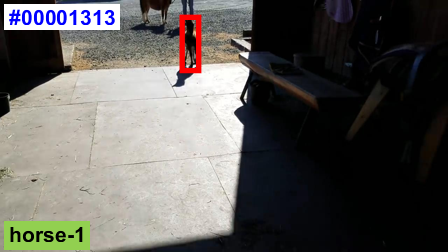} \\
	\scriptsize{{\em Horse-1}: ``brown horse running on the ground''} \\
	\includegraphics[width=2cm]{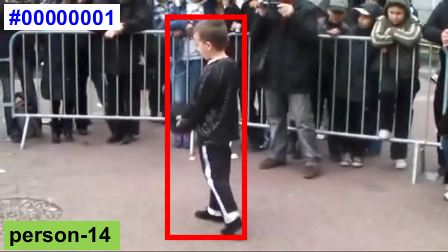}
	\includegraphics[width=2cm]{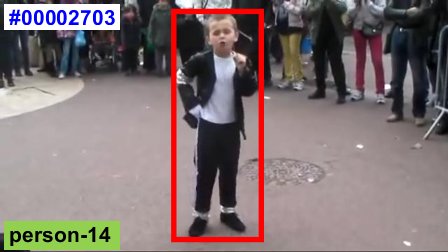}
	\includegraphics[width=2cm]{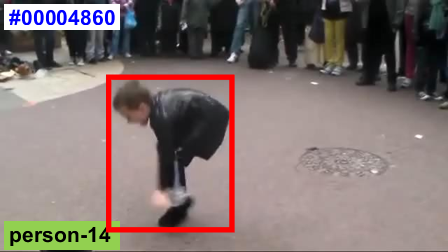} \includegraphics[width=2cm]{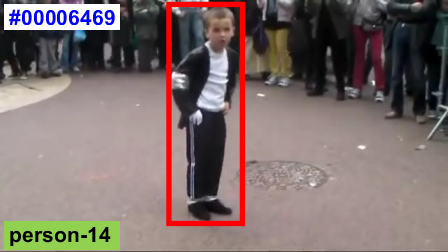} \\
	\scriptsize{{\em Person-14}: ``boy in black suit dancing in front of people''} \\
	\includegraphics[width=2cm]{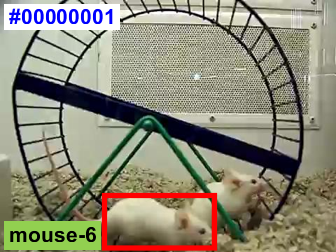}
	\includegraphics[width=2cm]{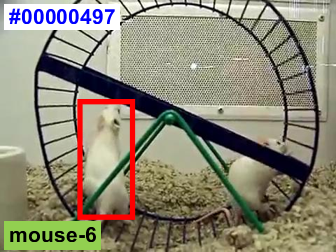}
	\includegraphics[width=2cm]{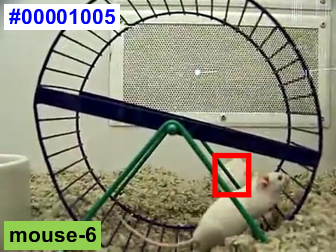} \includegraphics[width=2cm]{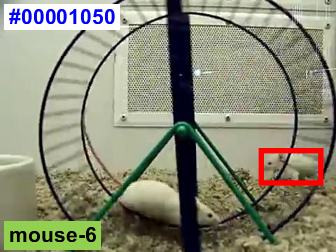} \\
	\scriptsize{{\em Mouse-6}: ``white mouse moving on the ground around another white mouse''} \\
	\caption{Example sequences and annotations of our LaSOT. We focus on long-term videos in which target objects may disappear, and then re-enter the view again. In addition, we provide natural language specification for each sequence. Best viewed in color.}
	\label{annotation_sample}
	\vspace{-2mm}
\end{figure}

\begin{figure}
	\centering
	\includegraphics[width=4.1cm]{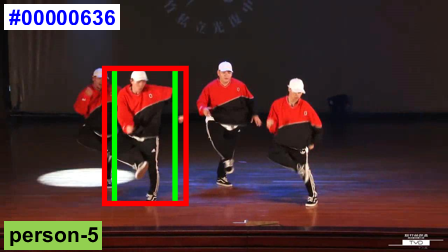}\hfill\includegraphics[width=4.1cm]{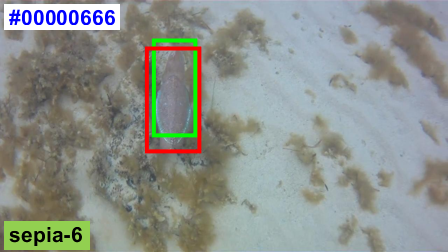} \\
	\includegraphics[width=7.5cm]{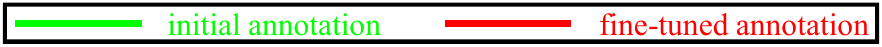} \\
	\caption{ Examples of fine-tuning initial annotations. 
	}
	\label{annotation-revision}
	\vspace{-1em}
\end{figure}

In order to provide consistent bounding box annotation, we define a deterministic annotation strategy. Given a video with a specific tracking target, for each frame, if the target object appears in the frame, a labeler manually draws/edits its bounding box as the tightest up-right one to fit any visible part of the target; otherwise, the labeler gives an absent label, either {\em out-of-view} or {\em full occlusion}, to the frame. Note that, such strategy can not guarantee to minimize the background area in the box, as observed in any other benchmarks. However, the strategy does provide a consistent annotation that is relatively stable for learning the dynamics.

While the above strategy works great most of the time, exceptions exist. Some objects, \eg a mouse, may have long and thin and highly deformable part, \eg a tail, which not only causes serious noise in object appearance and shape, but also provides little information for localizing of the target object. We carefully identify such objects and associated videos in LaSOT, and design specific rules for their annotation (\eg, exclude the tails of mice when drawing their bounding boxes). An example of such cases is shown in the last row of Fig.~\ref{annotation_sample}.

The natural language specification of a sequence is represented by a sentence that describes the color, behavior and surroundings of the target. For LaSOT, we provide 1,400 sentences for all videos. Note that the lingual description aims to provide auxiliary help for tracking. For instance, if a tracker generates proposals for further processing, the lingual specification can assist in reducing the ambiguity among them by serving as a global semantic guidance.

\renewcommand\arraystretch{1.05}
\begin{table*}[!t]\footnotesize
	\centering
	\caption{Descriptions of 14 different attributes in LaSOT.}
	\begin{tabular}{rp{20em}rp{28em}}
		\hline
		Attribute & Definition  & Attribute & Definition\\
		\hline \hline
		{\bf CM }   & Abrupt motion of the camera  & {\bf VC }   & Viewpoint affects target appearance significantly \\
		{\bf ROT }  & The target rotates in the image & {\bf SV }   & The ratio of bounding box is outside the rage [0.5, 2] \\
		{\bf DEF }  & The target is deformable during tracking & {\bf BC }   & The background has the similar appearance as the target \\
		{\bf FOC }  & The target is fully occluded in the sequence & {\bf MB }   & The target region is blurred due to target or camera motion \\
		{\bf IV }   & The illumination in the target region changes & {\bf ARC }  & The ratio of bounding box aspect ratio is outside the rage [0.5, 2] \\
		{\bf OV }   & The target completely leaves the video frame & {\bf LR }   & The target box is smaller than 1000 pixels in at least one frame \\
		{\bf POC }  & The target is partially occluded in the sequence & {\bf FM }   & The motion of the target is larger than the size of its bounding box \\
		\hline
	\end{tabular}%
	\label{tab:tab3}%
\end{table*}%

\begin{figure*}[!t]
	\centering
	\includegraphics[width=\linewidth]{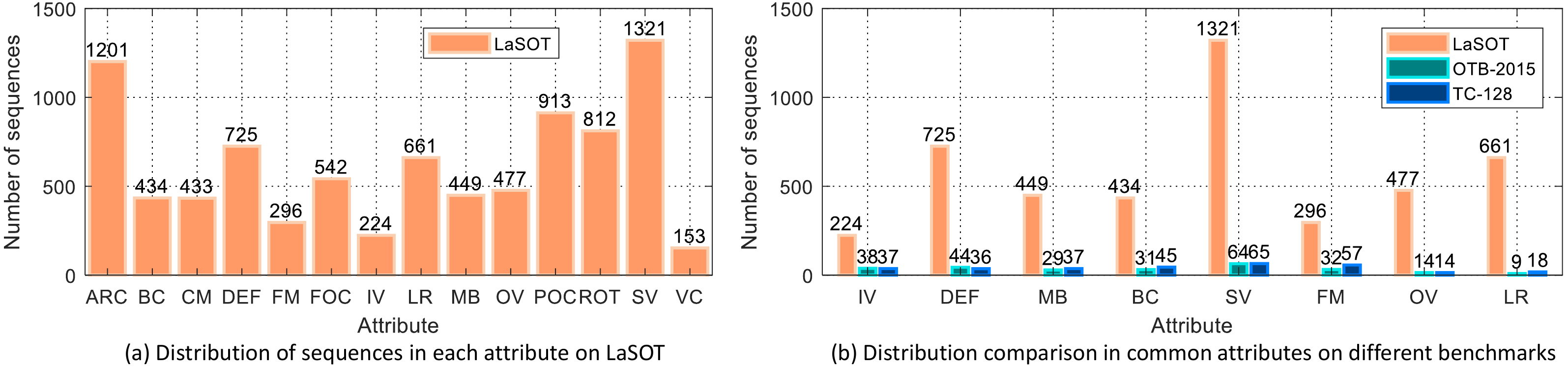}\\
	\caption{Distribution of sequences in each attribute on LaSOT and comparison with other benchmarks. Best viewed in color.}
	\label{fig:att_dis}
\end{figure*}

The greatest effort for constructing a high-quality dense tracking dataset is, apparently, the manual labeling, double-checking, and error correcting. For this task, we have assembled an annotation team containing several Ph.D. students working on related areas and about 10 volunteers.
To guarantee high-quality annotation, each video is processed by teams: a labeling team and a validation team. A labeling team is composed of a volunteer and an expert (Ph.D. student). The volunteer manually draws/edits the target bounding box in each frame, and the expert inspects the results and adjusts them if necessary. Then, the annotation results are reviewed by the validation team containing several (typically three) experts. If an annotation result is not unanimously agreed by the members of validation team, it will be sent back to the original labeling team to revise.

To improve the annotation quality as much as possible, our team checks the annotation results very carefully and revises them frequently. Around 40\% of the initial annotations fail in the first round of validation. And many frames are revised more than three times. Some challenging examples of frames that are initially labeled incorrectly or inaccurately are given in Fig.~\ref{annotation-revision}.
With all these efforts, we finally reach a benchmark with high-quality dense annotation, with some examples shown in Fig.~\ref{annotation_sample}.

\subsection{Attributes}

\label{sec_att}
To enable further performance analysis of trackers, we label each sequence with 14 attributes, including illumination variation (IV), full occlusion (FOC), partial occlusion (POC), deformation (DEF), motion blur (MB), fast motion (FM), scale variation (SV), camera motion (CM), rotation (ROT), background clutter (BC), low resolution (LR), viewpoint change (VC), out-of-view (OV) and aspect ratio change (ARC). The attributions are defined in Tab.~\ref{tab:tab3}, and Fig.~\ref{fig:att_dis} (a) shows the distribution of videos in each attribute.

From Fig.~\ref{fig:att_dis} (a), we observe that the most common challenge factors in LaSOT are scale changes (SV and ARC), occlusion (POC and FOC), deformation (DEF) and rotation (ROT), which are well-known challenges for tracking in real-world applications. Besides, Fig.~\ref{fig:att_dis} (b) demonstrates the distribution of attributes of LaSOT compared to OTB-2015~\cite{wu2015object} and TC-128~\cite{liang2015encoding} on overlapping attributes. From the figure we observe that more than 1,300 videos in LaSOT are involved with scale variations. Compared with OTB-2015 and TC-128 with less than 70 videos with scale changes, LaSOT is more challenging for scale changes. In addition, on the out-of-view attribute, LaSOT comprises 477 sequences, much larger than existing benchmarks.

\subsection{Evaluation Protocols}

Though there is no restriction to use LaSOT, we suggest two evaluation protocols for evaluating tracking algorithms, and conduct evaluations accordingly.

\vspace{0.1em}
\noindent {\bf Protocol \uppercase\expandafter{\romannumeral1}.} In protocol \uppercase\expandafter{\romannumeral1}, we use all 1,400 sequences to evaluate tracking performance. Researchers are allowed to employ any sequences except for those in LaSOT to develop tracking algorithms. Protocol \uppercase\expandafter{\romannumeral1} aims to provide large-scale evaluation of trackers.

\vspace{0.1em}
\noindent {\bf Protocol \uppercase\expandafter{\romannumeral2}.} In protocol \uppercase\expandafter{\romannumeral2}, we split LaSOT into  {\em training} and {\em testing} subsets. According to the 80/20 principle (\ie, the {\em Pareto} principle), we select 16 out of 20 videos in each category for training, and the rest is for testing\footnote{The training/testing split is shown in the {\bf supplementary material}.}. In specific, the {\em training} subset contains 1,120 videos with 2.83M frames, and the {\em testing} subset consists of 280 sequences with 690K frames. The evaluation of trackers is performed on the {\em testing} subset. Protocol \uppercase\expandafter{\romannumeral2} aims to provide a large set of videos for training and assessing trackers in the mean time.

\section{Evaluation}
\label{exp}

\subsection{Evaluation Metric}

\vspace{-2mm}Following popular protocols (\eg OTB-2015~\cite{wu2015object}), we perform an One-Pass Evaluation (OPE) and measure the {\bf precision}, {\bf normalized precision} and {\bf success} of different tracking algorithms under two protocols.

The precision is computed by comparing the distance between tracking result and groundtruth bounding box in pixels. Different trackers are ranked with this metric on a threshold (\eg, 20 pixels). Since the precision metric is sensitive to target size and image resolution, we normalize the precision as in~\cite{muller2018trackingnet}. With the normalized precision metric, we rank tracking algorithms using the Area Under the Curve (AUC) between 0 to 0.5. Please refer to~\cite{muller2018trackingnet} about the normalized precision metric. The success is computed as the Intersection over Union (IoU) between tracking result and groundtruth bounding box. The tracking algorithms are ranked using the AUC between 0 to 1.

\subsection{Evaluated Trackers}

\vspace{-2mm}We evaluate 35 algorithms on LaSOT to provide extensive baselines, comprising deep trackers (\eg, MDNet~\cite{nam2016learning}, TRACA~\cite{choi2018context}, CFNet~\cite{valmadre2017end}, SiamFC~\cite{bertinetto2016fully}, StructSiam~\cite{Zhang2018structured}, DSiam~\cite{guo2017learning}, SINT~\cite{tao2016siamese} and VITAL~\cite{song2018vital}), correlation filter trackers with hand-crafted features (\eg, ECO\_HC~\cite{danelljan2017eco}, DSST~\cite{danelljan2014accurate}, CN~\cite{danelljan2014adaptive}, CSK~\cite{henriques2012exploiting}, KCF~\cite{henriques2015high}, fDSST~\cite{danelljan2017discriminative}, SAMF~\cite{li2014scale}, SCT4~\cite{choi2016visual}, STC~\cite{zhang2014fast} and Staple~\cite{bertinetto2016staple}) or deep features (\eg, HCFT~\cite{ma2015hierarchical} and ECO~\cite{danelljan2017eco}) and regularization techniques (\eg, BACF~\cite{galoogahi2017learning}, SRDCF~\cite{danelljan2015learning}, CSRDCF~\cite{lukezic2017discriminative}, Staple\_CA~\cite{mueller2017context} and STRCF~\cite{li2018learning}), ensemble trackers (\eg, PTAV~\cite{fan2017parallel}, LCT~\cite{ma2015long}, MEEM~\cite{zhang2014meem} and TLD~\cite{kalal2012tracking}), sparse trackers (\eg, L1APG~\cite{bao2012real} and ASLA~\cite{jia2012visual}),  other representatives (\eg, CT~\cite{zhang2012real}, IVT~\cite{ross2008incremental}, MIL~\cite{babenko2009visual} and Struck~\cite{HareST11}). Tab.~\ref{tab:summary_tracker} summarizes these trackers with their representation schemes and search strategies in a chronological order.

\subsection{Evaluation Results with Protocol \uppercase\expandafter{\romannumeral1}}

\renewcommand\arraystretch{1.02}
\begin{table}[!t]\scriptsize
	\centering
	\caption{Summary of evaluated trackers. Representation: Sparse - Sparse Representation, Color - Color Names or Histograms, Pixel - Pixel Intensity, HoG - Histogram of Oriented Gradients, H or B - Haar or Binary, Deep - Deep Feature. Search: PF - Particle Filter, RS - Random Sampling, DS - Dense Sampling.}
	\begin{tabular}{@{}R{1.3cm}@{}R{1.2cm}@{}C{0.7cm}@{}ccccccccc@{}}
		\hline
		\multicolumn{1}{r}{\multirow{3}[0]{*}{ }} & \multicolumn{1}{r}{\multirow{3}[0]{*}{ }} & \multicolumn{7}{c}{\bf Representation}            & \multicolumn{3}{c}{\bf Search}  \\
		\cmidrule(lr){3-9}  \cmidrule(lr){10-12}
		&       & \rotatebox{90}{PCA} & \rotatebox{90}{Sparse} & \rotatebox{90}{Color} & \rotatebox{90}{Pixel} & \rotatebox{90}{HoG} & \rotatebox{90}{H or B} & \rotatebox{90}{Deep} & \rotatebox{90}{PF} & \rotatebox{90}{RS} & \rotatebox{90}{DS}   \\
		\hline \hline
		IVT~\cite{ross2008incremental}           & IJCV08 &    \cmark   &      &     &    &       &        &      &   \cmark    &        &      \\
		MIL~\cite{babenko2009visual}             & CVPR09 &             &       &       &   &     &   H   &       &       &       &   \cmark       \\
		Struck~\cite{HareST11}                   & ICCV11 &             &       &       &   &     &  H    &       &       &       &   \cmark     \\
		L1APG~\cite{bao2012real}                 & CVPR12 &             &   \cmark    &    &    &       &       &       &  \cmark   &       &         \\
		ASLA~\cite{jia2012visual}                & CVPR12 &             &   \cmark    &    &    &       &       &       &   \cmark   &       &         \\
		CSK~\cite{henriques2012exploiting}       & ECCV12 &       &       &       & \cmark  &     &       &       &       &       &   \cmark      \\
		CT~\cite{zhang2012real}                  & ECCV12 &       &       &       &   &     &   H    &       &       &       &   \cmark      \\
		TLD~\cite{kalal2012tracking}             & PAMI12 &       &       &       &  &      &  B     &       &       &       &   \cmark      \\
		CN~\cite{danelljan2014adaptive}          & CVPR14 &       &       &   \cmark    & \cmark  &     &       &       &       &       &    \cmark     \\
		DSST~\cite{danelljan2014accurate}        & BMVC14 &       &       &       & \cmark  &  \cmark   &       &       &       &       &   \cmark      \\
		MEEM~\cite{zhang2014meem}                & ECCV14 &       &       &       & \cmark  &     &       &       &       &  \cmark     &         \\
		STC~\cite{zhang2014fast}                 & ECCV14 &       &       &       & \cmark  &     &       &       &       &       &    \cmark     \\
		SAMF~\cite{li2014scale}                  & ECCVW14 &       &       &   \cmark    & \cmark  & \cmark    &       &       &       &       &   \cmark      \\
		LCT~\cite{ma2015long}                    & CVPR15 &       &       &       &  \cmark  &  \cmark  &       &       &       &       &   \cmark      \\
		SRDCF~\cite{danelljan2015learning}       & ICCV15 &       &       &       &   &  \cmark   &       &       &       &       &    \cmark     \\
		HCFT~\cite{ma2015hierarchical}           & ICCV15 &       &       &       &   &     &       &    \cmark   &       &       &    \cmark     \\
		KCF~\cite{henriques2015high}             & PAMI15 &       &       &       &  &  \cmark    &       &       &       &       &   \cmark      \\
		Staple~\cite{bertinetto2016staple}       & CVPR16 &       &       &   \cmark    &   &  \cmark   &       &       &       &       &    \cmark     \\
		SINT~\cite{tao2016siamese}               & CVPR16 &       &       &       &   &     &       &   \cmark    &       &    \cmark   &         \\
		SCT4~\cite{choi2016visual}               & CVPR16 &       &       &       &   &  \cmark   &       &       &       &       &   \cmark      \\
		MDNet~\cite{nam2016learning}             & CVPR16 &       &       &       &    &    &       &    \cmark   &      &    \cmark    &       \\
		SiamFC~\cite{bertinetto2016fully}        & ECCVW16&       &       &       &   &     &       &    \cmark    &       &       &     \cmark     \\
		Staple\_CA\cite{mueller2017context}      & CVPR17 &       &       &    \cmark   &    &  \cmark  &       &       &       &       &    \cmark     \\
		ECO\_HC~\cite{danelljan2017eco}          & CVPR17 &       &       &       &   &   \cmark  &       &       &       &       &    \cmark     \\
		ECO~\cite{danelljan2017eco}              & CVPR17 &       &       &       &   &     &       &  \cmark     &       &       &    \cmark     \\
		CFNet~\cite{valmadre2017end}             & CVPR17 &       &       &       &   &     &       &   \cmark    &       &       &    \cmark     \\
		CSRDCF~\cite{lukezic2017discriminative}  & CVPR17 &       &       &  \cmark     & \cmark  &  \cmark   &       &       &       &       &   \cmark      \\
		PTAV~\cite{fan2017parallel}              & ICCV17 &       &       &       & \cmark &  \cmark    &       &     \cmark  &       &       &     \cmark    \\
		DSiam~\cite{guo2017learning}             & ICCV17 &       &       &       &    &    &       &    \cmark   &       &       &   \cmark      \\
		BACF~\cite{galoogahi2017learning}        & ICCV17 &       &       &       &    & \cmark   &       &       &       &       &   \cmark      \\
		fDSST~\cite{danelljan2017discriminative} & PAMI17 &       &       &       & \cmark  &  \cmark   &       &       &       &       &    \cmark     \\
		VITAL~\cite{song2018vital}               & CVPR18 &       &       &       &    &    &       &    \cmark   &       &   \cmark         \\
		TRACA~\cite{choi2018context}             & CVPR18 &       &       &       &   &     &       &  \cmark     &       &       &    \cmark     \\
		STRCF~\cite{li2018learning}              & CVPR18 &       &       &       &   &   \cmark  &       &       &       &       &    \cmark     \\
		StructSiam~\cite{Zhang2018structured}    & ECCV18 &       &       &       &    &    &       &   \cmark    &       &       &   \cmark     \\
		\hline
	\end{tabular}%
	\vspace{-3mm}
	\label{tab:summary_tracker}%
\end{table}%

\begin{figure*}[!t]
	\centering
	\includegraphics[width=5.7cm]{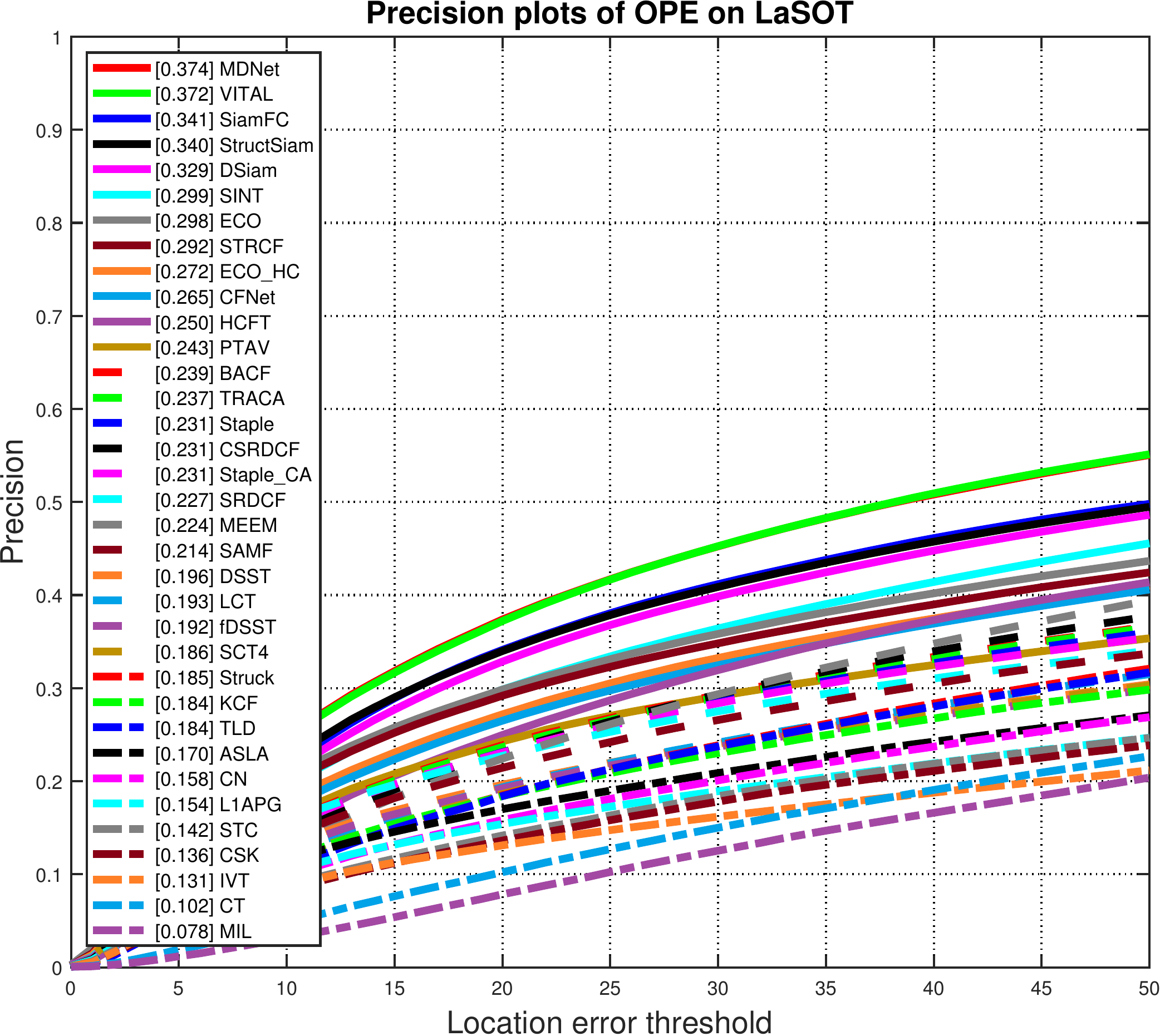}
	\includegraphics[width=5.7cm]{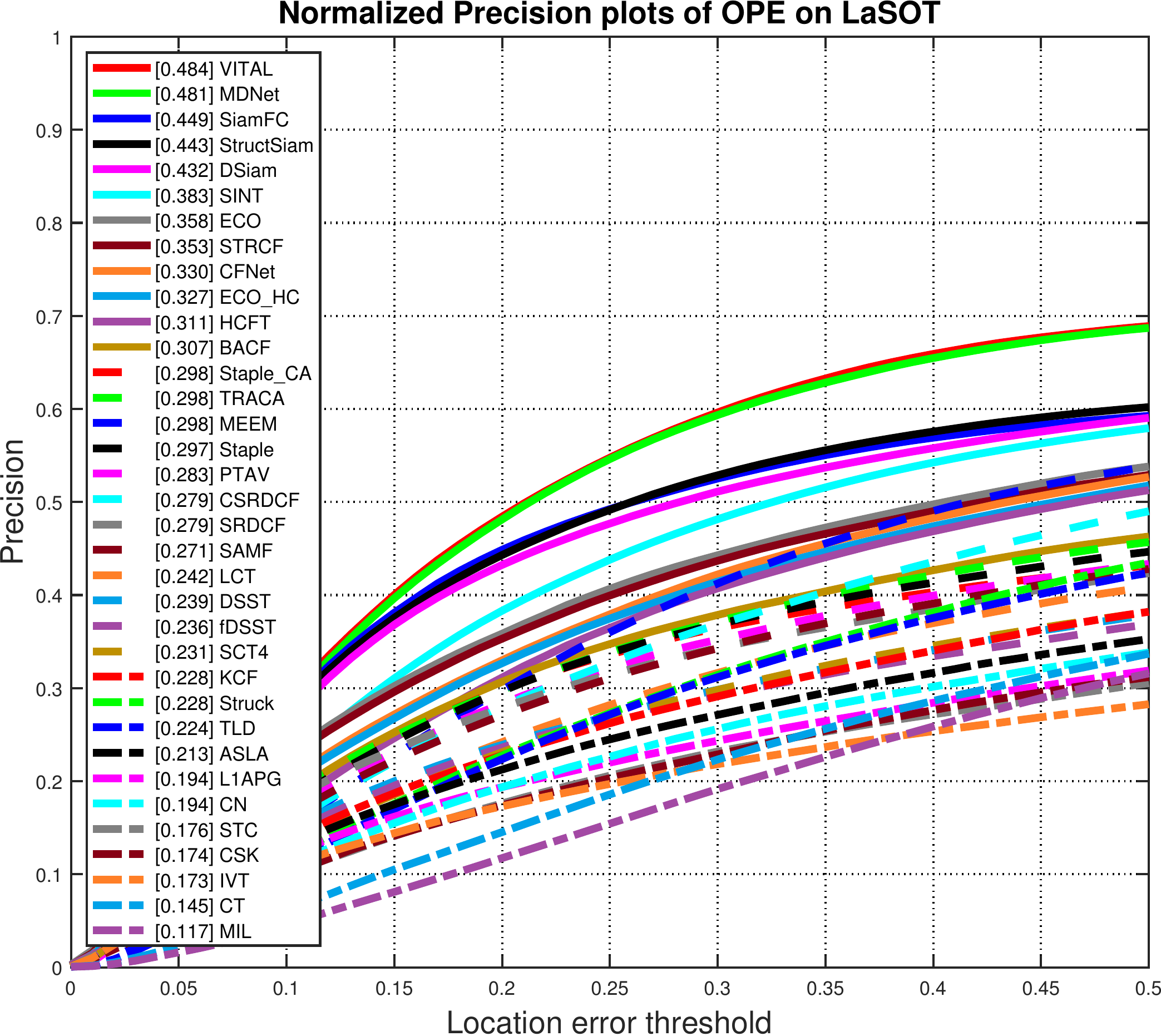}
	\includegraphics[width=5.7cm]{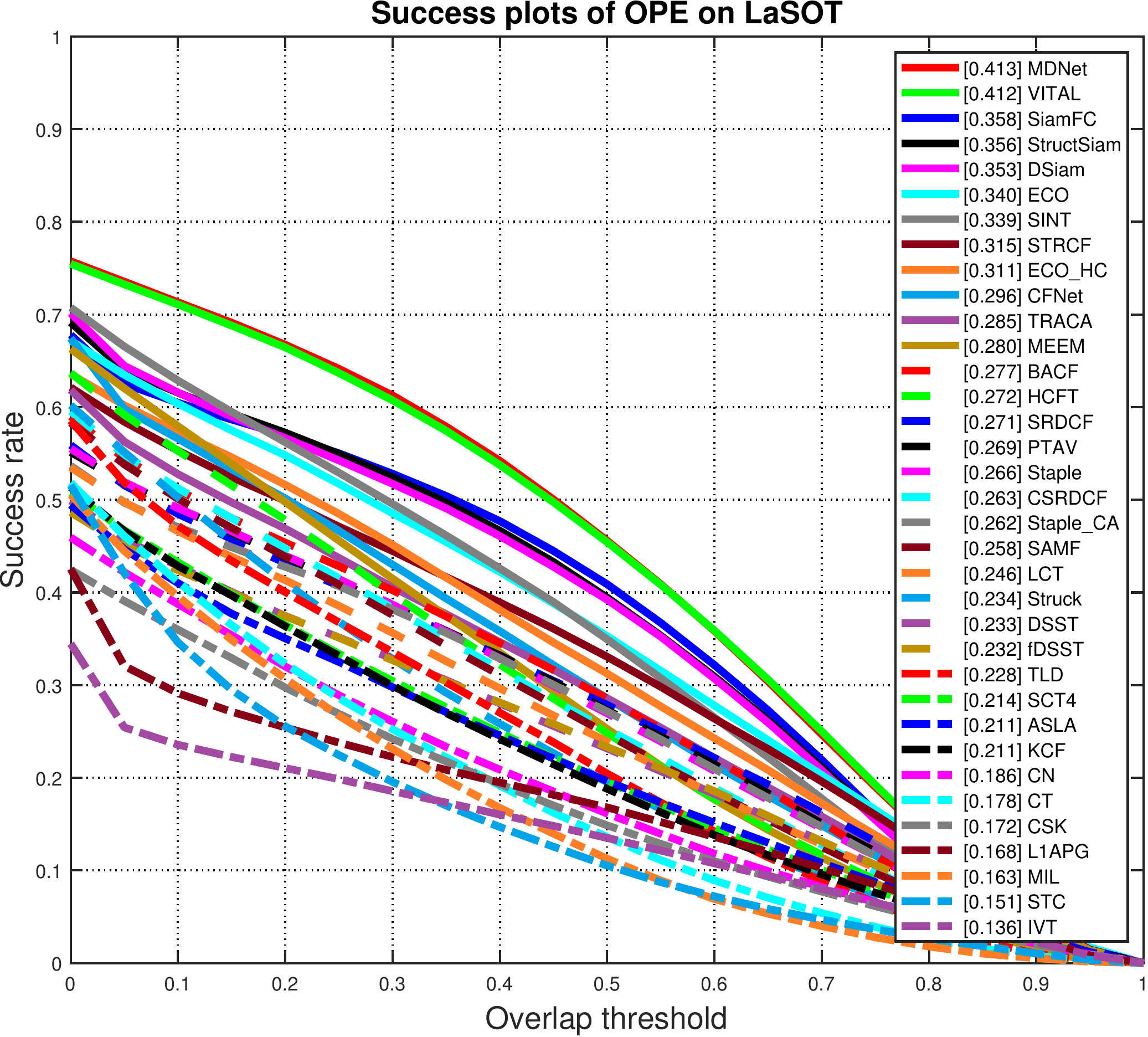}\\
	\caption{Evaluation results on LaSOT under protocol \uppercase\expandafter{\romannumeral1} using precision, normalized precision and success. Best viewed in color.}
	\label{fig:protocol_1_overall_res}
\end{figure*}

\begin{figure*}[!t]
	\centering
	\includegraphics[width=5.7cm]{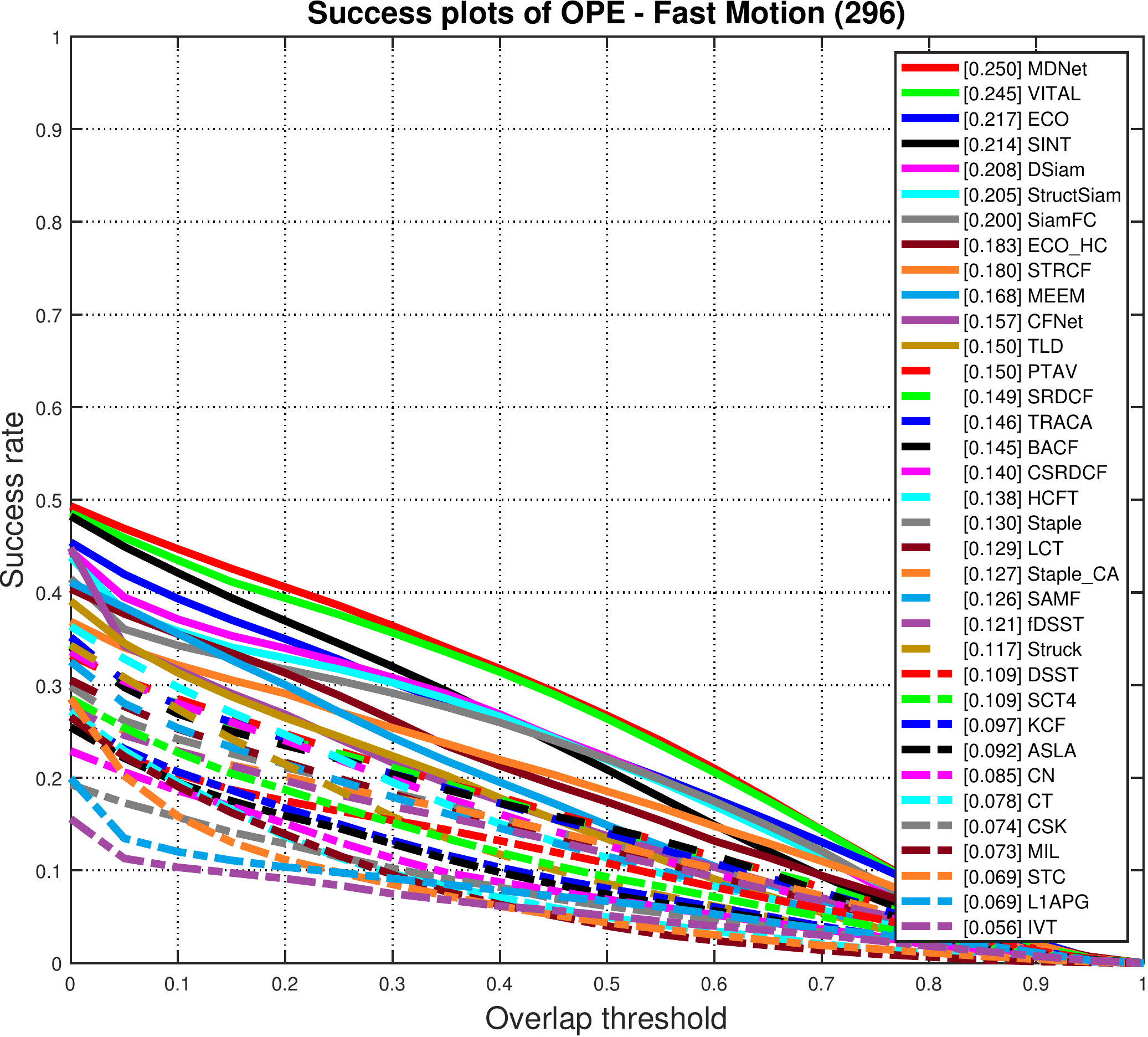}
	\includegraphics[width=5.7cm]{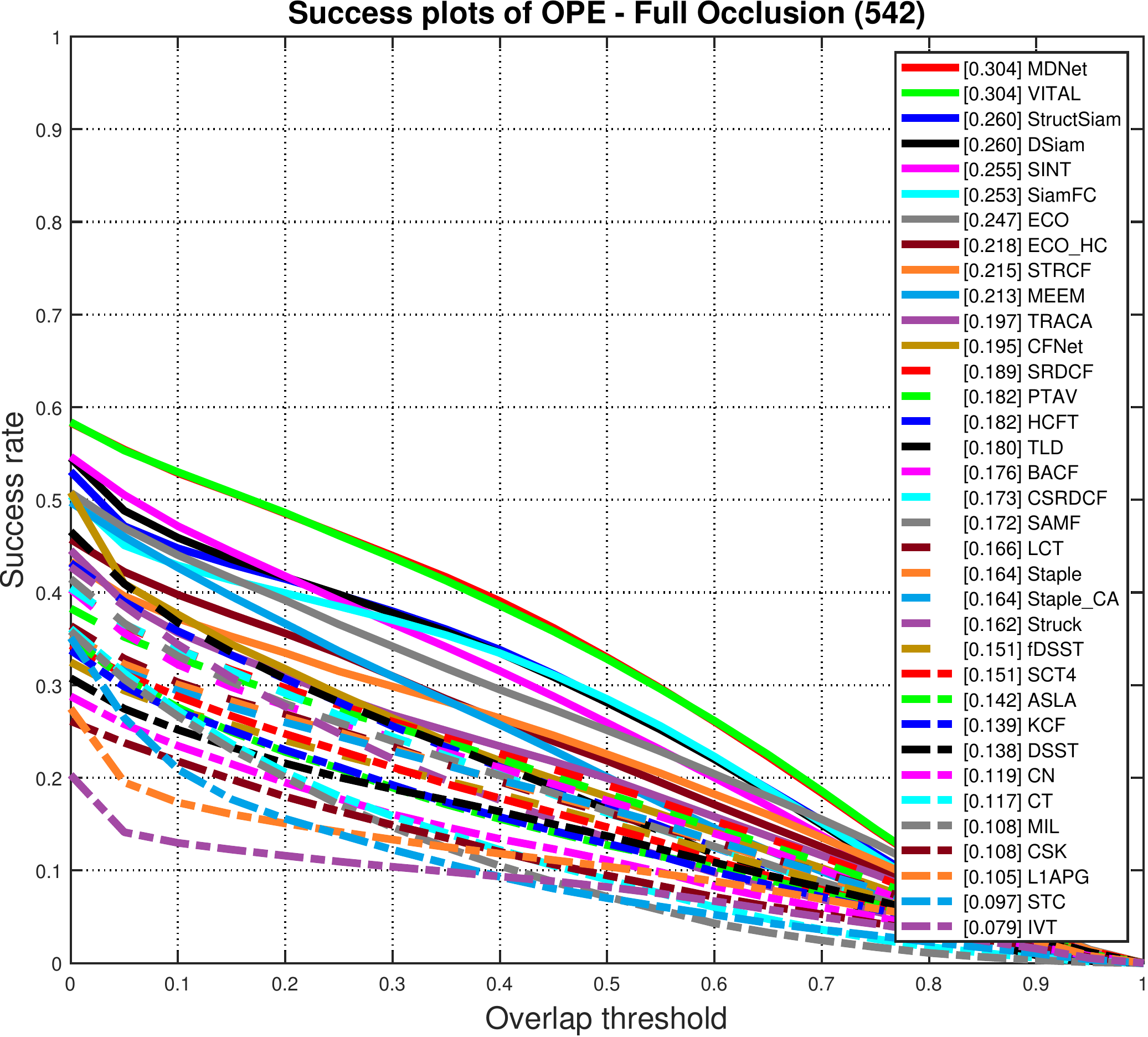}
	\includegraphics[width=5.7cm]{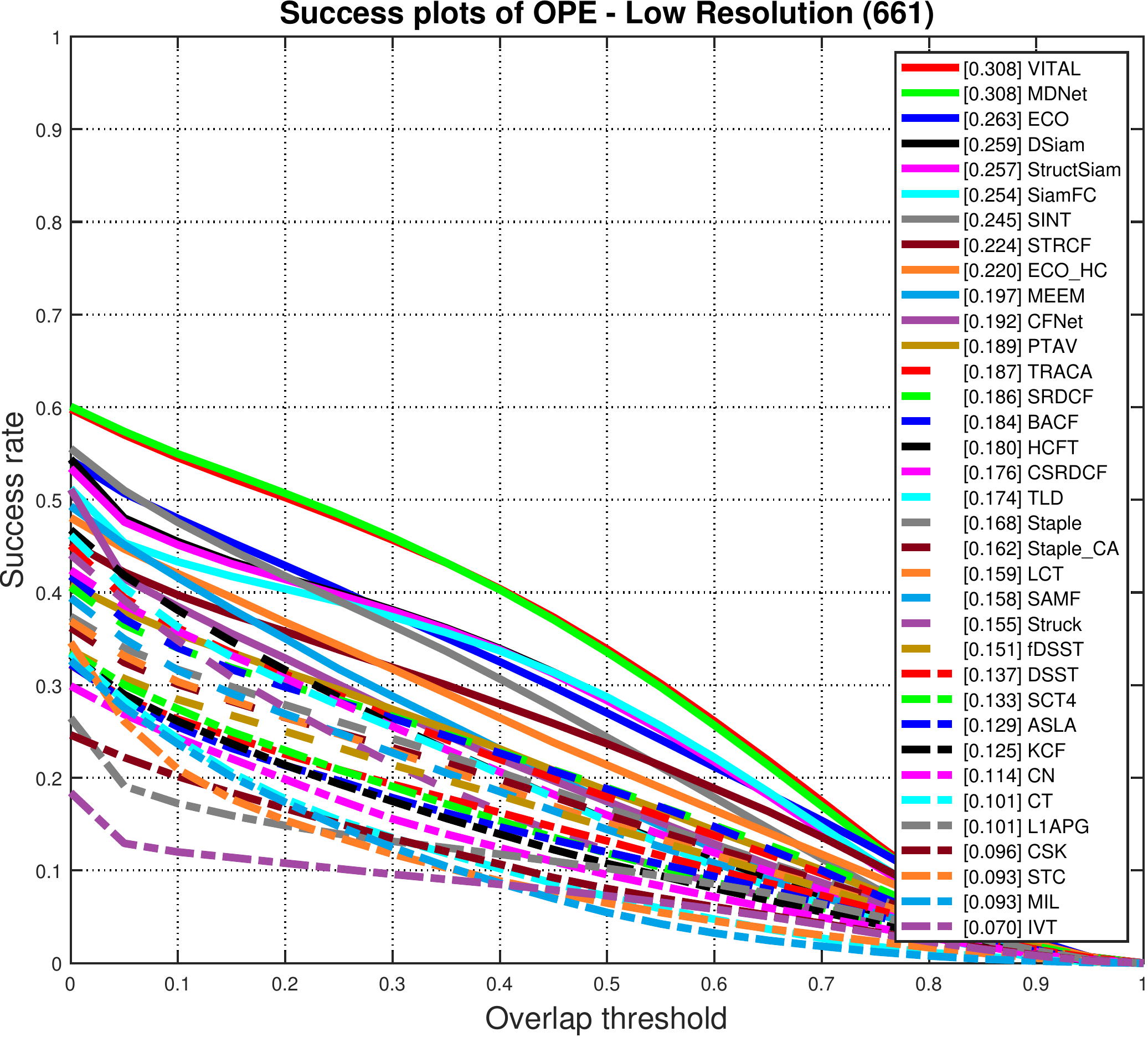}
	\caption{Performances of trackers on three most challenging attributes under protocol \uppercase\expandafter{\romannumeral1} using success. Best viewed in color.}
	\vspace{-.5em}
	\label{fig:protocol_1_att_res}
\end{figure*}

\begin{figure*}[!t]
	\centering
	\begin{tabular}{@{\hspace{.0mm}}c@{\hspace{1.75mm}} @{\hspace{.0mm}}c@{\hspace{.0mm}}}
		\includegraphics[width=2.8cm]{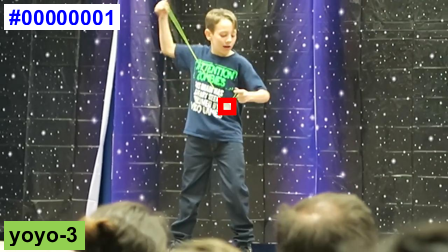} \includegraphics[width=2.8cm]{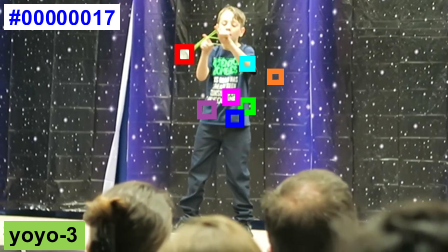} \includegraphics[width=2.8cm]{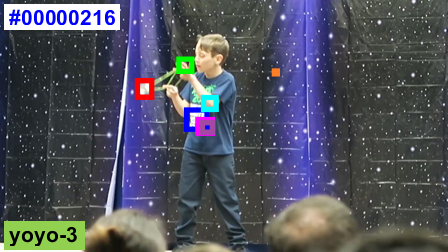}& \includegraphics[width=2.8cm]{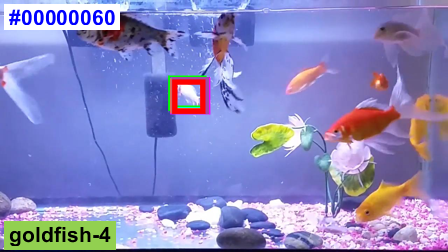} \includegraphics[width=2.8cm]{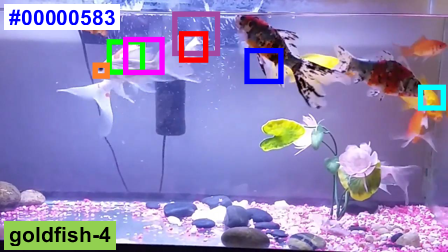} \includegraphics[width=2.8cm]{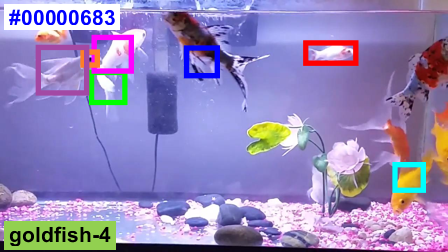}\\
		\includegraphics[width=2.8cm]{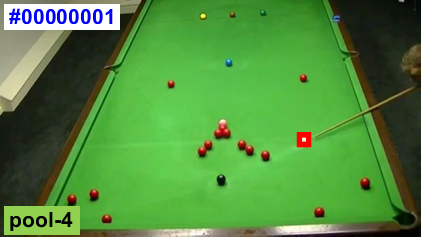} \includegraphics[width=2.8cm]{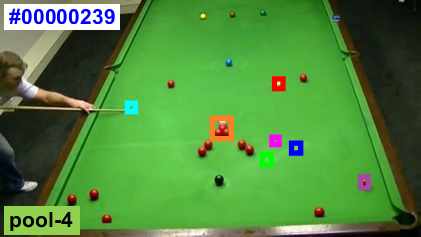} \includegraphics[width=2.8cm]{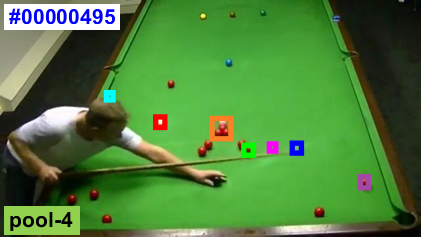}& \includegraphics[width=2.8cm]{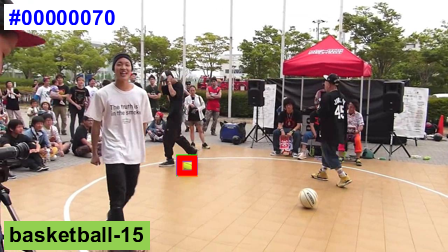} \includegraphics[width=2.8cm]{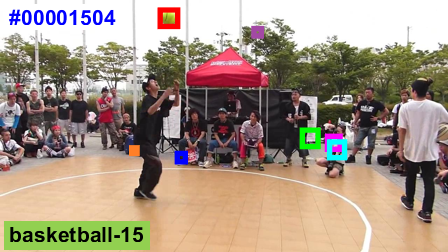} \includegraphics[width=2.8cm]{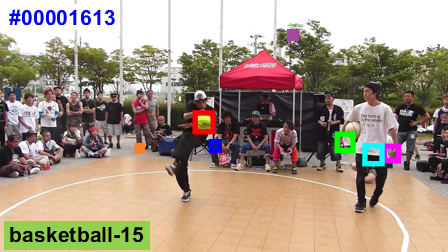}\\
		\includegraphics[width=2.8cm]{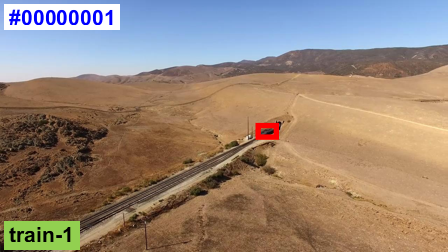} \includegraphics[width=2.8cm]{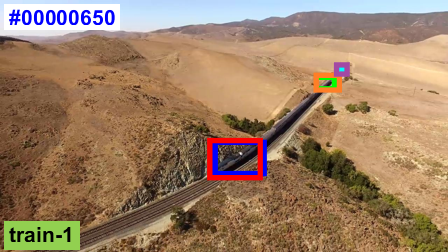} \includegraphics[width=2.8cm]{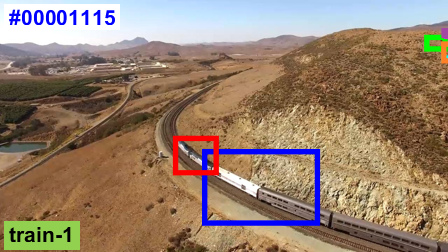}& \includegraphics[width=2.8cm]{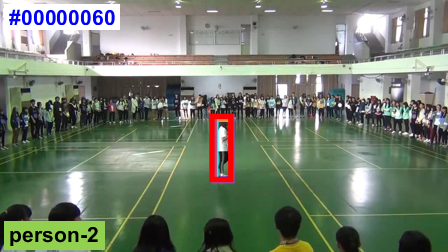} \includegraphics[width=2.8cm]{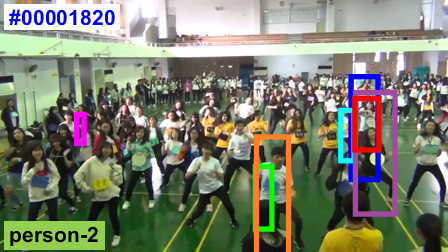} \includegraphics[width=2.8cm]{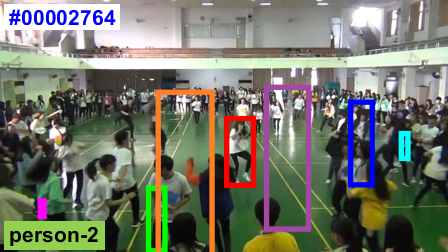}\\
		\multicolumn{2}{c}{\includegraphics[width=15cm,height=0.25cm]{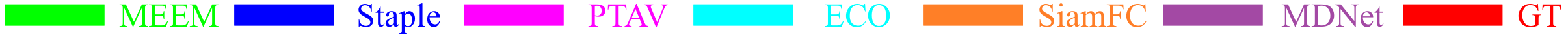}}\\
	\end{tabular}
	\caption{Qualitative evaluation in six typical hard challenges: {\em Yoyo-3} with fast motion, {\em Goldfish-4} with full occlusion, {\em Pool-4} with low-resolution, {\em Basketball-15} with out-of-view, {\em Train-1} with aspect ration change and {\em Person-2} with background clutter. Best viewed in color.}
	\vspace{-5mm}	\label{qua_res}
\end{figure*}

\begin{figure*}[!t]
	\centering
	\includegraphics[width=5.7cm]{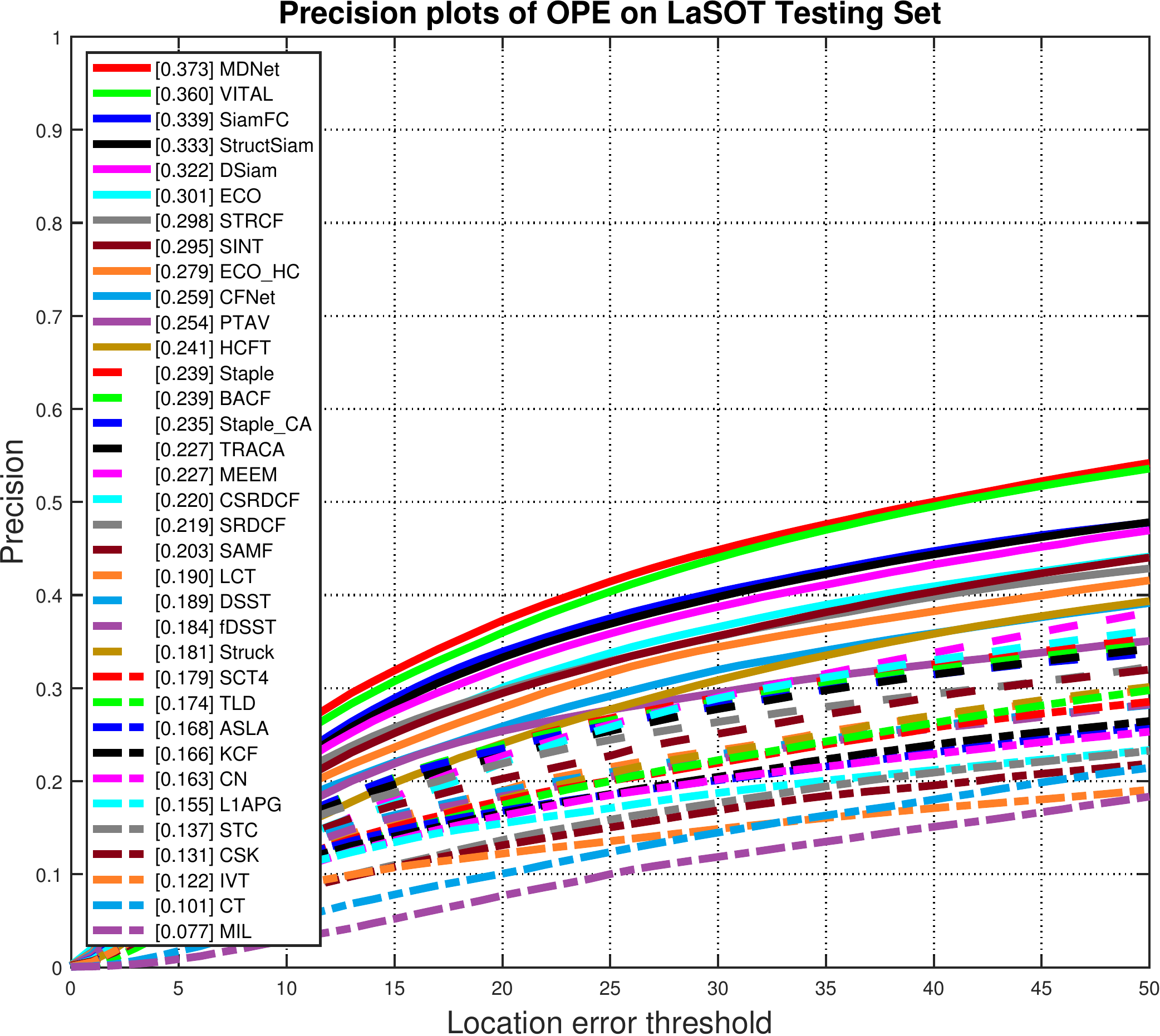}
	\includegraphics[width=5.7cm]{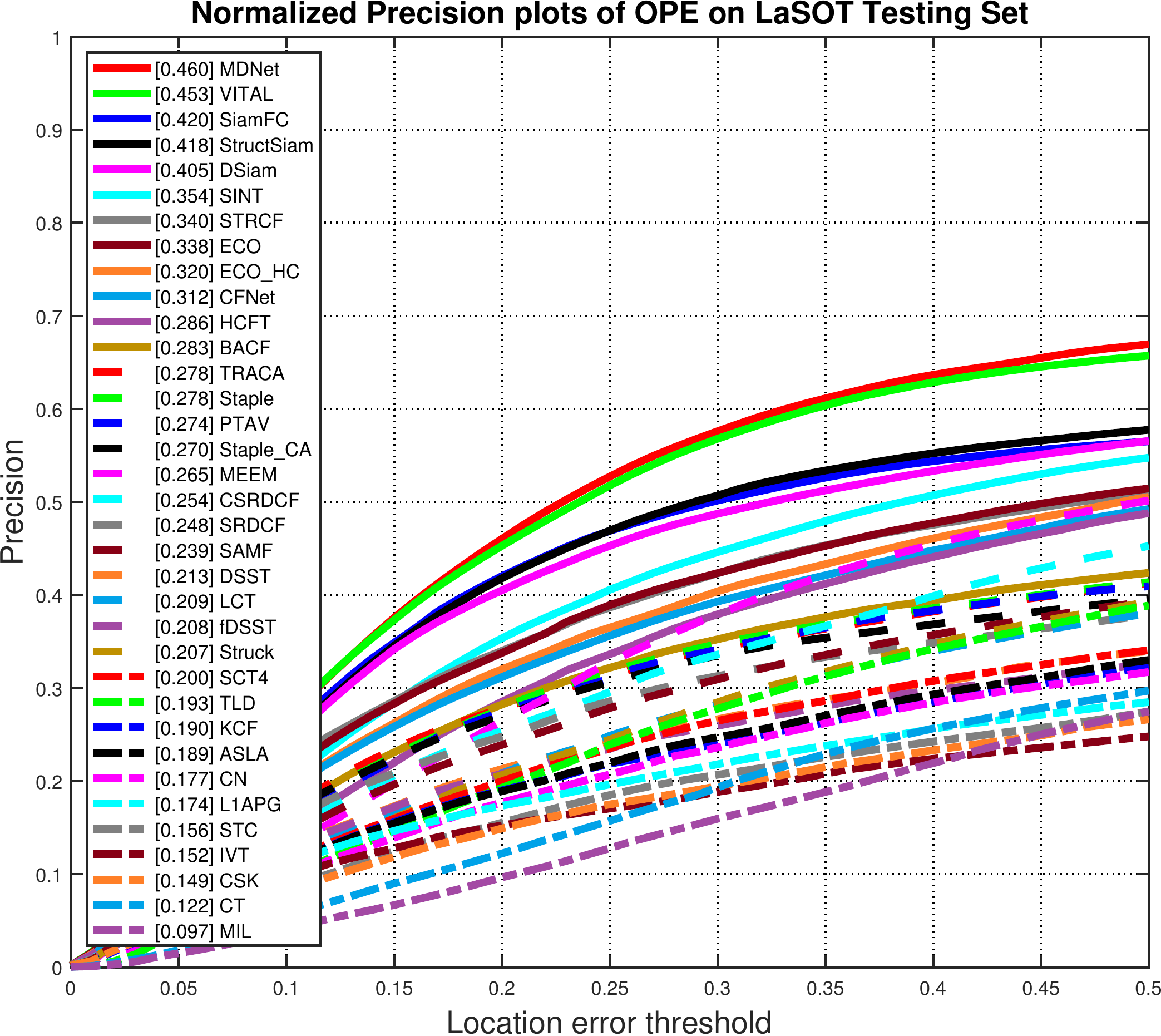}
	\includegraphics[width=5.7cm]{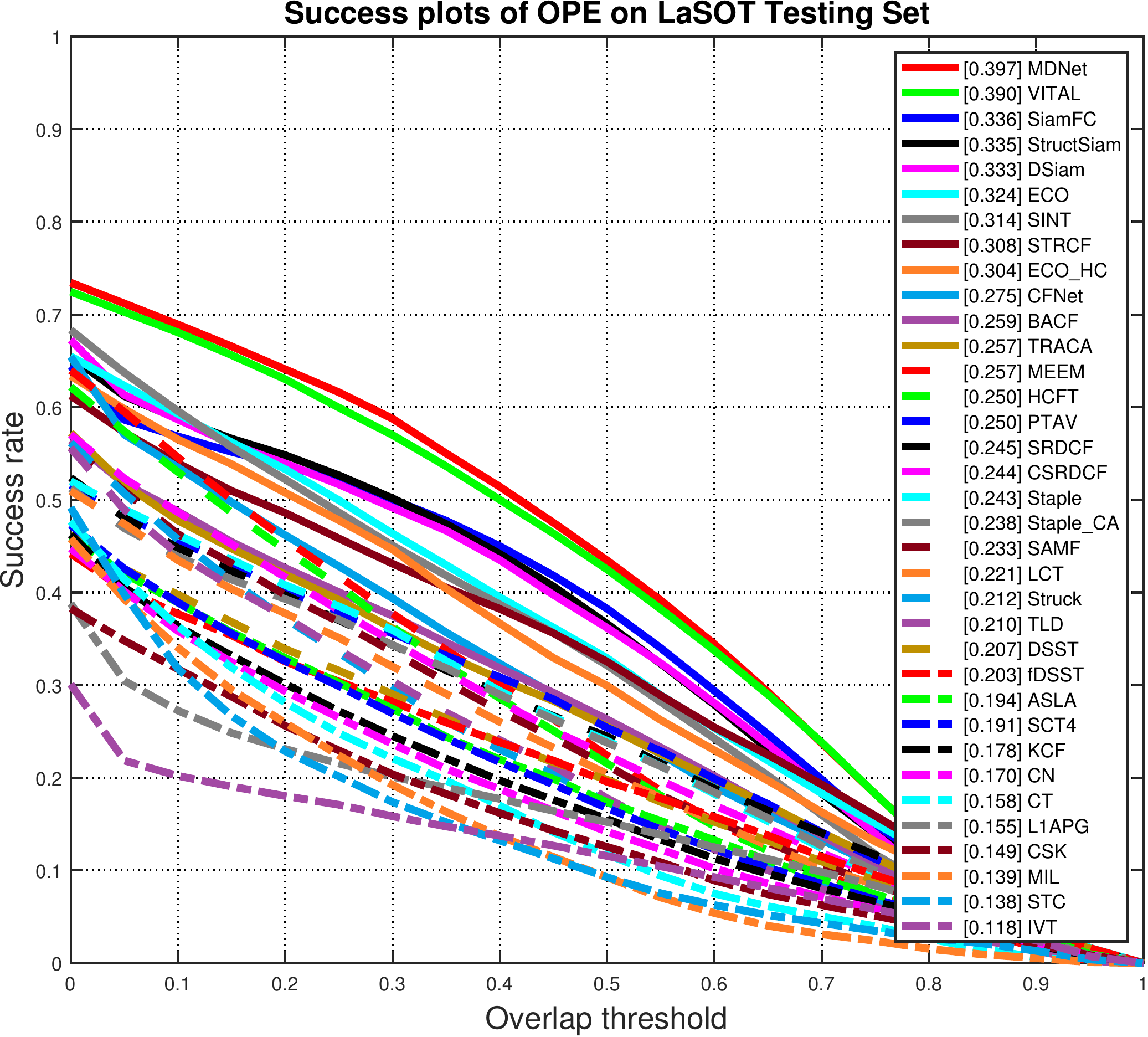}\\
	\caption{Evaluation results on LaSOT under protocol \uppercase\expandafter{\romannumeral2} using precision, normalized precision and success. Best viewed in color.}
	\vspace{-1em}
	\label{fig:protocol_2_overall_res}
\end{figure*}

\vspace{-2mm}\noindent
{\bf Overall performance.} Protocol \uppercase\expandafter{\romannumeral1} aims at providing large-scale evaluations on all 1,400 videos in LaSOT. Each tracker is used as it is for evaluation, without any modification. We report the evaluation results in OPE using precision, normalized precision and success, as shown in Fig.~\ref{fig:protocol_1_overall_res}. MDNet achieves the best precision score of 0.374 and success score of 0.413, and VITAL obtains the best normalized precision score of 0.484. Both MDNet and VITAL are trained in an online fashion, resulting in expensive computation and slow running speeds. SimaFC tracker, which learns off-line a matching function from a large set of videos using deep network, achieves competitive results with 0.341 precision score, 0.449 normalized precision score and 0.358 success score, respectively. Without time-consuming online model adaption, SiamFC runs efficiently in real-time. The best correlation filter tracker is ECO with 0.298 precision score, 0.358 normalized precision score and 0.34 success score.

Compared to the typical tracking performances on existing dense benchmarks (\eg, OTB-2015~\cite{wu2015object}), the performances on LaSOT are severely degraded because of a large mount of non-rigid target objects and challenging factors involved in LaSOT. An interesting observation from Fig.~\ref{fig:protocol_1_overall_res} is that all the top seven trackers leverage deep feature, demonstrating its advantages in handling appearance changes.

\vspace{0.1em}
\noindent
{\bf Attribute-based performance.} To analyze different challenges faced by existing trackers, we evaluate all tracking algorithms on 14 attributes. We show the results on three most challenging attributes, \ie, {\em fast motion}, {\em out-of-view} and {\em full occlusion}, in Fig.~\ref{fig:protocol_1_att_res} and refer the readers to {\bf supplementary material} for detailed attribute-based evaluation.

\vspace{0.1em}
\noindent
{\bf Qualitative evaluation.} To qualitatively analyze different trackers and provide guidance for future research, we show the qualitative evaluation results of six representative trackers, including MDNet, SiamFC, ECO, PTAV, Staple and MEEM, in six typical hard challenges containing {\em fast motion}, {\em full occlusion}, {\em low resolution}, {\em out-of-view}, {\em aspect ratio change} and {\em background clutter} in Fig.~\ref{qua_res}. From Fig.~\ref{qua_res}, we observe that, for videos with {\em fast motion}, {\em full occlusion} and {\em out-of-view} (\eg, {\em Yoyo-3}, {\em Goldfish-4} and {\em Basketball-15}), the trackers are prone to lose the target because existing trackers usually perform localization from a small local region. To handle these challenges, a potential solution is to leverage an instance-specific detector to locate the target for subsequent tracking. Trackers easily drift in video with {\rm low resolution} (\eg, {\em Pool-4}) due to the ineffective representation for small target. A solution for deep feature based trackers is to combine features from multiple scales to incorporate details into representation. Video with {\em aspect ratio change} is difficult as most existing trackers either ignore this issue or adopt a simple method (\eg, random search or pyramid strategy) to deal with it. Inspired from the success of deep learning based object detection, a generic regressor can be leveraged to reduce the effect of {\em aspect ratio change} (and {\em scale change}) on tracking. For sequence with {\em background clutter}, trackers drift due to less discriminative representation for target and background. A possible solution to alleviate this problem is to utilize the contextual information to enhance the discriminability.

\subsection{Evaluation Results with Protocol \uppercase\expandafter{\romannumeral2}}
\vspace{-2mm}
Under protocol \uppercase\expandafter{\romannumeral2}, we split LaSOT into {\em training} and {\em testing} sets. Researchers are allowed to leverage the sequences in the {\em training} set to develop their trackers and assess their performances on the {\em test} set. In order to provide baselines and comparisons on the {\em testing} set, we evaluate the 35 tracking algorithms. Each tracker is used as it is for evaluation without any modification or re-training. The evaluation results are shown in Fig.~\ref{fig:protocol_2_overall_res} using precision, normalized precision and success. We observe consistent results as in protocol \uppercase\expandafter{\romannumeral1}. MDNet and VITAL show top performances with precision scores of 0.373 and 0.36, normalized precision scores of 0.46 and 0.453 and success scores of 0.397 and 0.39. Next, SiamFC achieves the third-ranked performance with a 0.339 precision score, a 0.42 normalized precision score and a 0.336 success score, respectively. Despite slightly lower scores in accuracy than MDNet and VITAL, SiamFC runs much faster and achieves real-time running speed, showing good balance between accuracy and efficiency. For attribute-based evaluation of trackers on LaSOT {\em testing} set, we refer the readers to {\bf supplementary material} because of limited space.

In addition to evaluating each tracking algorithm as it is, we conduct experiments by re-training two representative deep trackers, MDNet~\cite{nam2016learning} and SiamFC~\cite{bertinetto2016fully}, on the {\em training} set of LaSOT and assessing them. The evaluation results show similar performances for these trackers as without retraining. A potential reason is that our re-training may not follow the same configurations used by the original authors. Besides, since LaSOT are in general more challenging than previous datasets (\eg, all sequences are {\em long-term}), dedicated configuration may be needed for training these trackers. We leave this part as a future work since it is beyond the scope of this benchmark.

\subsection{Retraining Experiment on LaSOT}

\renewcommand\arraystretch{1.05}
\begin{table}[!t]\small
	\centering
	\caption{Retraining of SiamFC~\cite{bertinetto2016fully} on LaSOT.}
	\begin{tabular}{crcc}
		\hline
		&       & \multicolumn{2}{c}{SiamFC-3s (color)} \\
		\cline{3-4}
		\multicolumn{2}{c}{Training data} & \multicolumn{1}{c}{\tabincell{c}{ImageNet \\ Video~\cite{russakovsky2015imagenet}}} & \multicolumn{1}{c}{\tabincell{c}{LaSOT \\ training set}} \\
		\hline
		\multirow{2}[0]{*}{OTB-2013~\cite{wu2013online}} & Precision &   0.803    &  0.816 ($\uparrow${\bf 1.3\%}) \\
		& Success &   0.588    &  0.608 ($\uparrow${\bf 2.0\%})\\
		\hline
		\multirow{2}[0]{*}{OTB-2015~\cite{wu2015object}} & Precision &    0.756   & 0.777 ($\uparrow${\bf 2.1\%})\\
		& Success &   0.565    & 0.582 ($\uparrow${\bf 1.7\%})\\
		\hline
	\end{tabular}%
	\label{tab:retrain}%
	\vspace{-5mm}
\end{table}%

\vspace{-2mm}We conduct the experiment by retraining SiamFC~\cite{bertinetto2016fully} on the training set of LaSOT to demonstrate how deep learning based tracker is improved using more data. Tab.~\ref{tab:retrain} reports the results on OTB-2013~\cite{wu2013online} and OTB-2015~\cite{wu2015object} and comparisons with the performance of original SiamFC trained on ImageNet Video~\cite{russakovsky2015imagenet}. Note that, we utilize color images for training, and apply a pyramid with 3 scales for tracking, \ie, SiamFC-3s (color). All parameters for training and tracking are kept the same in these two experiments. From Tab.~\ref{tab:retrain}, we observe consistent performance gains on the two benchmarks, showing the importance of specific large-scale training set for deep trackers.

\section{Conclusion}
\label{con}

\vspace{-2mm}We present LaSOT with high-quality dense bounding box annotations for visual object tracking. To the best of our knowledge, LaSOT is the {\em largest} tracking benchmark with high quality annotations to date. By releasing LaSOT, we expect to provide the tracking community a dedicated platform for training deep trackers and assessing long-term tracking performance. Besides, LaSOT provides lingual annotations for each sequence, aiming to encourage the exploration on integrating visual and lingual features for robust tracking. By releasing LaSOT, we hope to narrow the gap between the increasing number of deep trackers and the lack of large dedicated datasets for training, and meanwhile provide more veritable evaluations for different trackers in the wild. Extensive evaluations on LaSOT under two protocols imply a large room to improvement for visual tracking.

\vspace{0.1em}
\noindent
{\small {\bf Acknowledgement.} We sincerely thank B. Huang, X. Li, Q. Zhou, L. Chen, J. Liang, J. Wang and anonymous volunteers for their help in constructing LaSOT. This work is supported in part by the China National Key Research and Development Plan (Grant No. 2016YFB1001200), and in part by US NSF Grants 1618398, 1449860 and 1350521, and Yong Xu thanks the supports by National Nature Science Foundation of China (U1611461 and 61672241), the Cultivation Project of Major Basic Research of NSF-Guangdong Province (2016A030308013).}

\begin{onecolumn}

\begin{center}
	{\Large \bf LaSOT: A High-quality Benchmark for Large-scale Single Object Tracking \vspace*{6pt} \\ ----- Supplementary Material -----}
	\vspace*{1pt}
\end{center}

\renewcommand\arraystretch{0.99}
\begin{table}[!htbp]\scriptsize
	\centering
	\caption{Details of 70 object categories in LaSOT and comparison with existing dense benchmark. Best viewed when zoomed-in.}
	\begin{tabular}{c@{}C{0.9cm}c@{}C{0.9cm}c@{}C{0.9cm}c@{}C{0.9cm}c@{}C{0.9cm}c@{}C{0.9cm}c@{}C{0.9cm}}
		\hline
		\multicolumn{2}{c}{\bf NUS-PRO~\cite{li2016nus}} & \multicolumn{2}{c}{\bf OTB-2015~\cite{wu2015object}} & \multicolumn{2}{c}{\bf TC-128~\cite{liang2015encoding}} & \multicolumn{2}{c}{\bf UAV123~\cite{mueller2016benchmark}} & \multicolumn{2}{c}{\bf VOT-17~\cite{kristan2017visual}} & \multicolumn{2}{c}{\bf NfS~\cite{galoogahi2017need}} & \multicolumn{2}{c}{\bf LaSOT} \\
		class & \# entries & class & \# entries & class & \# entries & class & \# entries & class & \# entries & class & \# entries & \multicolumn{1}{c}{\;\;\;class} & \# entries \\
		\cmidrule(lr){1-2} \cmidrule(lr){3-4} \cmidrule(lr){5-6} \cmidrule(lr){7-8} \cmidrule(lr){9-10} \cmidrule(lr){11-12} \cmidrule(lr){13-14} \cmidrule(lr){3-4}
		person & 193   & person & 36    & person & 45    & person & 48    & person & 19    & ball  & 21    & airplane & 20 \\
		head  & 60    & head  & 26    & head  & 16    & car   & 30    & head  & 5     & person & 20    & basketball & 20 \\
		car   & 31    & car   & 12    & sphere & 8     & drone & 10    & fish  & 4     & animal & 10    & bear  & 20 \\
		airplane & 20    & toy   & 8     & 2D print & 5     & wakeboard & 10    & motorcycle & 4     & vehicle & 9     & bicycle & 20 \\
		boat  & 20    & 2D print & 4     & bicycle & 5     & boat  & 9     & car   & 3     & shuffeboard & 8     & bird  & 20 \\
		helicopter & 20    & cuboid & 3     & car   & 5     & building & 5     & drone & 3     & face  & 6     & boat  & 20 \\
		motorcycle & 20    & bird  & 2     & ball  & 4     & truck & 5     & ant   & 2     & cup   & 4     & book  & 20 \\
		drone & 1     & motorcycle & 1     & toy   & 4     & bicycle & 3     & ball  & 2     & dollar & 4     & bottle & 20 \\
		-     & -     & deer  & 1     & hand  & 3     & bird  & 3     & bird  & 2     & aircraft & 4     & bus   & 20 \\
		-     & -     & bottle & 1     & kite  & 3     & -     & -     & toy   & 2     & airboard & 2     & car   & 20 \\
		-     & -     & panda & 1     & logo  & 3     & -     & -     & bag   & 1     & fish  & 2     & cat   & 20 \\
		-     & -     & board & 1     & cuboid & 3     & -     & -     & book  & 1     & motorcycle & 2     & cattle & 20 \\
		-     & -     & can   & 1     & boat  & 2     & -     & -     & butterfly & 1     & drone & 2     & chameleon & 20 \\
		-     & -     & dog   & 1     & cup   & 2     & -     & -     & cable & 1     & bicycle & 2     & coin  & 20 \\
		-     & -     & transformer & 1     & fish  & 2     & -     & -     & crab  & 1     & bird  & 2     & crab  & 20 \\
		-     & -     & bicycle & 1     & guitar & 2     & -     & -     & cat   & 1     & bag   & 1     & crocodile & 20 \\
		-     & -     & -     & -     & bird  & 2     & -     & -     & flamingo & 1     & yoyo  & 1     & cup   & 20 \\
		-     & -     & -     & -     & microphone & 2     & -     & -     & frisbee & 1     & -     & -     & deer  & 20 \\
		-     & -     & -     & -     & torso & 2     & -     & -     & glove & 1     & -     & -     & dog   & 20 \\
		-     & -     & -     & -     & motorcycle & 2     & -     & -     & hand  & 1     & -     & -     & drone & 20 \\
		-     & -     & -     & -     & airplane & 2     & -     & -     & helicopter & 1     & -     & -     & electricFan & 20 \\
		-     & -     & -     & -     & board & 1     & -     & -     & leaf  & 1     & -     & -     & elephant & 20 \\
		-     & -     & -     & -     & bottle & 1     & -     & -     & rabbit & 1     & -     & -     & flag  & 20 \\
		-     & -     & -     & -     & can   & 1     & -     & -     & sheep & 1     & -     & -     & fox   & 20 \\
		-     & -     & -     & -     & deer  & 1     & -     & -     & -     & -     & -     & -     & frog  & 20 \\
		-     & -     & -     & -     & ring  & 1     & -     & -     & -     & -     & -     & -     & gameTarget & 20 \\
		-     & -     & -     & -     & torus & 1     & -     & -     & -     & -     & -     & -     & gecko & 20 \\
		-     & -     & -     & -     & -     & -     & -     & -     & -     & -     & -     & -     & giraffe & 20 \\
		-     & -     & -     & -     & -     & -     & -     & -     & -     & -     & -     & -     & goldfish & 20 \\
		-     & -     & -     & -     & -     & -     & -     & -     & -     & -     & -     & -     & gorilla & 20 \\
		-     & -     & -     & -     & -     & -     & -     & -     & -     & -     & -     & -     & guitar & 20 \\
		-     & -     & -     & -     & -     & -     & -     & -     & -     & -     & -     & -     & hand  & 20 \\
		-     & -     & -     & -     & -     & -     & -     & -     & -     & -     & -     & -     & hat   & 20 \\
		-     & -     & -     & -     & -     & -     & -     & -     & -     & -     & -     & -     & helmet & 20 \\
		-     & -     & -     & -     & -     & -     & -     & -     & -     & -     & -     & -     & hippo & 20 \\
		-     & -     & -     & -     & -     & -     & -     & -     & -     & -     & -     & -     & horse & 20 \\
		-     & -     & -     & -     & -     & -     & -     & -     & -     & -     & -     & -     & kangaroo & 20 \\
		-     & -     & -     & -     & -     & -     & -     & -     & -     & -     & -     & -     & kite  & 20 \\
		-     & -     & -     & -     & -     & -     & -     & -     & -     & -     & -     & -     & leopard & 20 \\
		-     & -     & -     & -     & -     & -     & -     & -     & -     & -     & -     & -     & licensePlate & 20 \\
		-     & -     & -     & -     & -     & -     & -     & -     & -     & -     & -     & -     & lion  & 20 \\
		-     & -     & -     & -     & -     & -     & -     & -     & -     & -     & -     & -     & lizard & 20 \\
		-     & -     & -     & -     & -     & -     & -     & -     & -     & -     & -     & -     & microphone & 20 \\
		-     & -     & -     & -     & -     & -     & -     & -     & -     & -     & -     & -     & monkey & 20 \\
		-     & -     & -     & -     & -     & -     & -     & -     & -     & -     & -     & -     & motorcycle & 20 \\
		-     & -     & -     & -     & -     & -     & -     & -     & -     & -     & -     & -     & mouse & 20 \\
		-     & -     & -     & -     & -     & -     & -     & -     & -     & -     & -     & -     & person & 20 \\
		-     & -     & -     & -     & -     & -     & -     & -     & -     & -     & -     & -     & pig   & 20 \\
		-     & -     & -     & -     & -     & -     & -     & -     & -     & -     & -     & -     & pool  & 20 \\
		-     & -     & -     & -     & -     & -     & -     & -     & -     & -     & -     & -     & rabbit & 20 \\
		-     & -     & -     & -     & -     & -     & -     & -     & -     & -     & -     & -     & racing & 20 \\
		-     & -     & -     & -     & -     & -     & -     & -     & -     & -     & -     & -     & robot & 20 \\
		-     & -     & -     & -     & -     & -     & -     & -     & -     & -     & -     & -     & rubicCube & 20 \\
		-     & -     & -     & -     & -     & -     & -     & -     & -     & -     & -     & -     & sepia & 20 \\
		-     & -     & -     & -     & -     & -     & -     & -     & -     & -     & -     & -     & shark & 20 \\
		-     & -     & -     & -     & -     & -     & -     & -     & -     & -     & -     & -     & shark & 20 \\
		-     & -     & -     & -     & -     & -     & -     & -     & -     & -     & -     & -     & sheep & 20 \\
		-     & -     & -     & -     & -     & -     & -     & -     & -     & -     & -     & -     & skateboard & 20 \\
		-     & -     & -     & -     & -     & -     & -     & -     & -     & -     & -     & -     & spider & 20 \\
		-     & -     & -     & -     & -     & -     & -     & -     & -     & -     & -     & -     & squirrel & 20 \\
		-     & -     & -     & -     & -     & -     & -     & -     & -     & -     & -     & -     & surfboard & 20 \\
		-     & -     & -     & -     & -     & -     & -     & -     & -     & -     & -     & -     & swing & 20 \\
		-     & -     & -     & -     & -     & -     & -     & -     & -     & -     & -     & -     & tank  & 20 \\
		-     & -     & -     & -     & -     & -     & -     & -     & -     & -     & -     & -     & tiger & 20 \\
		-     & -     & -     & -     & -     & -     & -     & -     & -     & -     & -     & -     & train & 20 \\
		-     & -     & -     & -     & -     & -     & -     & -     & -     & -     & -     & -     & truck & 20 \\
		-     & -     & -     & -     & -     & -     & -     & -     & -     & -     & -     & -     & turtle & 20 \\
		-     & -     & -     & -     & -     & -     & -     & -     & -     & -     & -     & -     & umbrella & 20 \\
		-     & -     & -     & -     & -     & -     & -     & -     & -     & -     & -     & -     & volleyball & 20 \\
		-     & -     & -     & -     & -     & -     & -     & -     & -     & -     & -     & -     & yoyo  & 20 \\
		\hline
	\end{tabular}%
	\label{tab:lasot_cat}%
\end{table}%
	
\setcounter{section}{0}
\section{Details of 70 Object Categories in LaSOT and Comparison with Existing Dense Benchmarks}

LaSOT consists of 70 object categories with each containing 20 videos, as shown in Tab.~\ref{tab:lasot_cat}. Most of 70 classes are chosen form the 1,000 classes in ImageNet~\cite{deng2009imagenet}, with a few exceptions such as {\em drone} and {\em gametarget}, which are carefully selected by the experts for tracking. The selection of each category must be agreed upon by all the experts to ensure its usability for visual tracking. In addition, we also compare the object categories of different dense benchmarks. As shown in Tab.~\ref{tab:lasot_cat}, the number of object categories in LaSOT is two times more than that of existing benchmarks (\eg, TC-128~\cite{liang2015encoding} with 27 classes). Moreover, LaSOT eliminates the category bias of dataset for tracking while others do not.

\section{Traing/Testing Split in Protocol \uppercase\expandafter{\romannumeral2}}

In protocol \uppercase\expandafter{\romannumeral2}, we split LaSOT into {\em training} and {\em testing} sets. The {\em training} set contains of 1,120 videos (\ie, 16 sequences for each category) with 2.83M frames in total. The rest 280 videos (\ie, 4 sequences for each category) with 690K frames are used for testing.

\renewcommand\arraystretch{1.05}
\begin{table}[htbp]
	\centering
	\caption{Comparison between {\em training} and {\em testing} sets of LaSOT.}
	\begin{tabular}{rccccccc}
		\hline
		& Video & Min frames & Mean frames & Median frames & Max frames & Total frames & Total duration \\
		\hline \hline
		LaSOT$_{\mathrm{training}}$ & 1,120  & 1,000  & 2,529 & 2,043 & 11,397 & 2.83$\mathbf{M}$  & 26.2 $\mathbf{hours}$ \\
		LaSOT$_{\mathrm{testing}}$ & 280   & 1,000  & 2,448 & 2,102 & 9,999 & 690$\mathbf{K}$   & 6.3 $\mathbf{hours}$ \\
		LaSOT & 1,400 & 1,000 & 2,506 & 2,053 & 11,397 & 3.52$\mathbf{M}$ & 32.5 $\mathbf{hours}$ \\
		\hline
	\end{tabular}%
	\label{tab:training_testing}%
\end{table}%

\begin{figure*}[!hbpt]
	\centering
	\includegraphics[width=\linewidth]{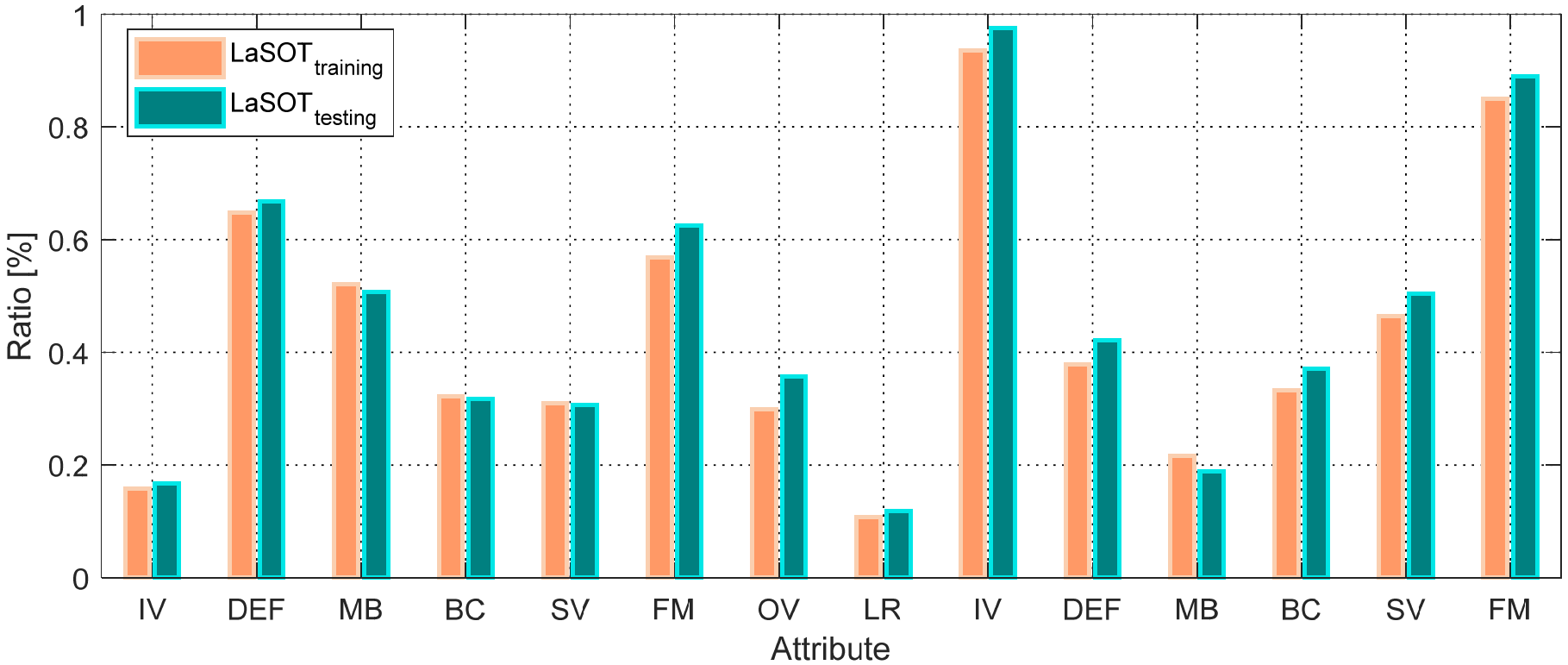}\\
	\caption{Comparison of sequence distribution in each attribute between {\em training} and {\em testing} sets. Best viewed in color.}
	\label{fig:train_test_att}
\end{figure*}

Tab.~\ref{tab:training_testing} reports the detailed comparison between the {\em training} and the {\em testing} sets of LaSOT. We observe that the {\em min frames}, {\em mean frames}, {\em median frames} and {\em max frames} are similar between these two subsets. In addition, as shown in Fig.~\ref{fig:train_test_att}, we can see that the ratios of sequences in all 14 attributes are similar. Both Tab.~\ref{tab:training_testing} and Fig.~\ref{fig:train_test_att} evidence the consistency of our training/testing split.

\newpage
\section{Detailed Attribute-based Performance under Protocol \uppercase\expandafter{\romannumeral1}}

Fig.~\ref{fig:protocol_1_all_att_res_precision} shows the performance of trackers on each attribute using precision under protocol \uppercase\expandafter{\romannumeral1}.
\begin{figure*}[!hbpt]
	\centering
	\includegraphics[width=4.55cm]{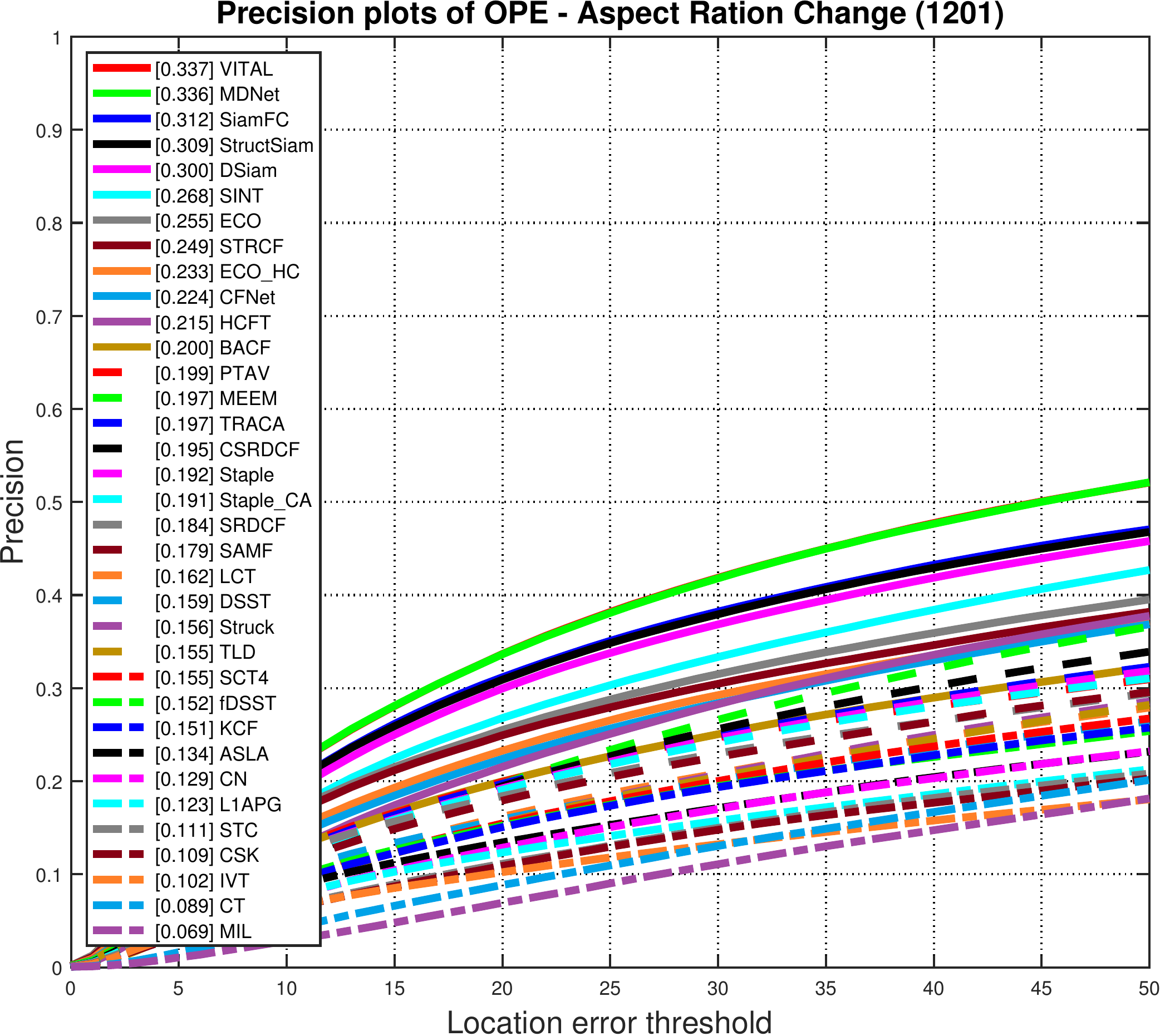}
	\includegraphics[width=4.55cm]{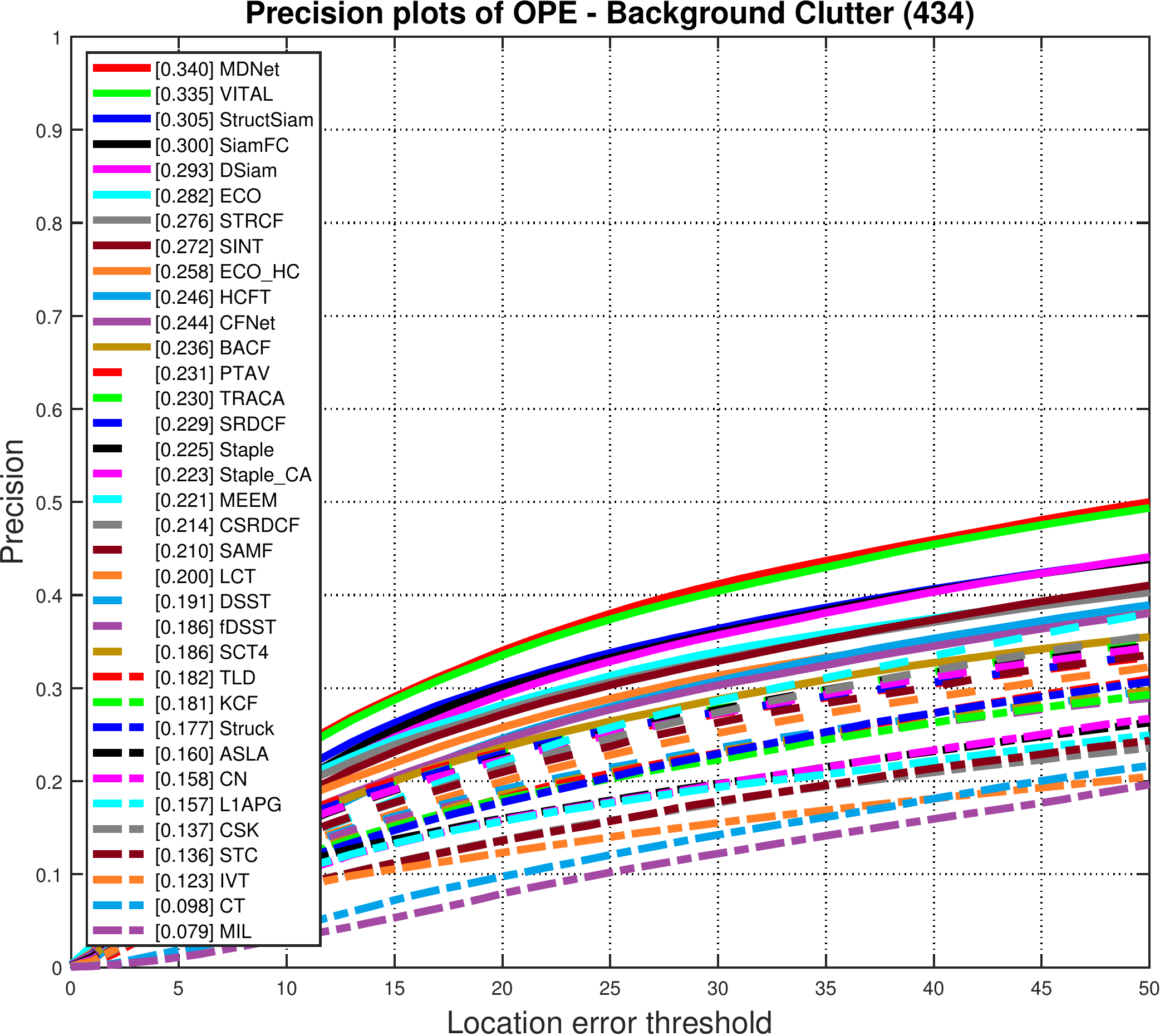}
	\includegraphics[width=4.55cm]{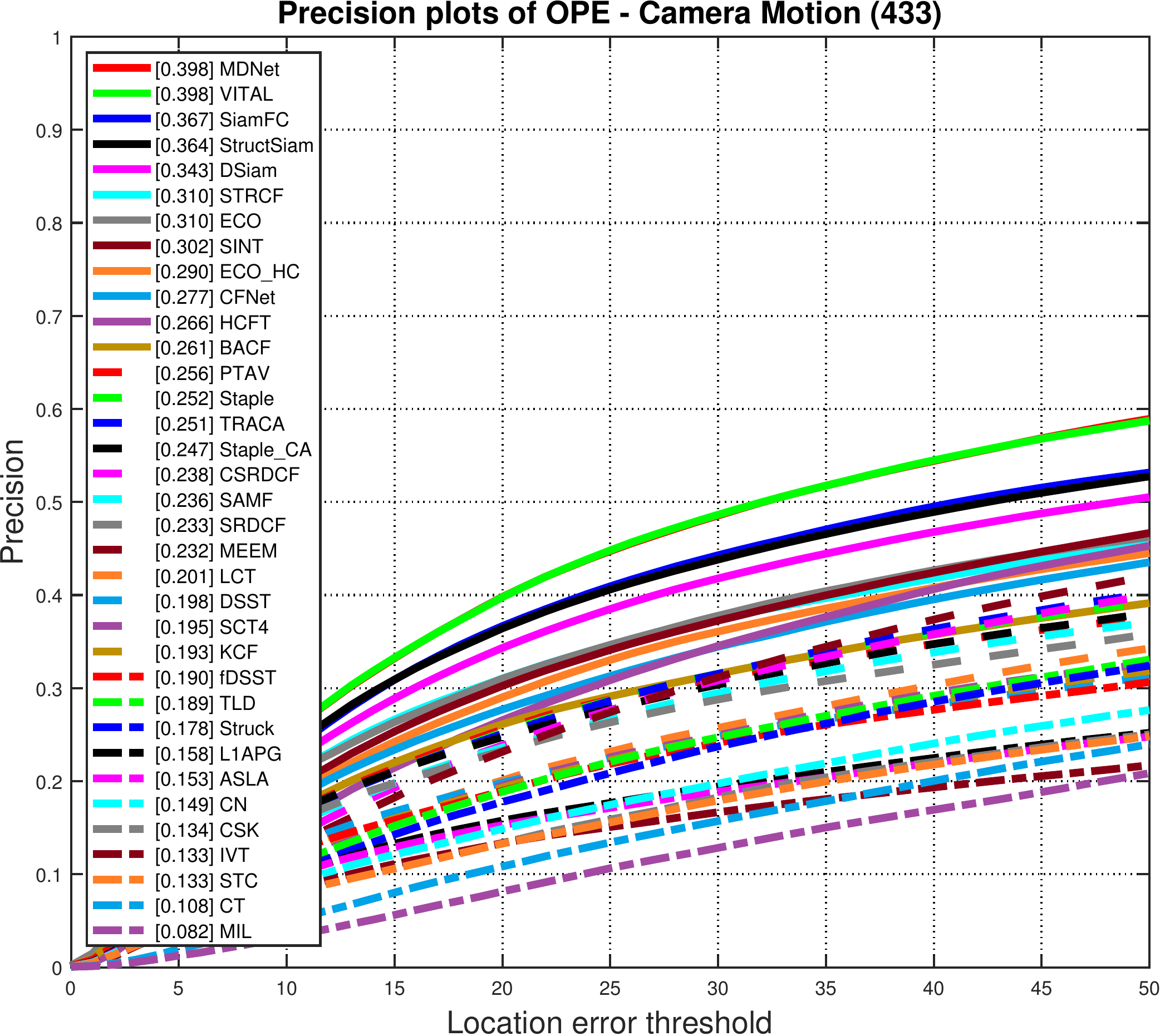}\\
	\includegraphics[width=4.55cm]{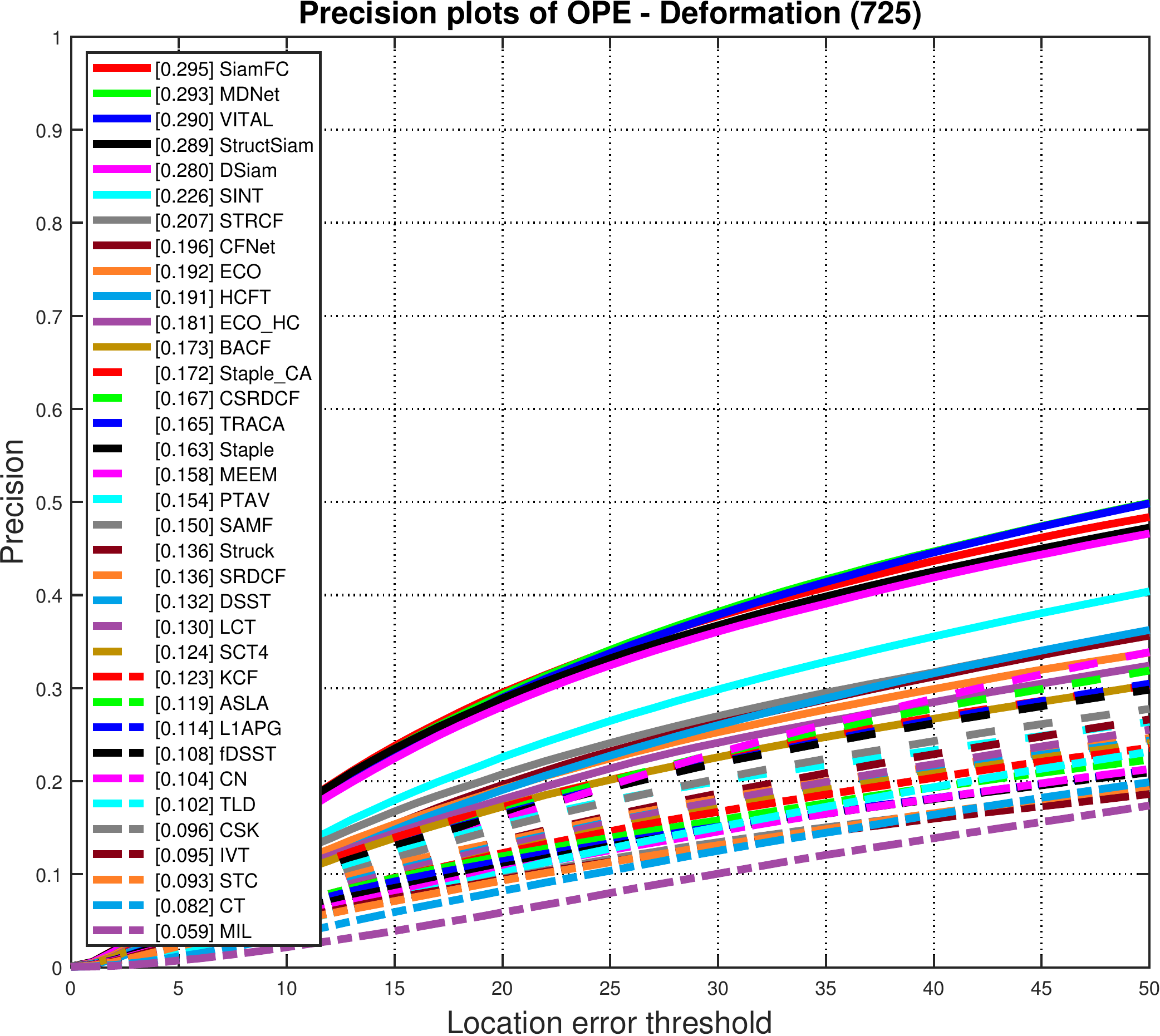}
	\includegraphics[width=4.55cm]{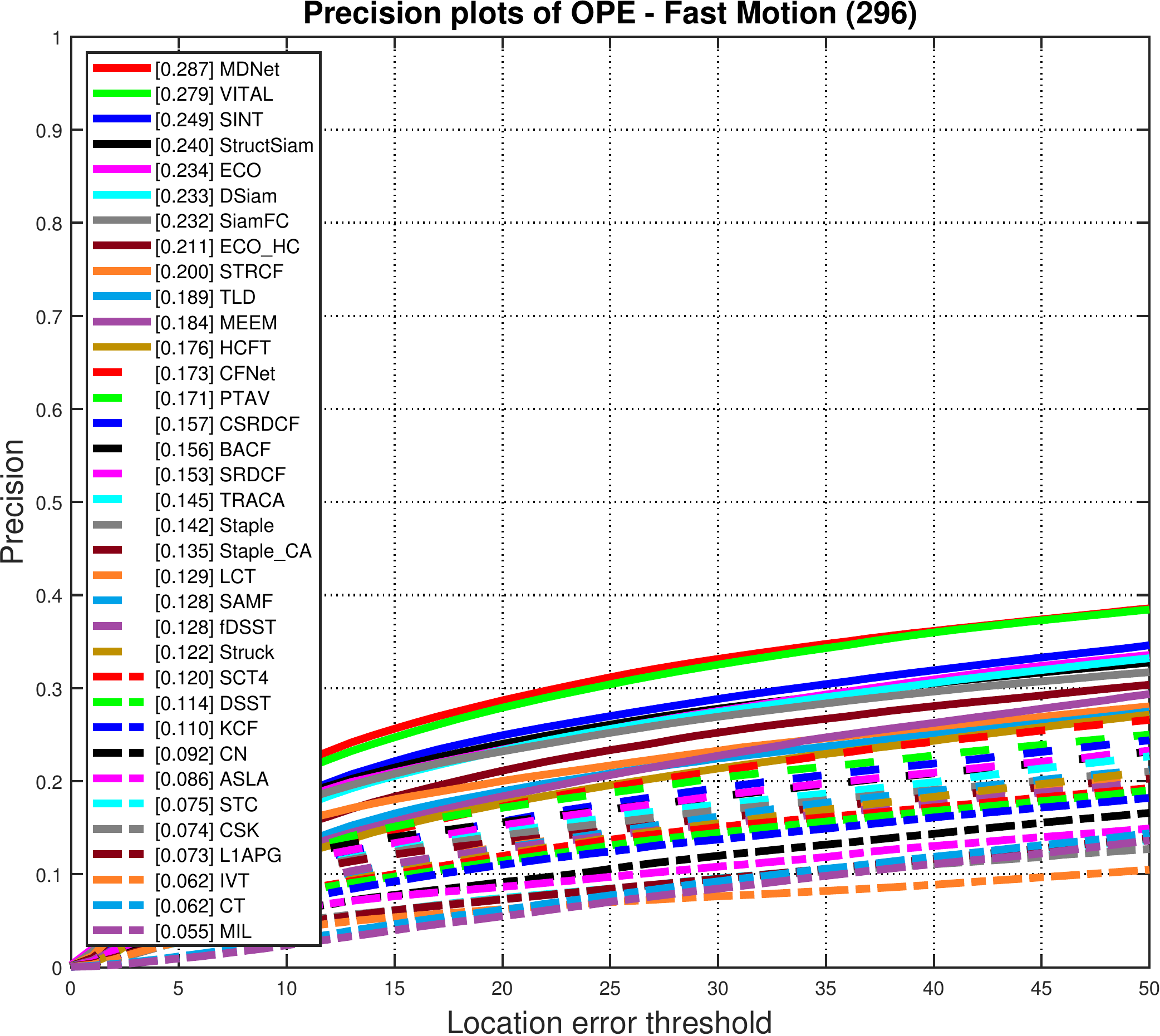}
	\includegraphics[width=4.55cm]{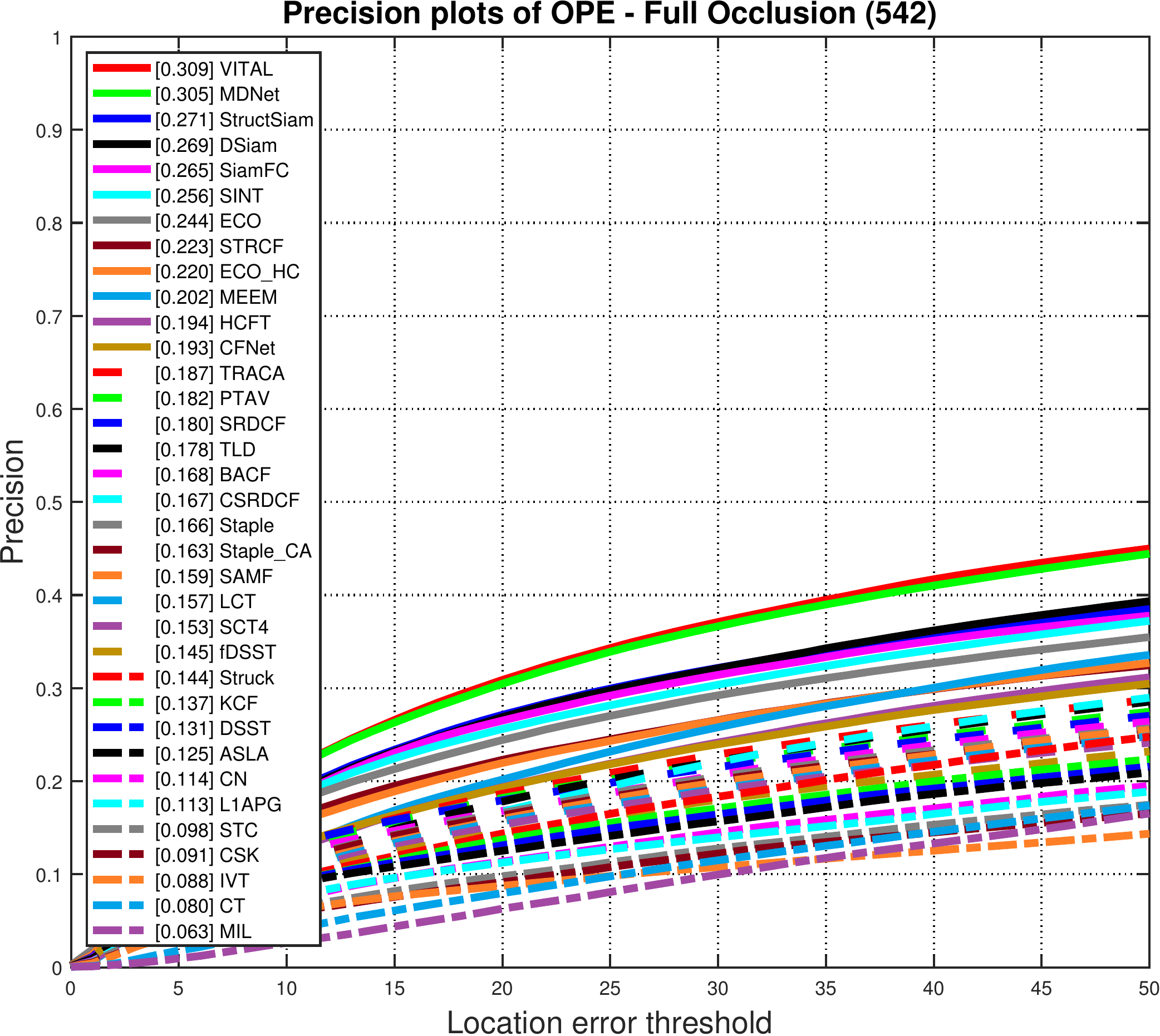}\\
	\includegraphics[width=4.55cm]{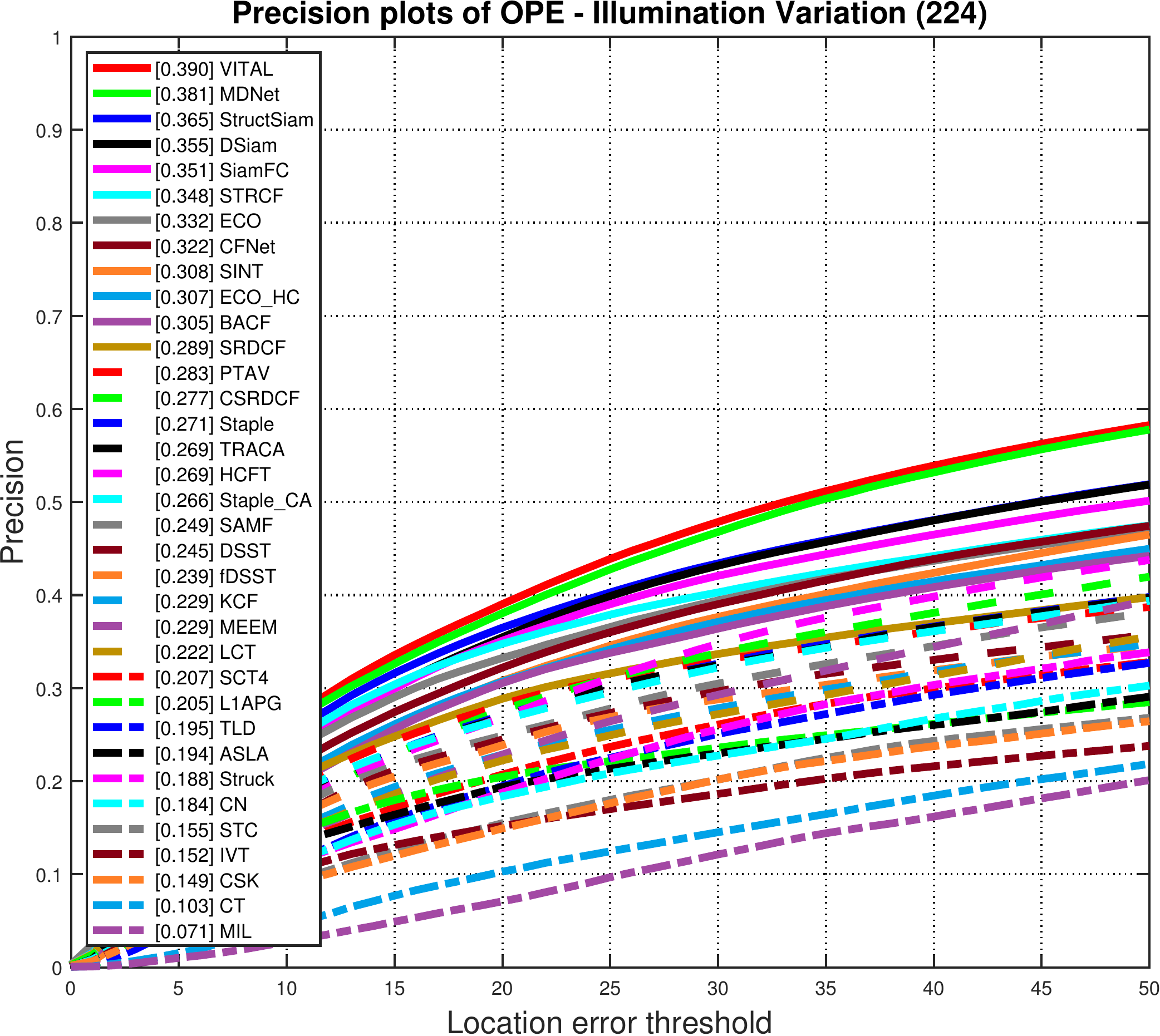}
	\includegraphics[width=4.55cm]{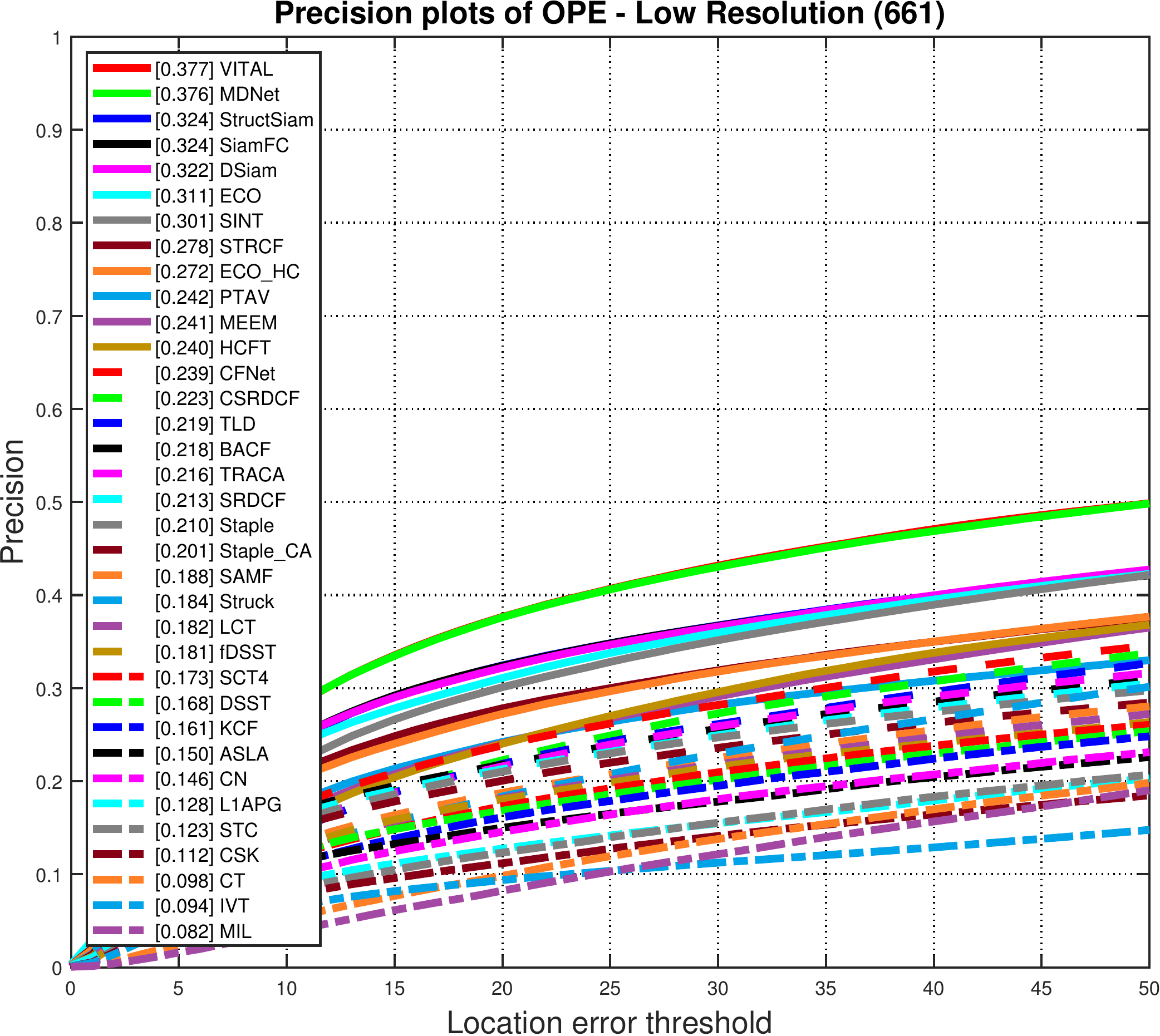}
	\includegraphics[width=4.55cm]{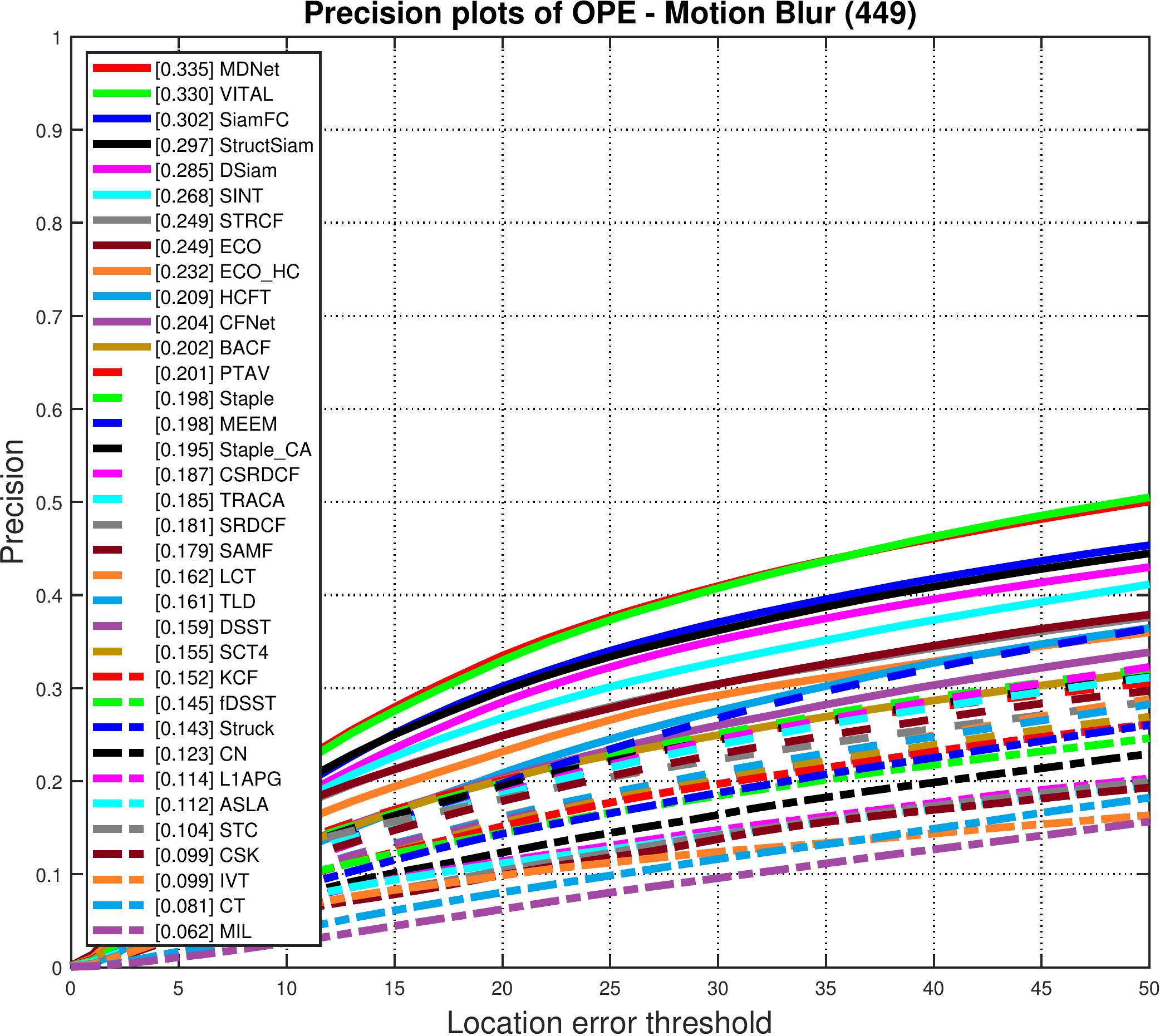}\\
	\includegraphics[width=4.55cm]{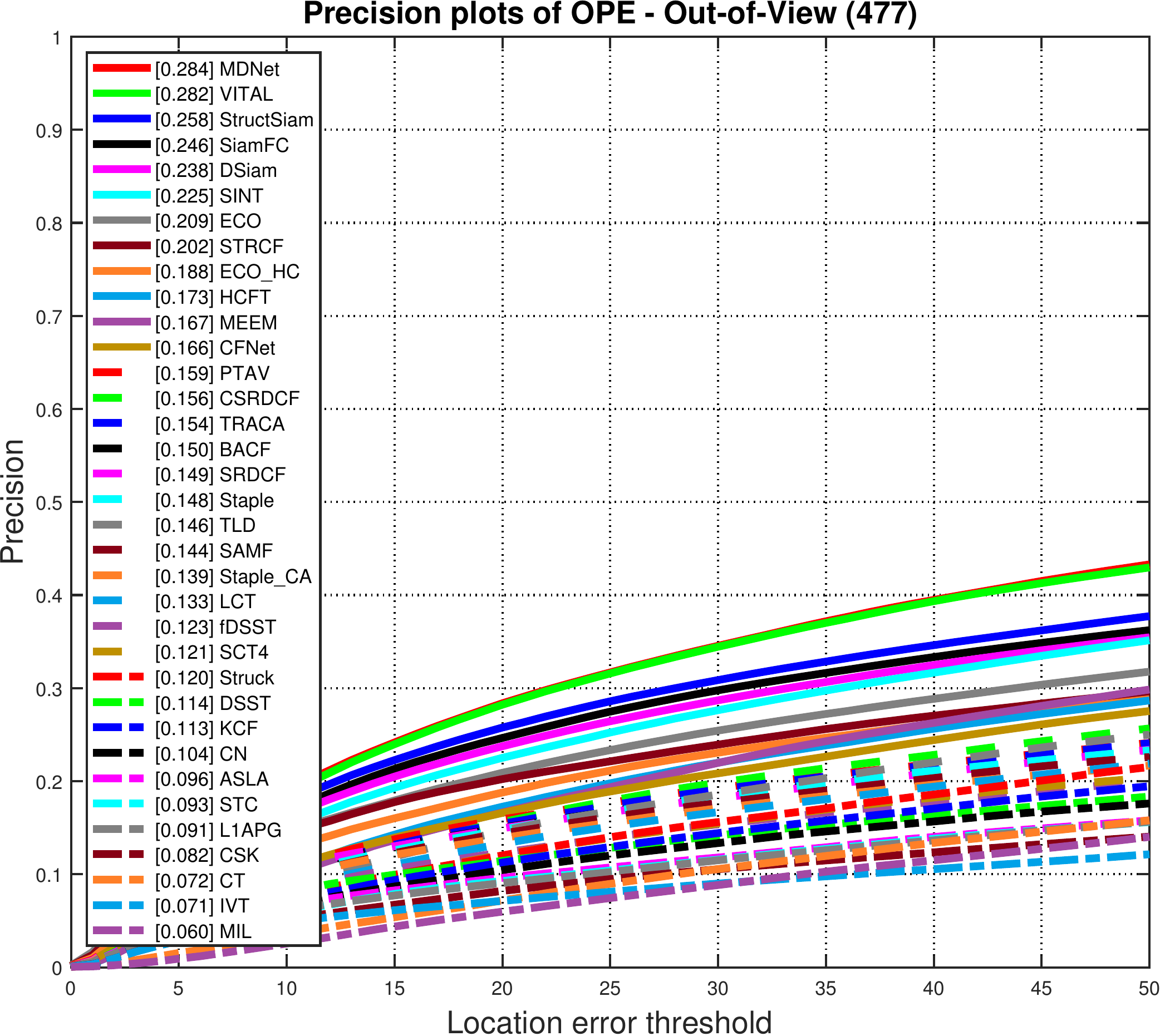}
	\includegraphics[width=4.55cm]{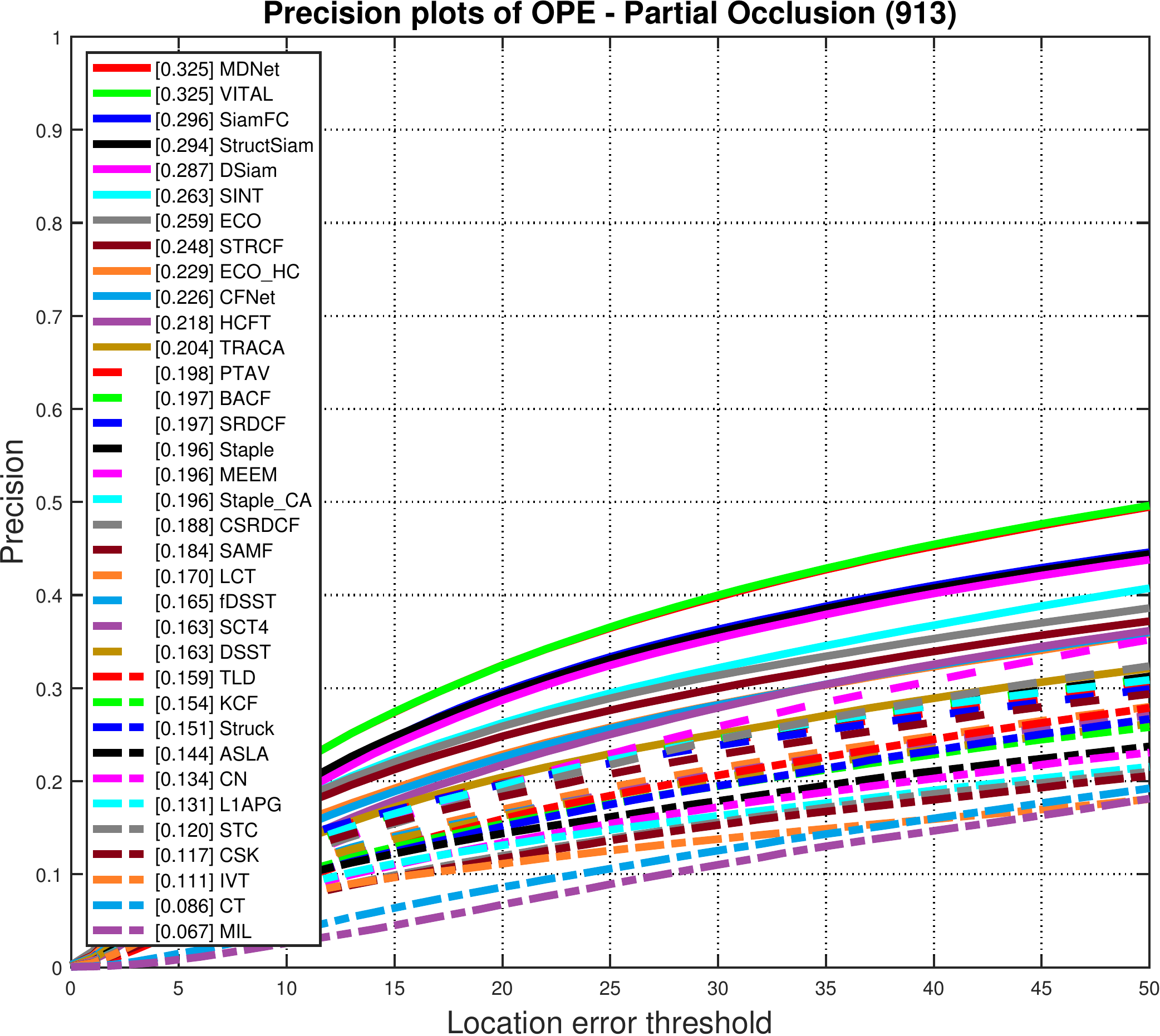}
	\includegraphics[width=4.55cm]{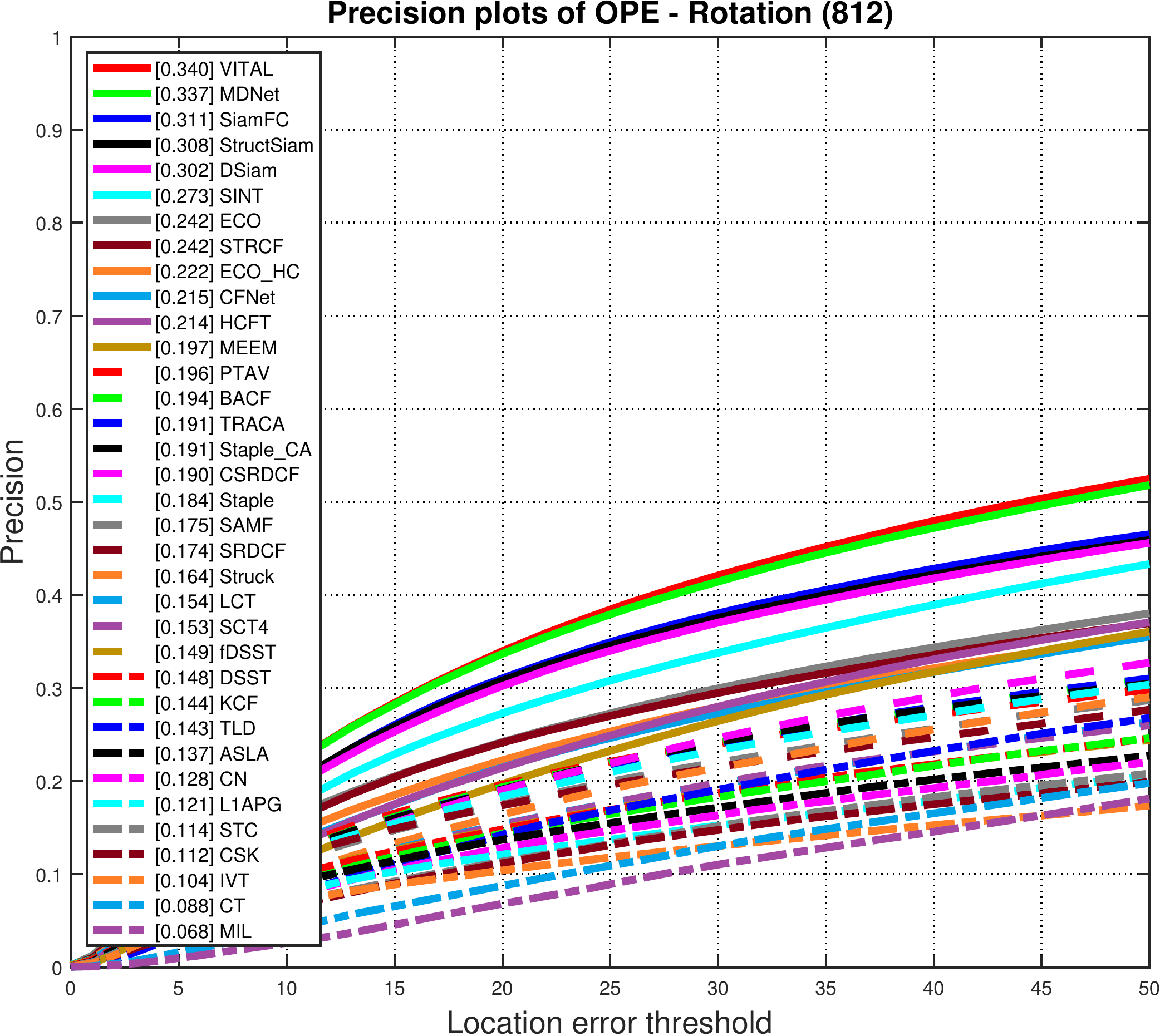}\\
	\includegraphics[width=4.55cm]{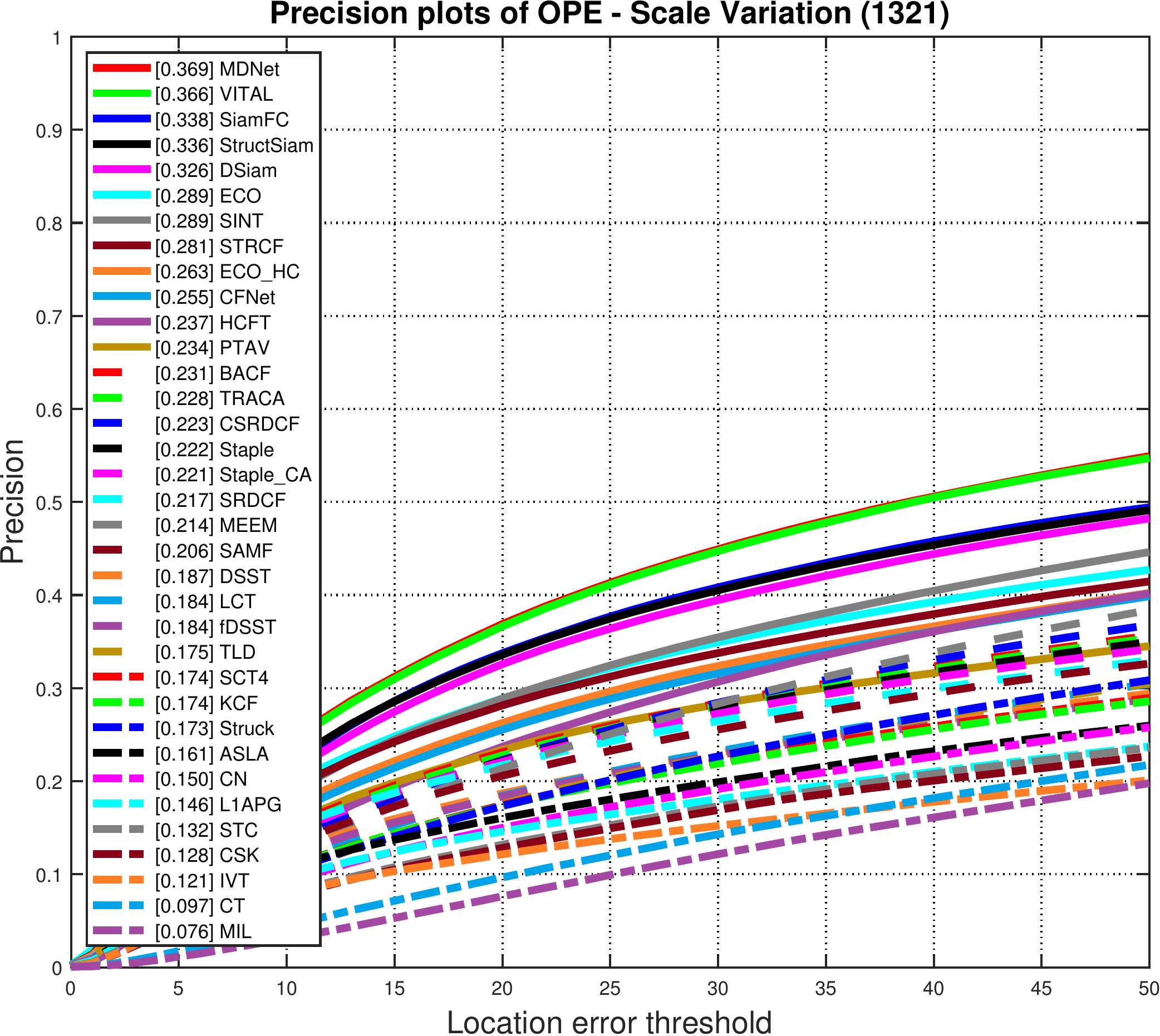}
	\includegraphics[width=4.55cm]{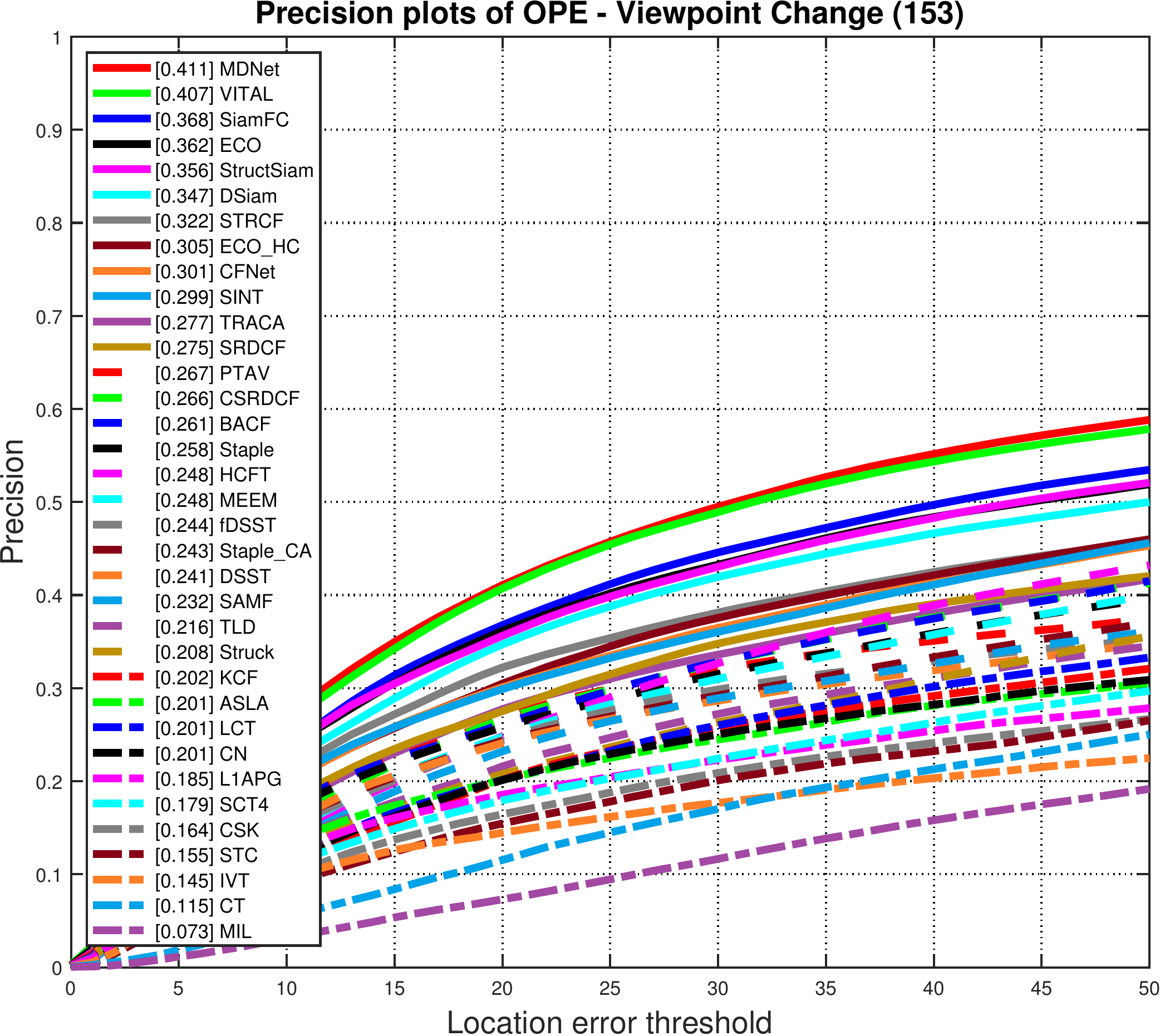}\\
	\caption{Performance of trackers on each attribute using precision under protocol \uppercase\expandafter{\romannumeral1}. Best viewed in color.}
	\label{fig:protocol_1_all_att_res_precision}
\end{figure*}

\newpage
Fig.~\ref{fig:protocol_1_all_att_res_norm_precision} shows the performance of trackers on each attribute using normalized precision under protocol \uppercase\expandafter{\romannumeral1}.
\begin{figure*}[!hbpt]
	\centering
	\includegraphics[width=4.55cm]{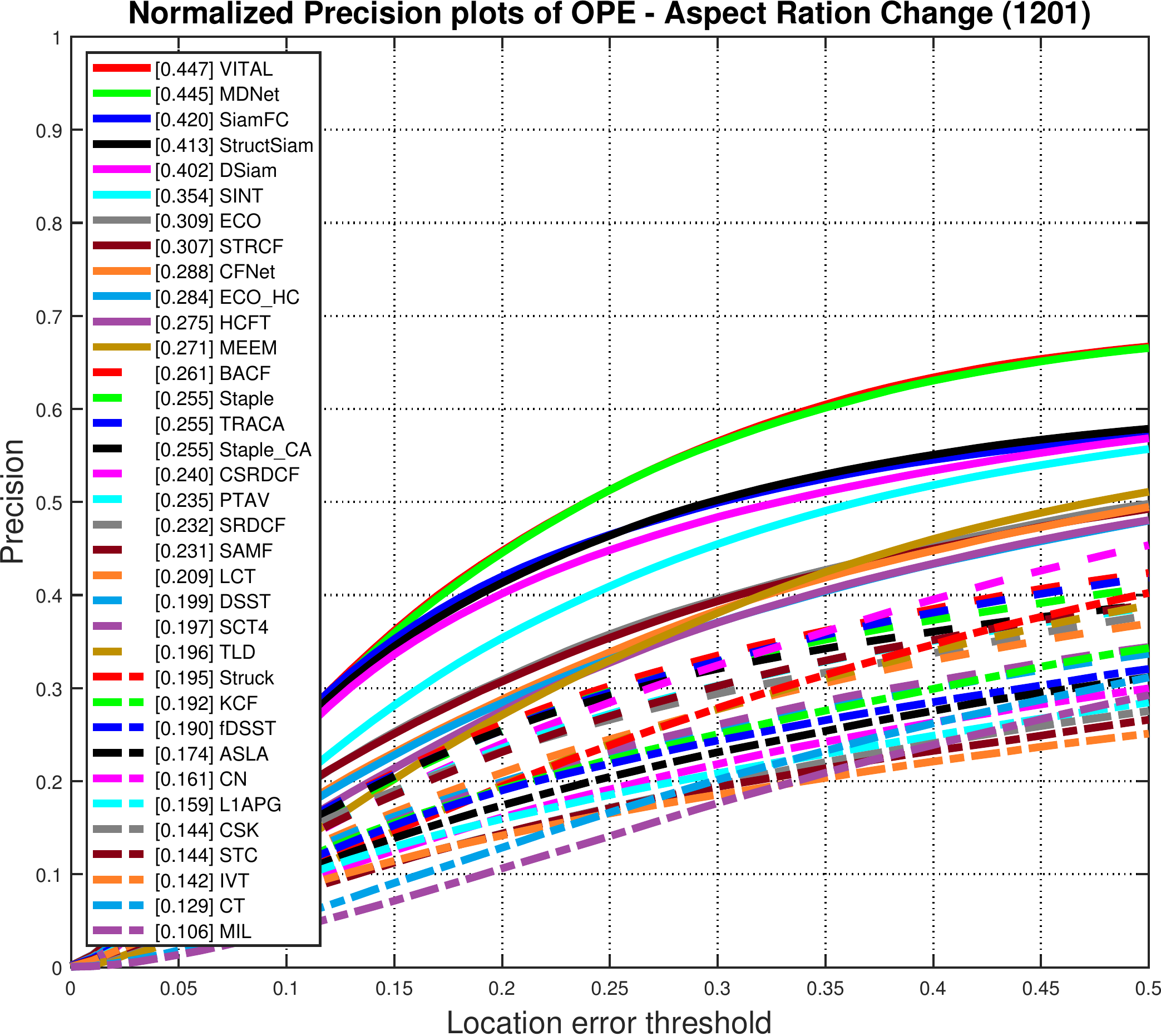}
	\includegraphics[width=4.55cm]{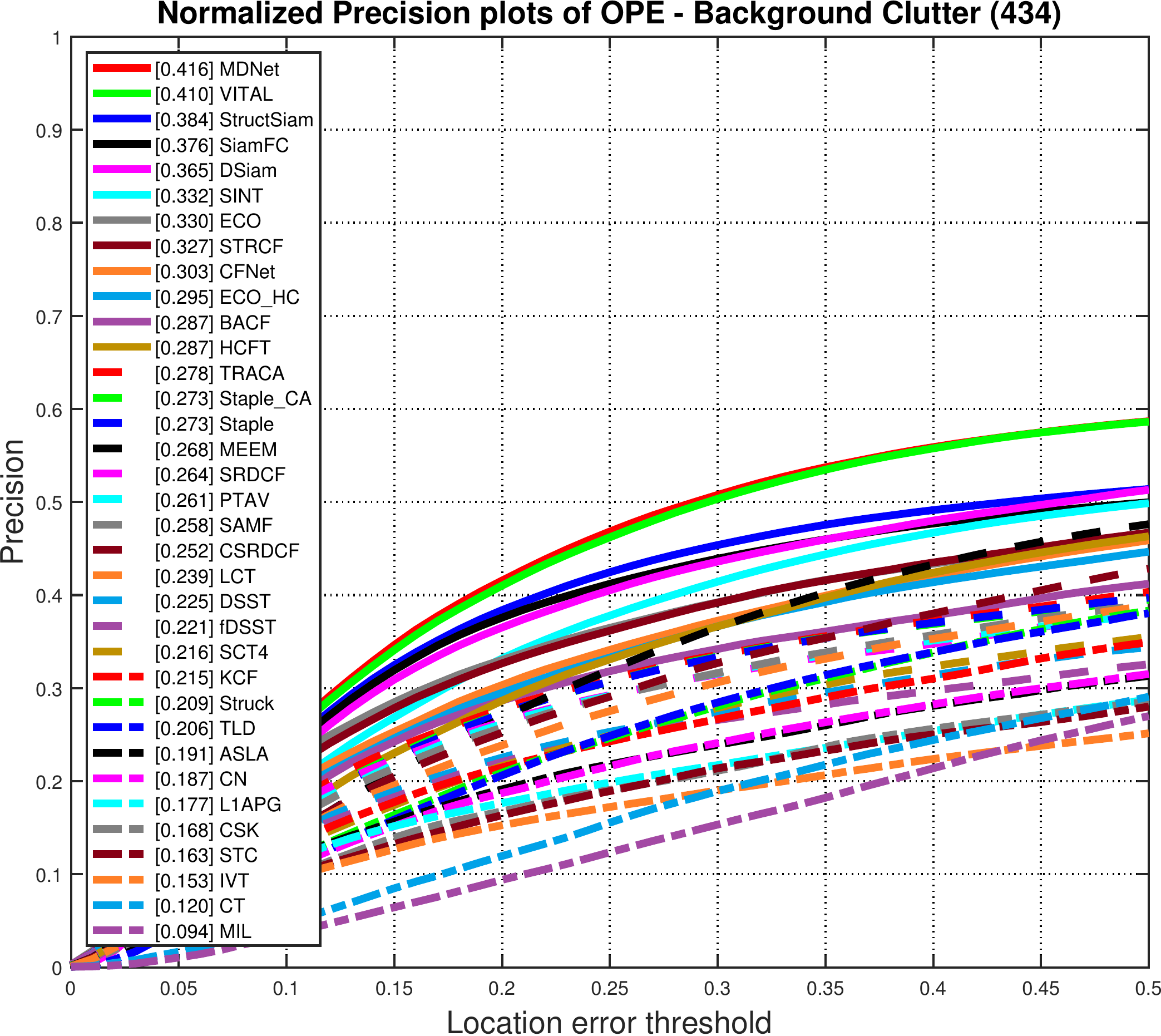}
	\includegraphics[width=4.55cm]{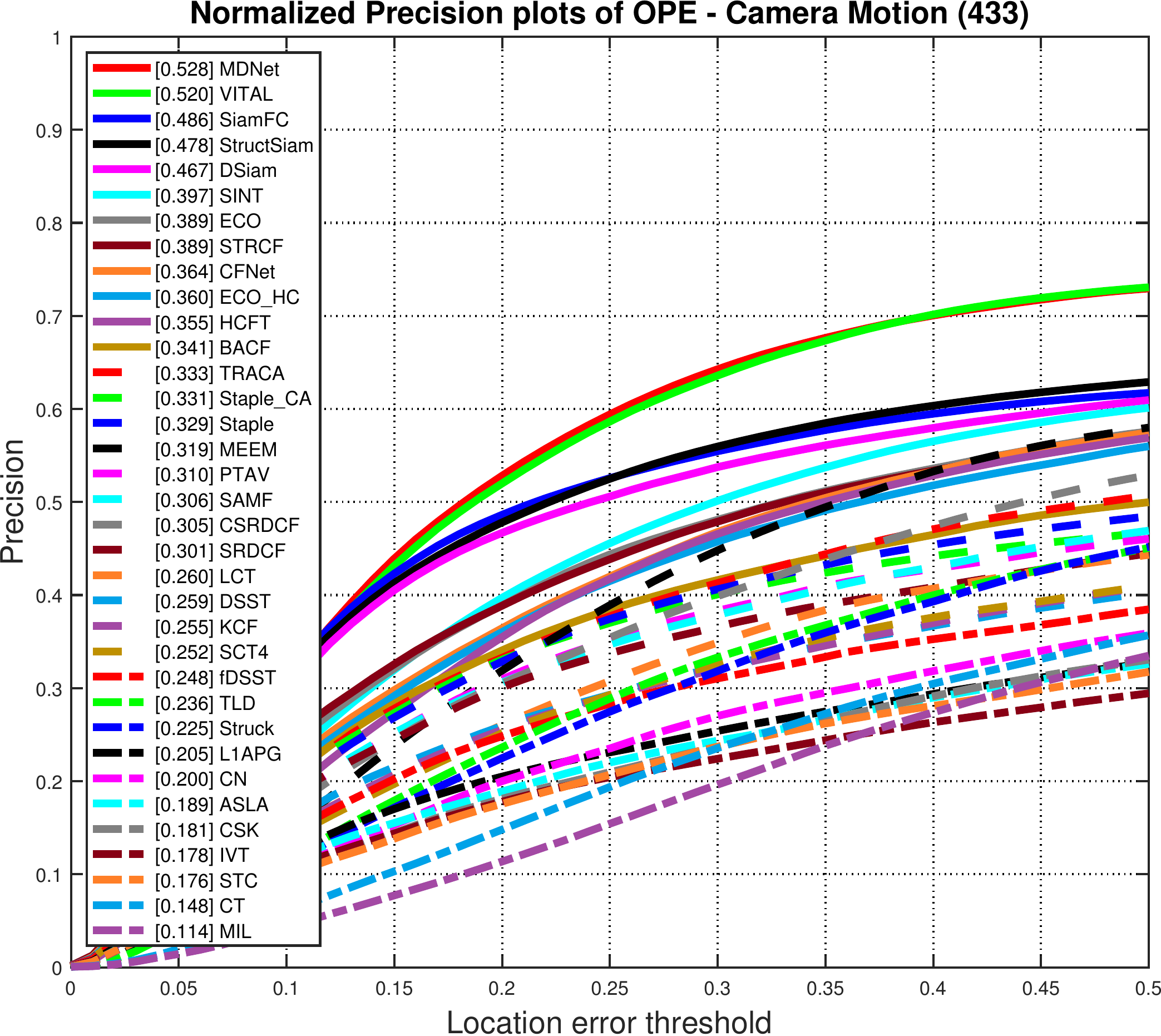}\\
	\includegraphics[width=4.55cm]{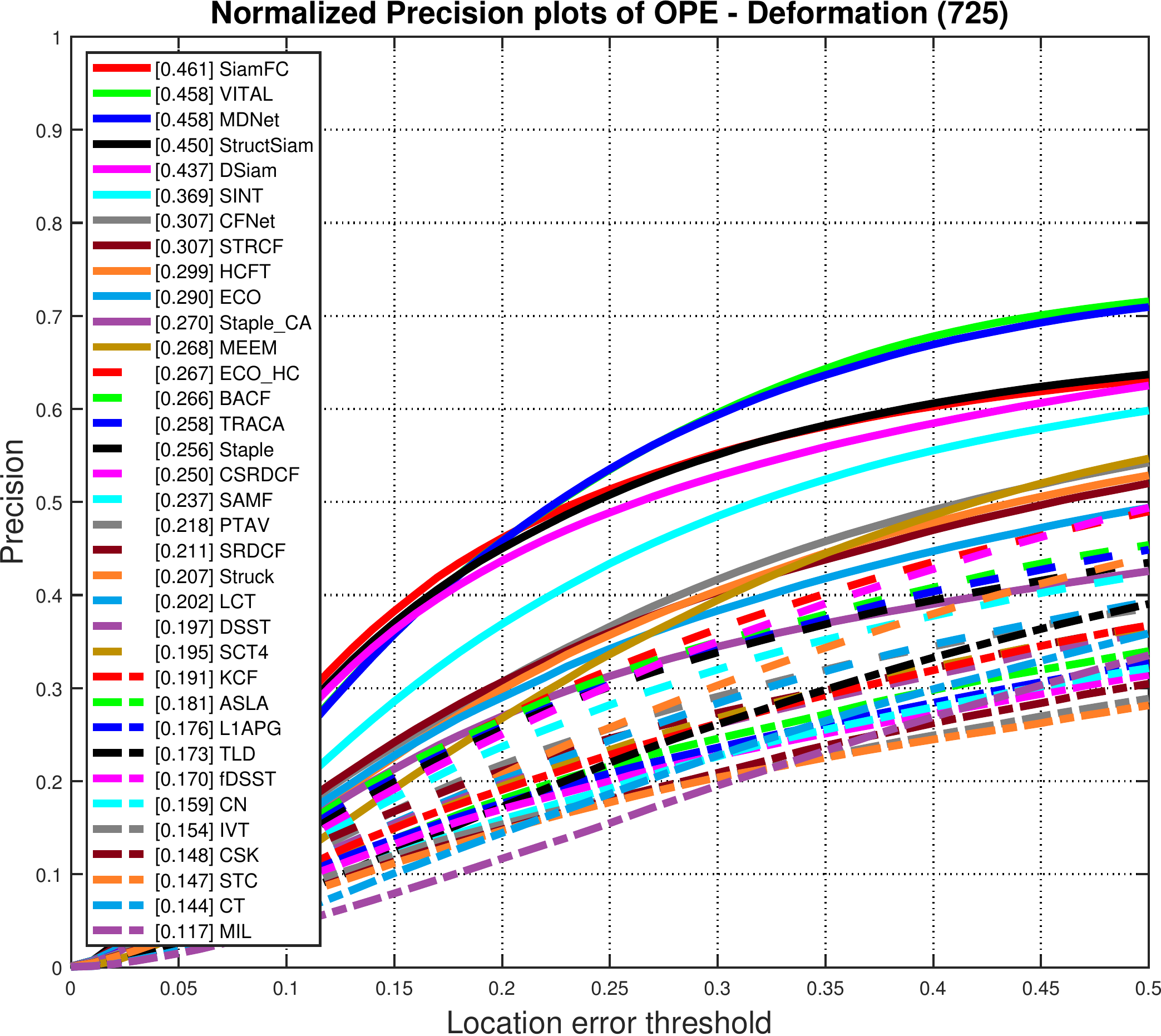}
	\includegraphics[width=4.55cm]{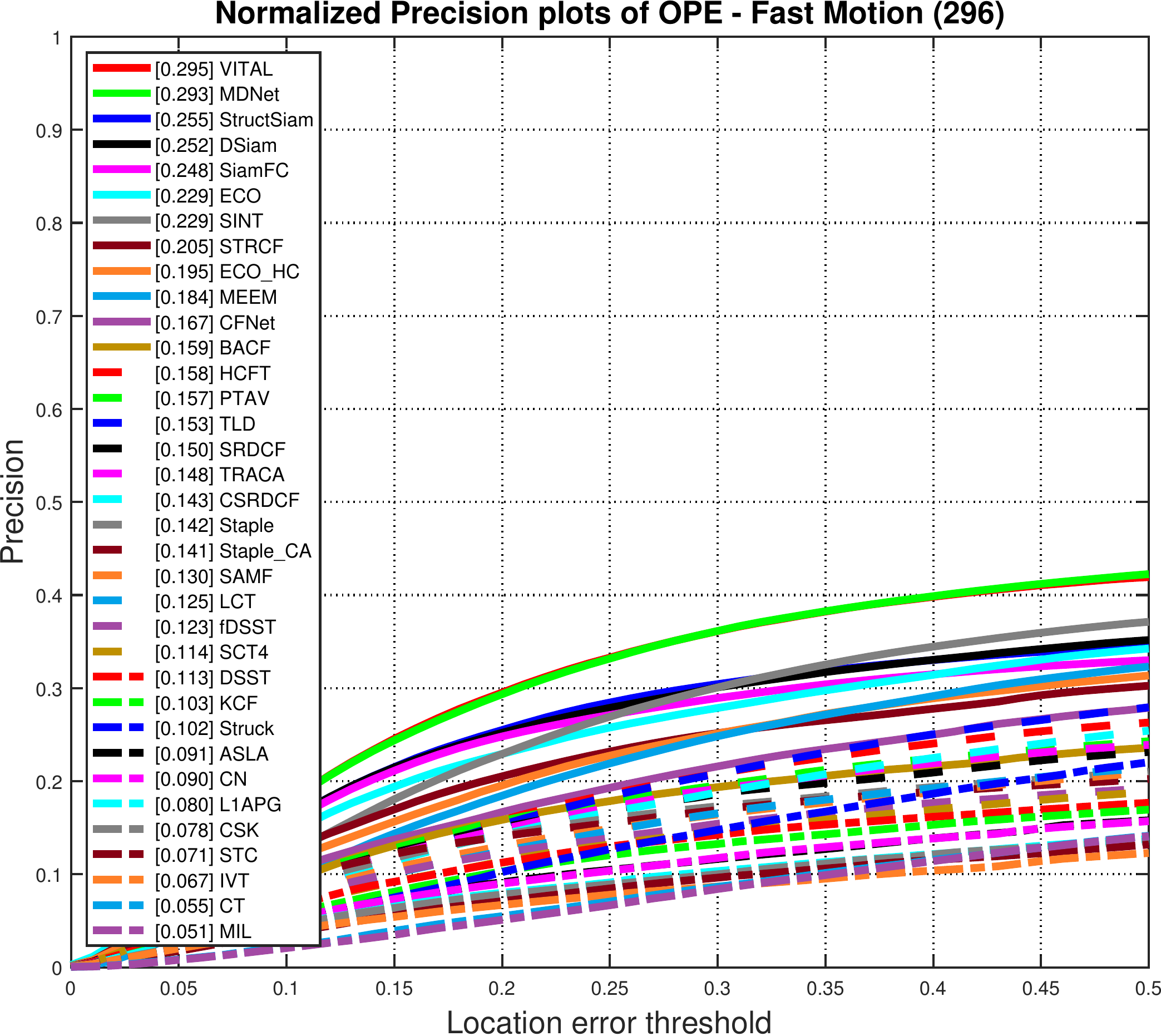}
	\includegraphics[width=4.55cm]{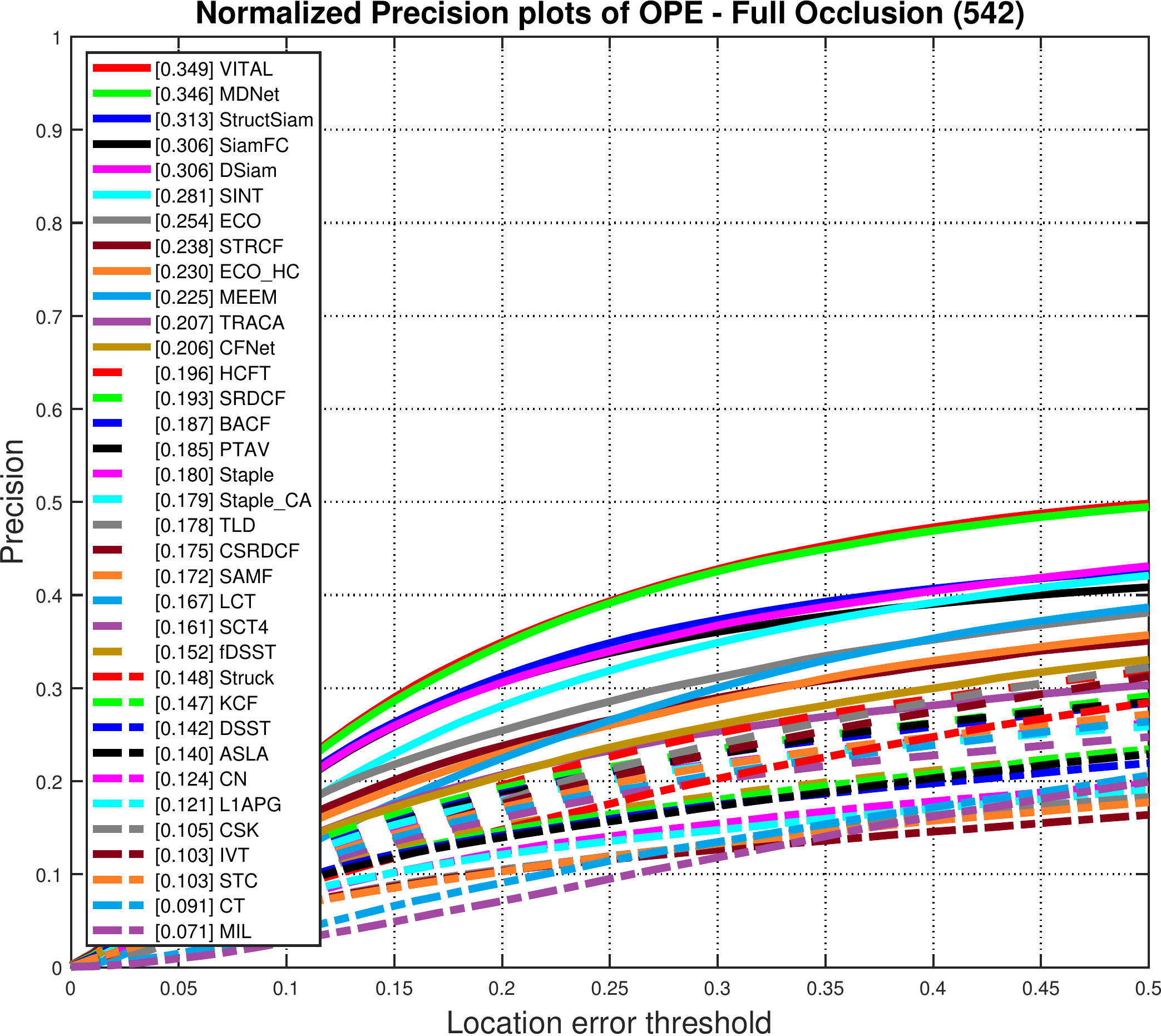}\\
	\includegraphics[width=4.55cm]{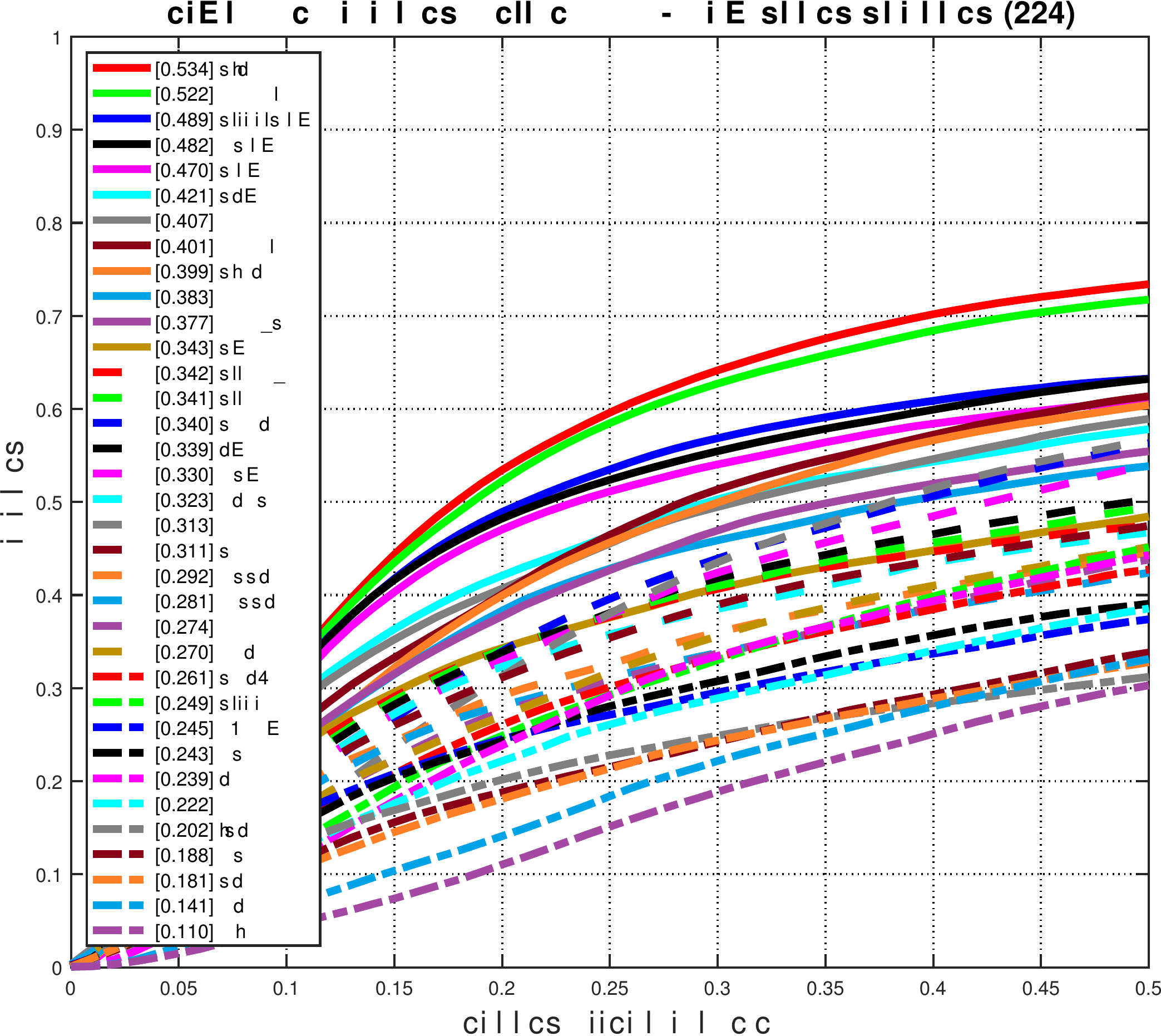}
	\includegraphics[width=4.55cm]{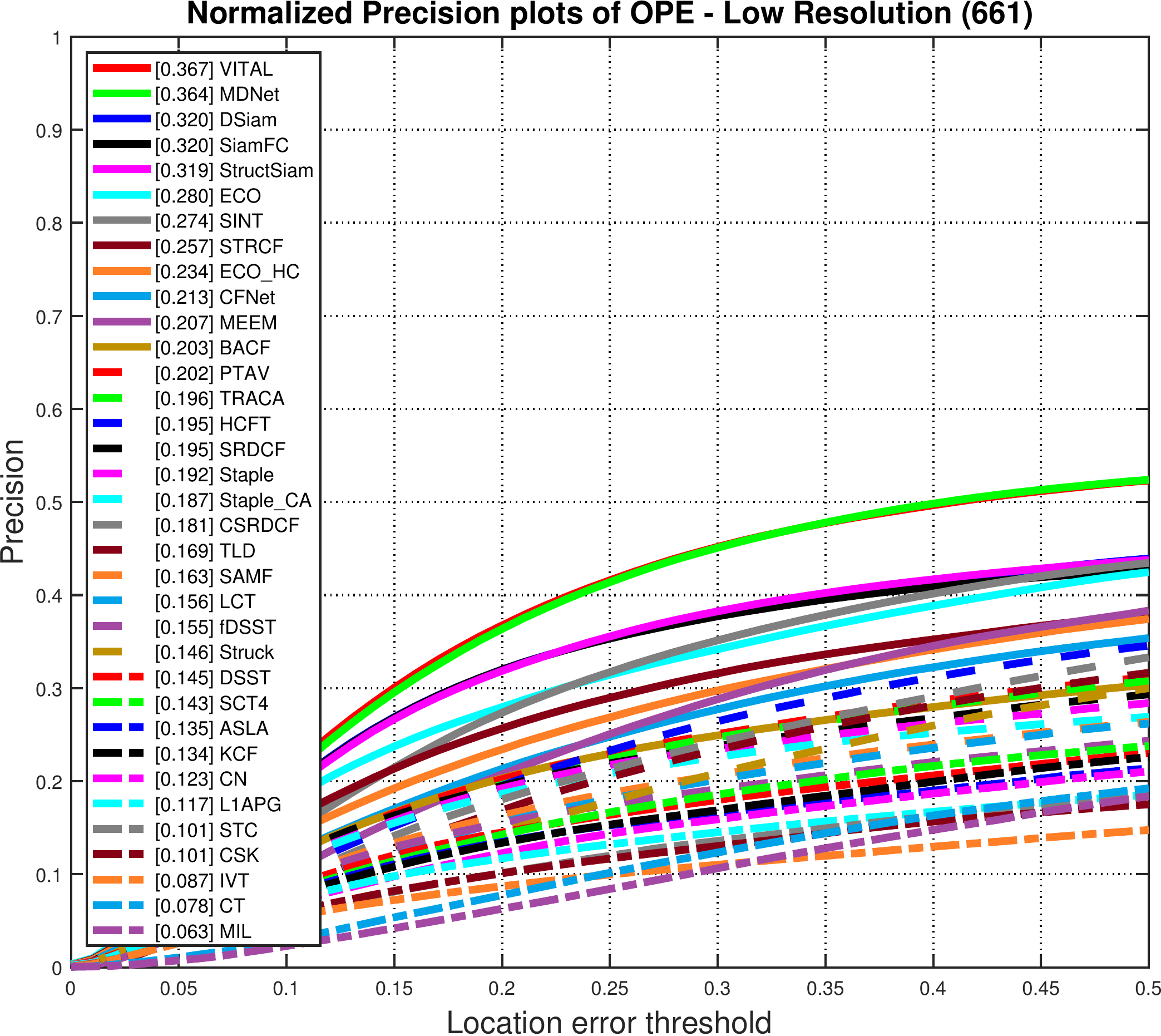}
	\includegraphics[width=4.55cm]{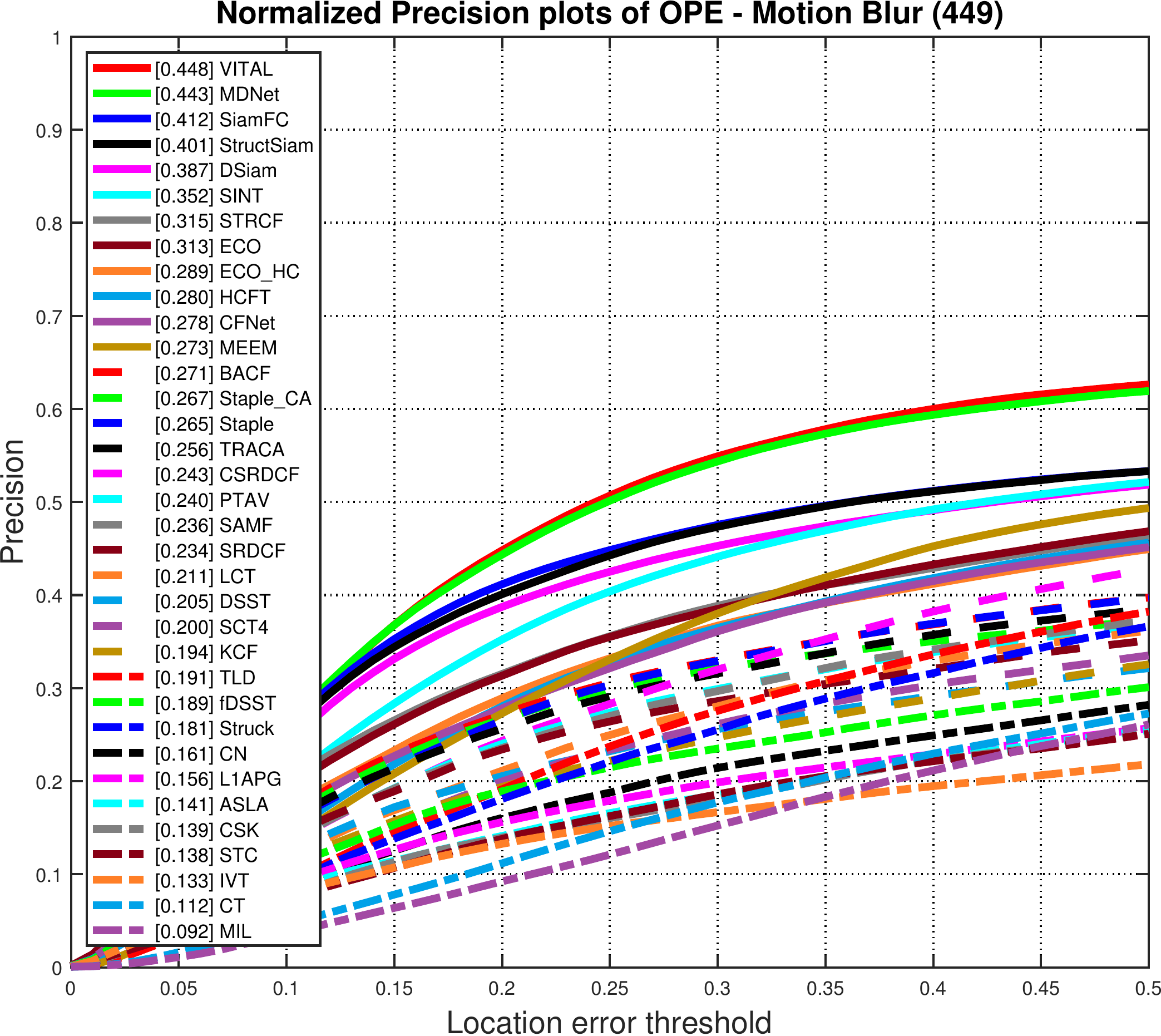}\\
	\includegraphics[width=4.55cm]{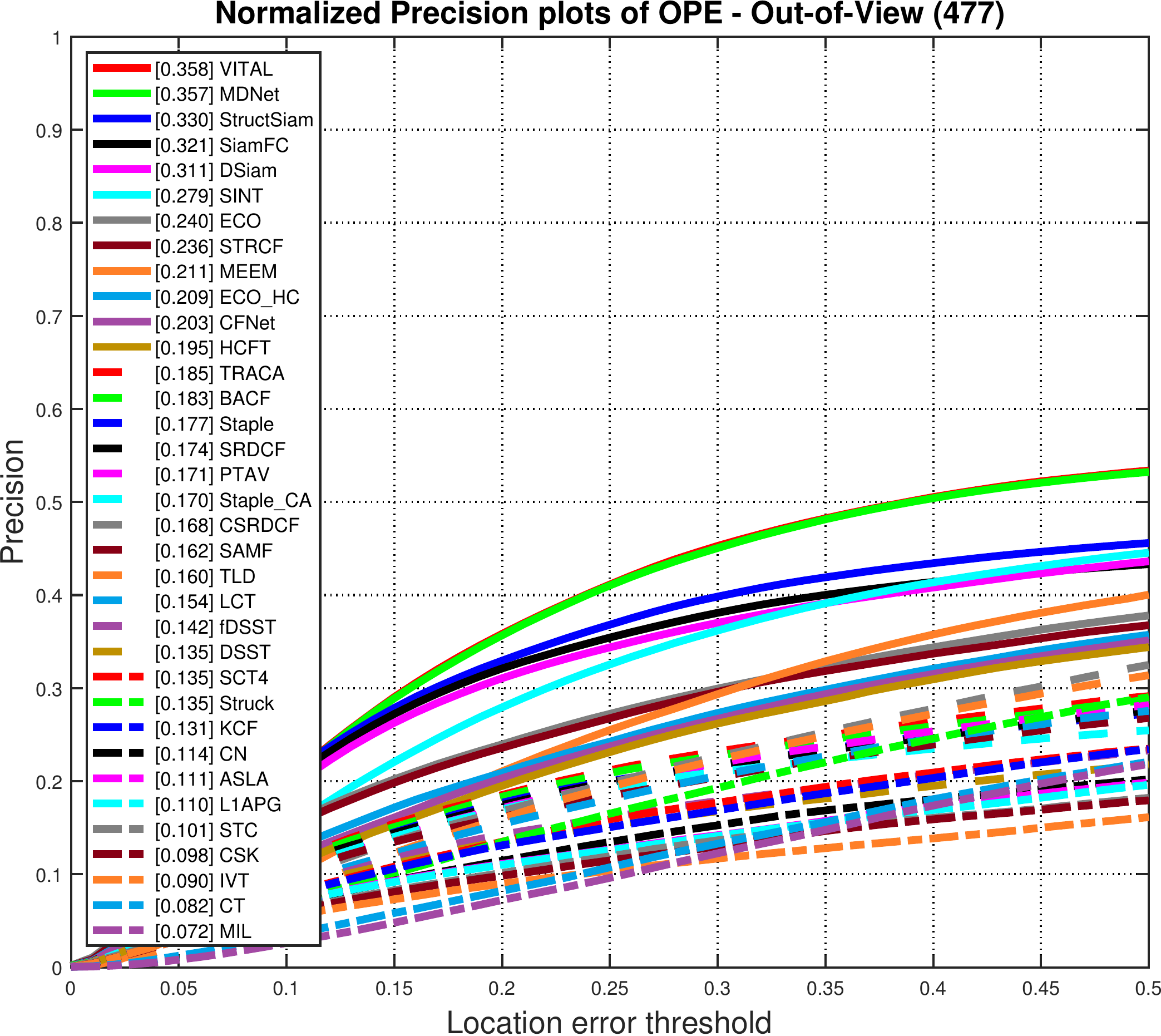}
	\includegraphics[width=4.55cm]{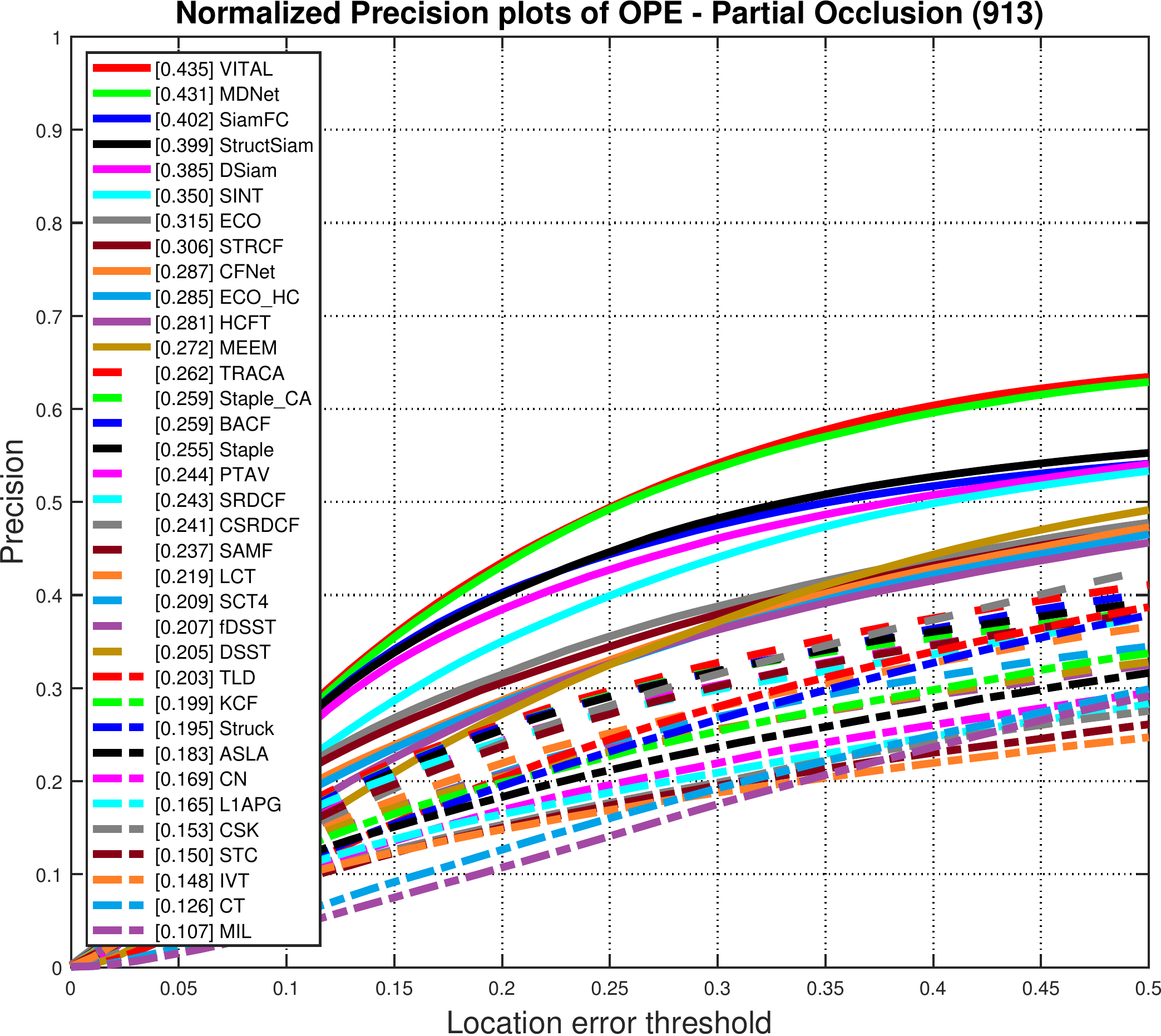}
	\includegraphics[width=4.55cm]{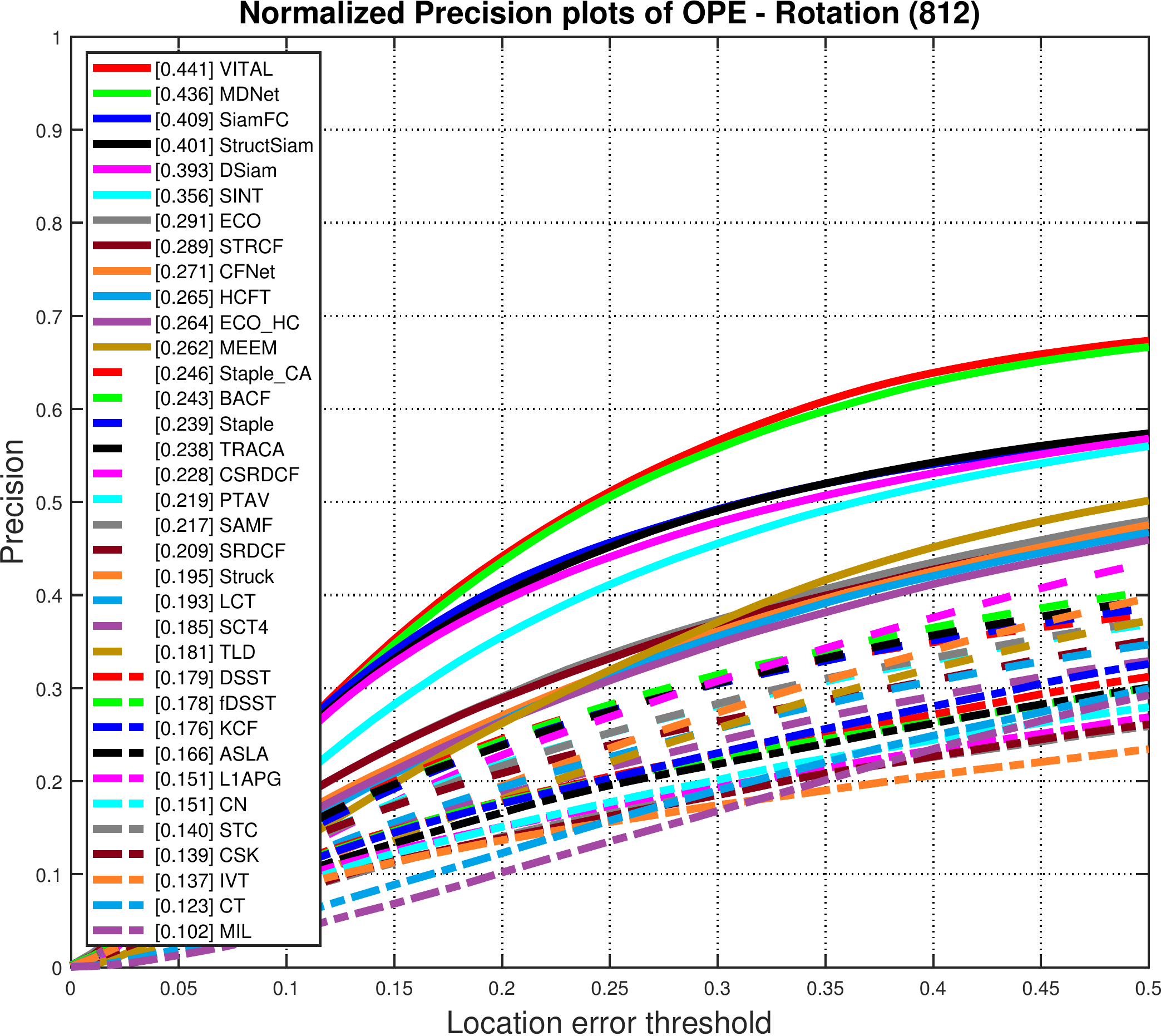}\\
	\includegraphics[width=4.55cm]{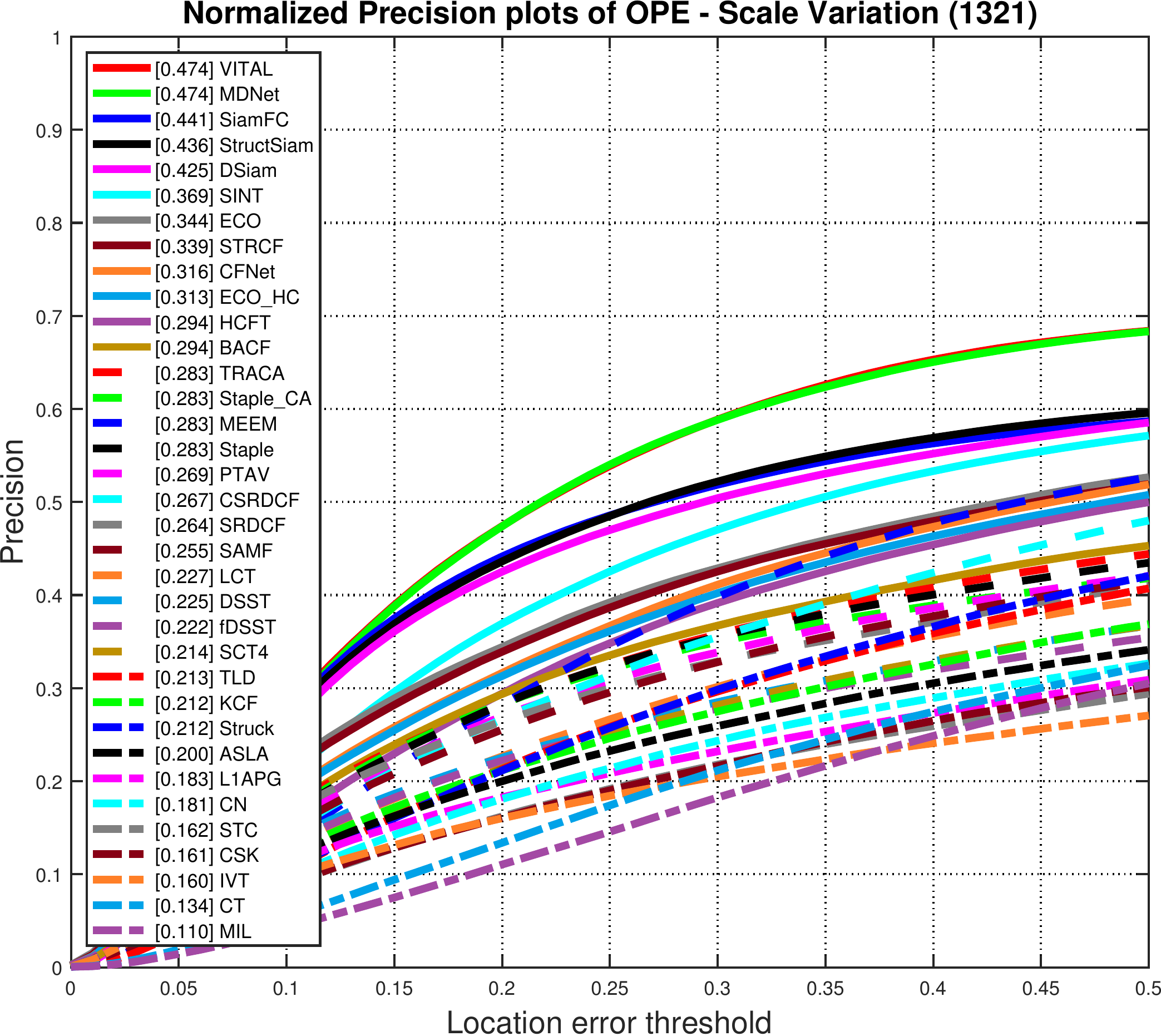}
	\includegraphics[width=4.55cm]{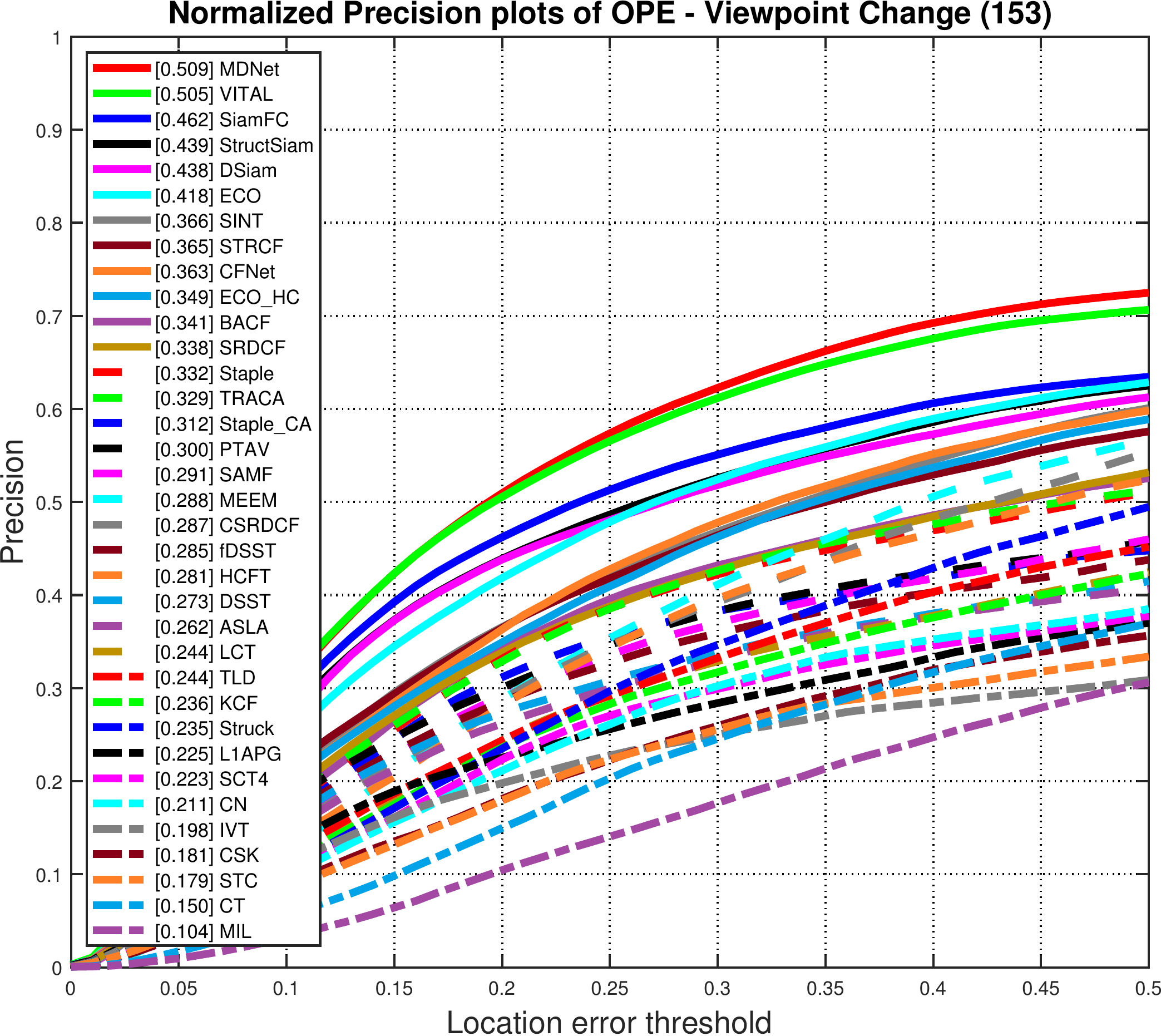}\\
	\caption{Performance of trackers on each attribute using precision under protocol \uppercase\expandafter{\romannumeral1}. Best viewed in color.}
	\label{fig:protocol_1_all_att_res_norm_precision}
\end{figure*}

\newpage
Fig.~\ref{fig:protocol_1_all_att_res_success} shows the performance of trackers on each attribute using success under protocol \uppercase\expandafter{\romannumeral1}.
\begin{figure*}[!hbpt]
	\centering
	\includegraphics[width=4.55cm]{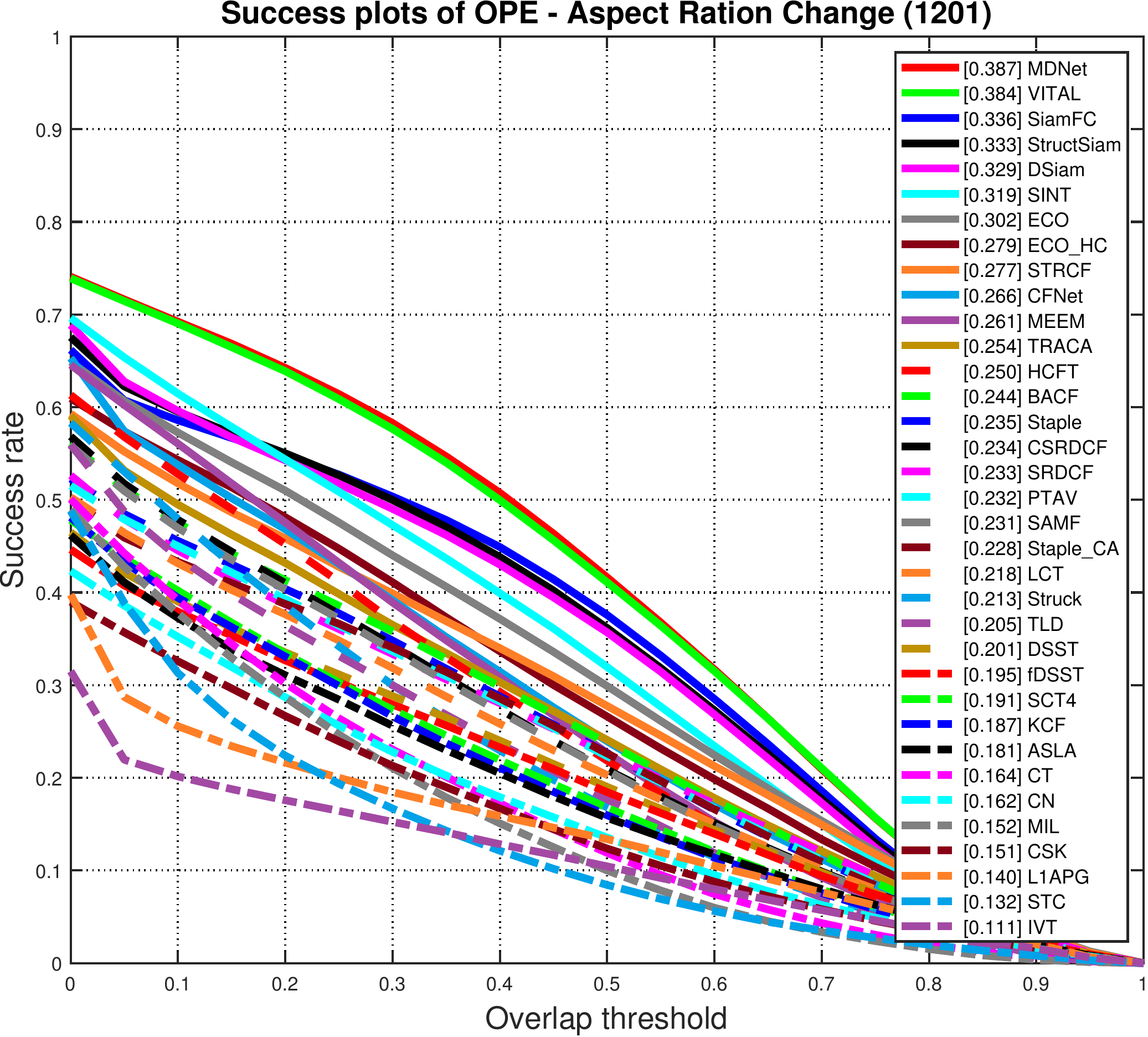}
	\includegraphics[width=4.55cm]{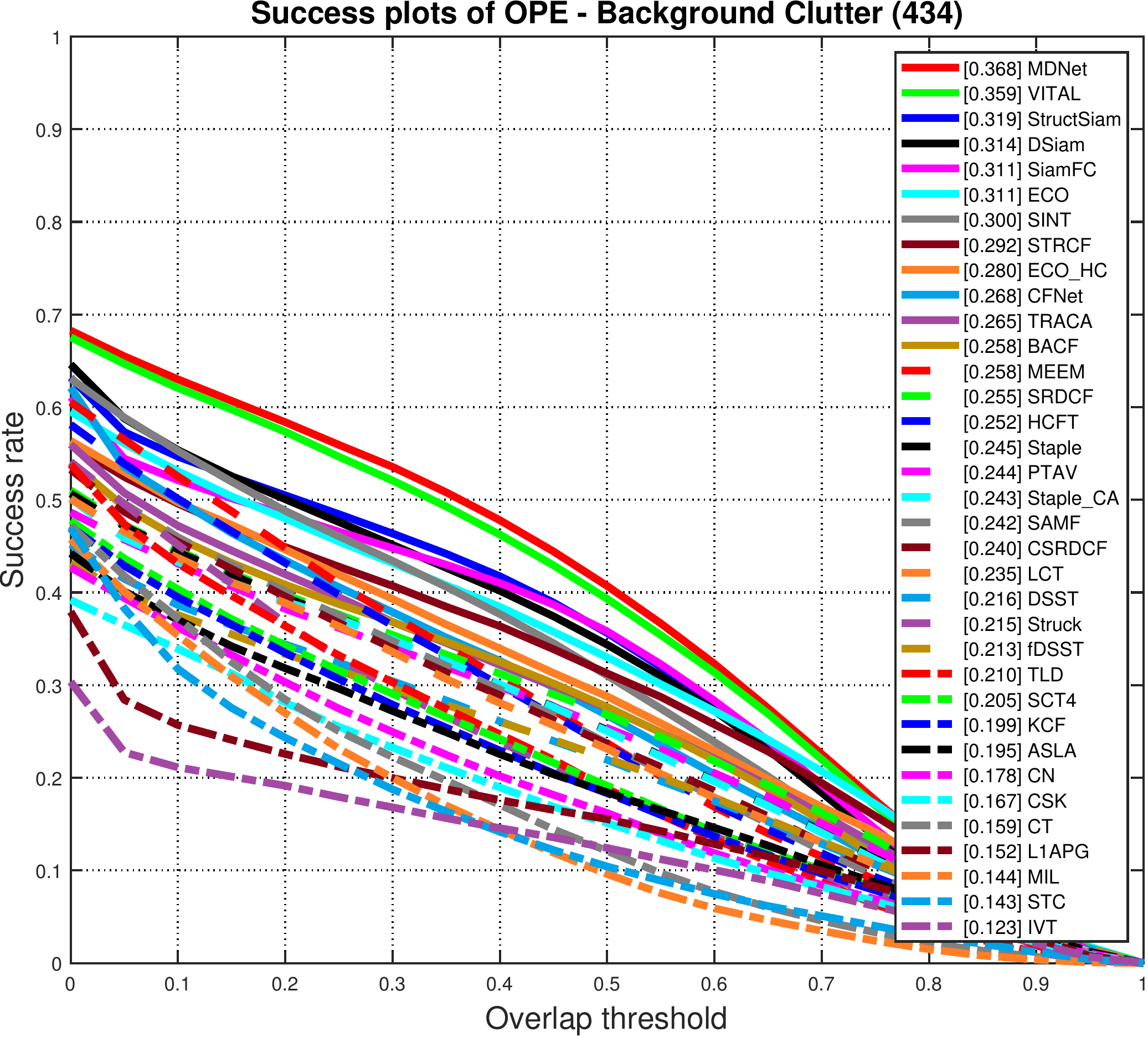}
	\includegraphics[width=4.55cm]{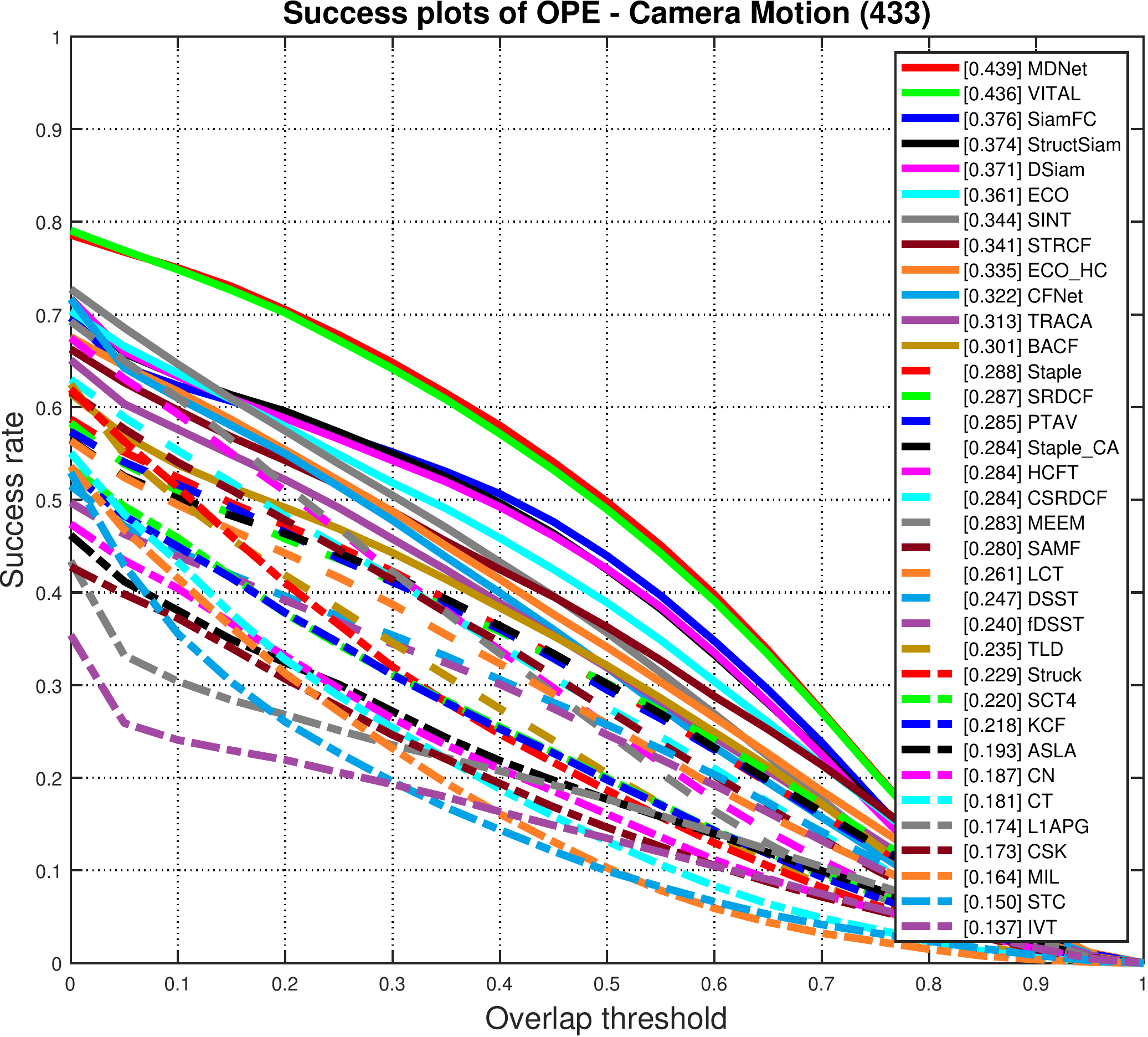}\\
	\includegraphics[width=4.55cm]{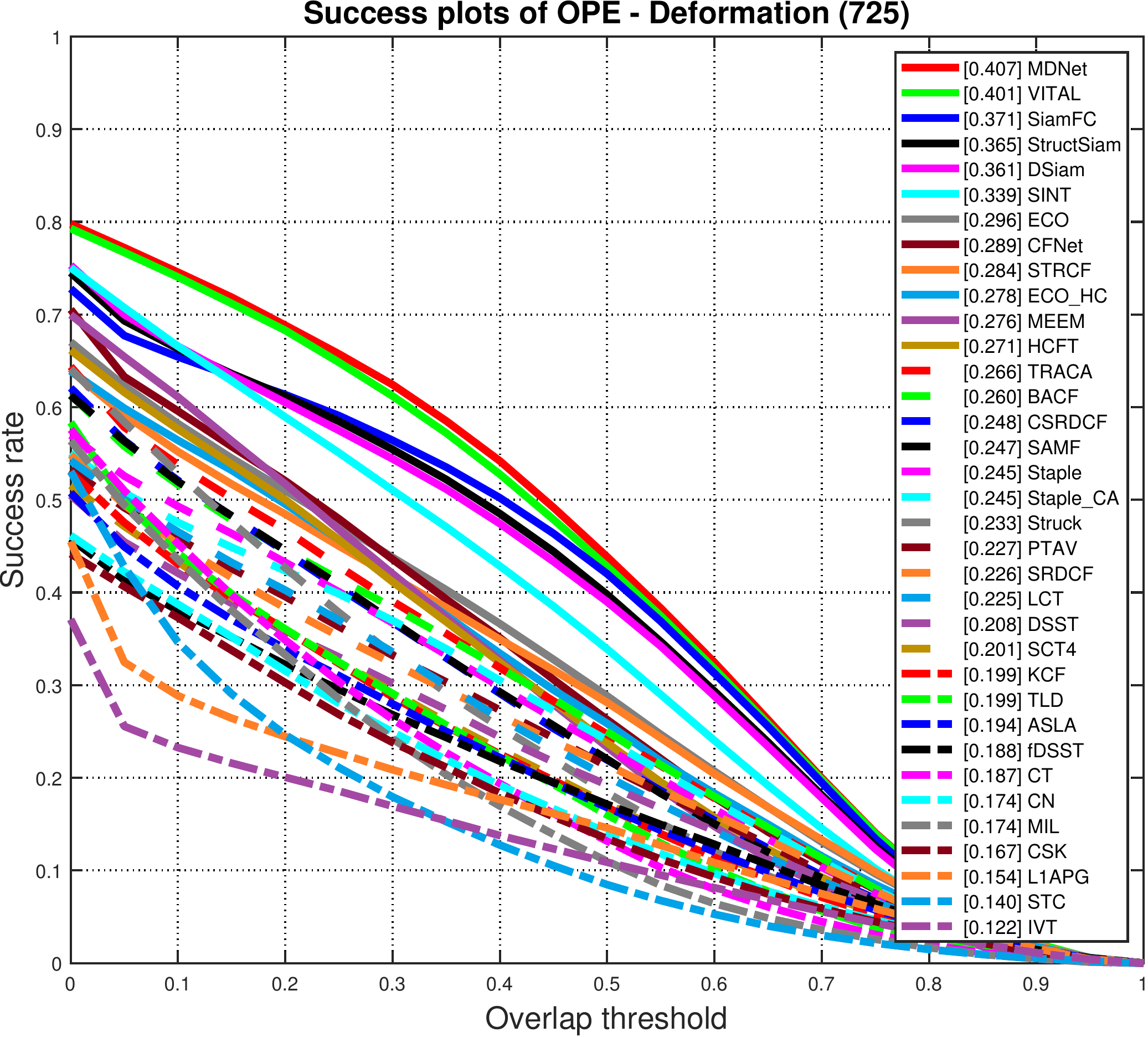}
	\includegraphics[width=4.55cm]{protocol_1_success_att/FM_overlap_OPE_AUC}
	\includegraphics[width=4.55cm]{protocol_1_success_att/FOC_overlap_OPE_AUC}\\
	\includegraphics[width=4.55cm]{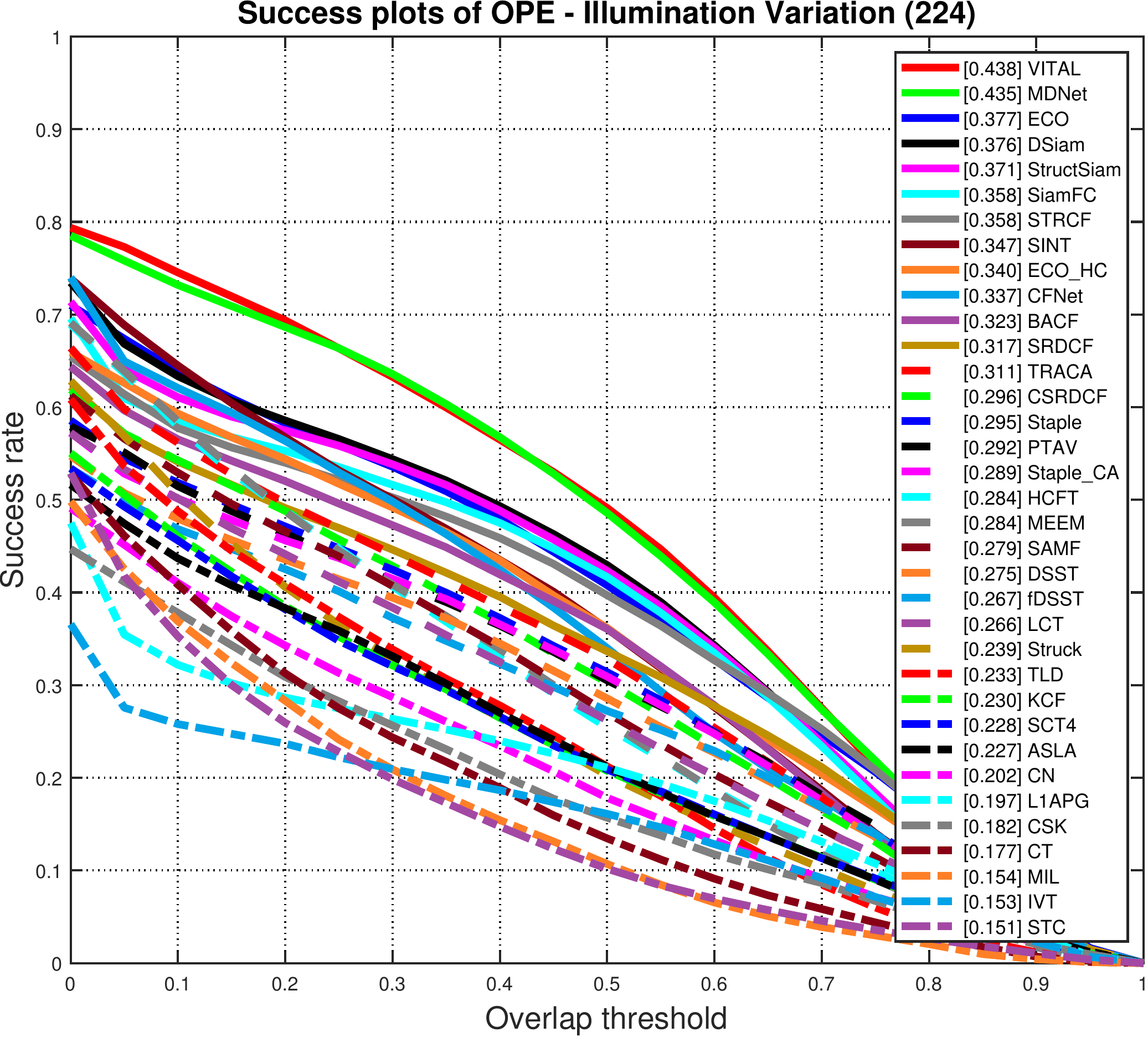}
	\includegraphics[width=4.55cm]{protocol_1_success_att/LR_overlap_OPE_AUC}
	\includegraphics[width=4.55cm]{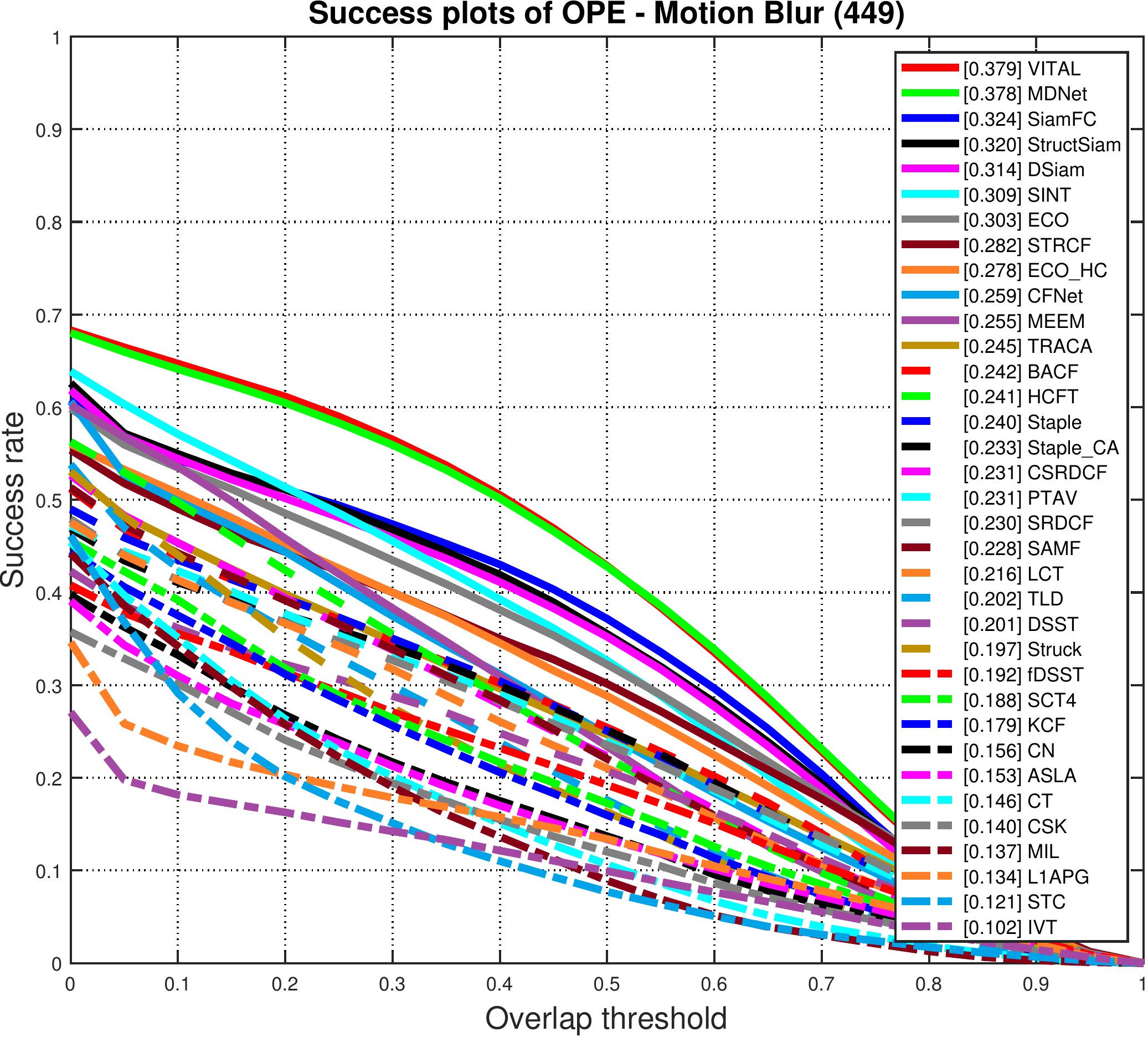}\\
	\includegraphics[width=4.55cm]{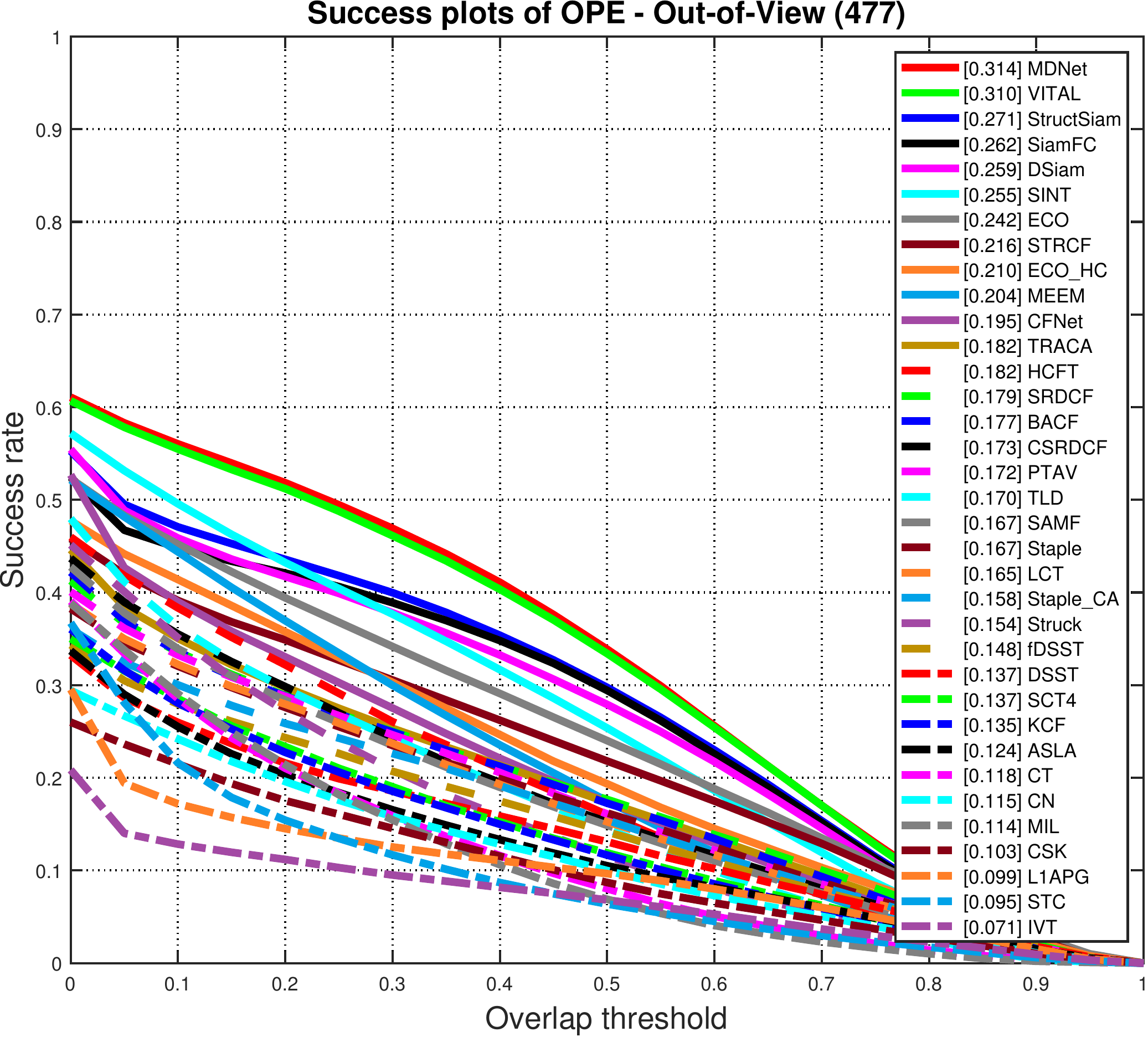}
	\includegraphics[width=4.55cm]{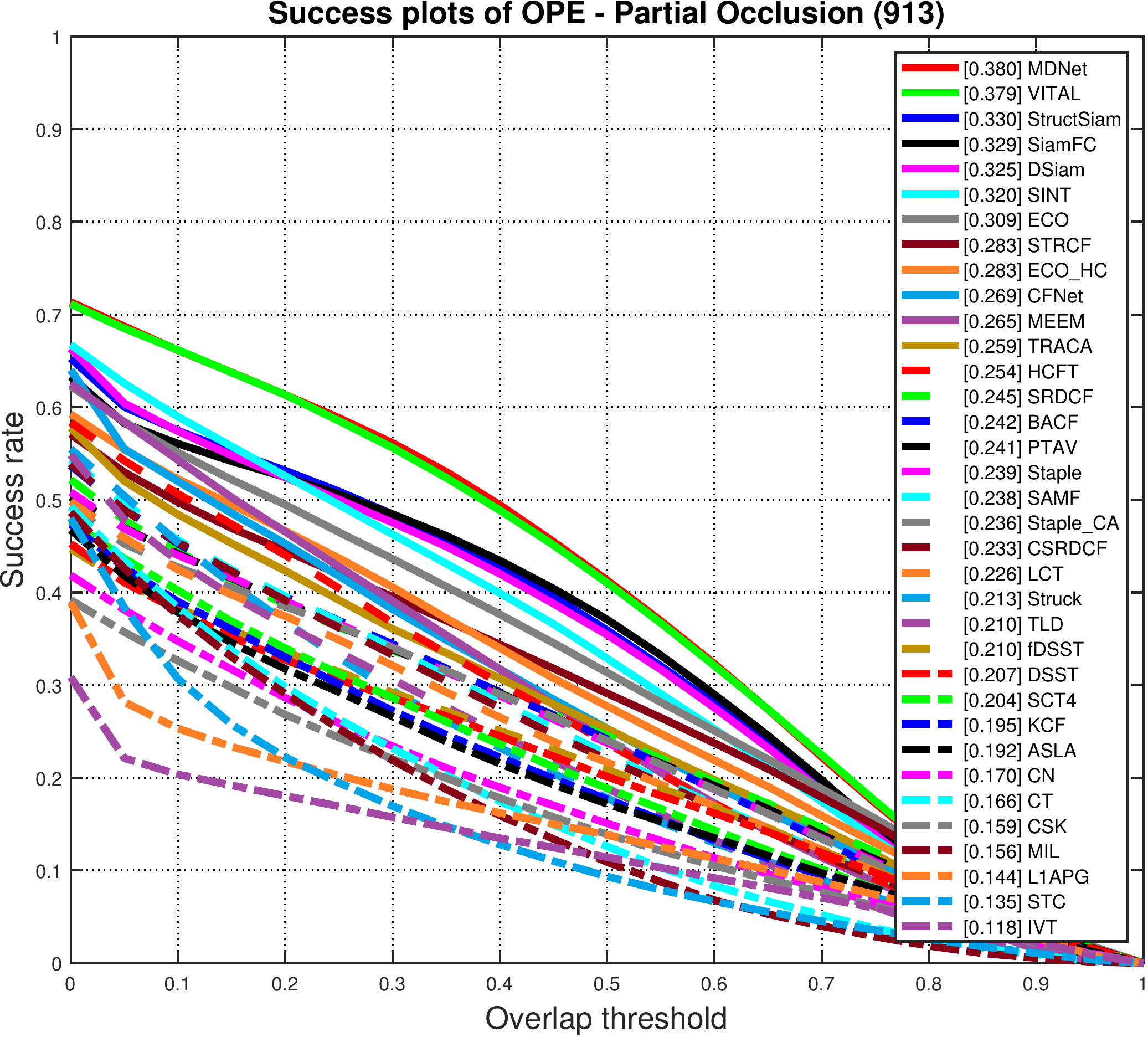}
	\includegraphics[width=4.55cm]{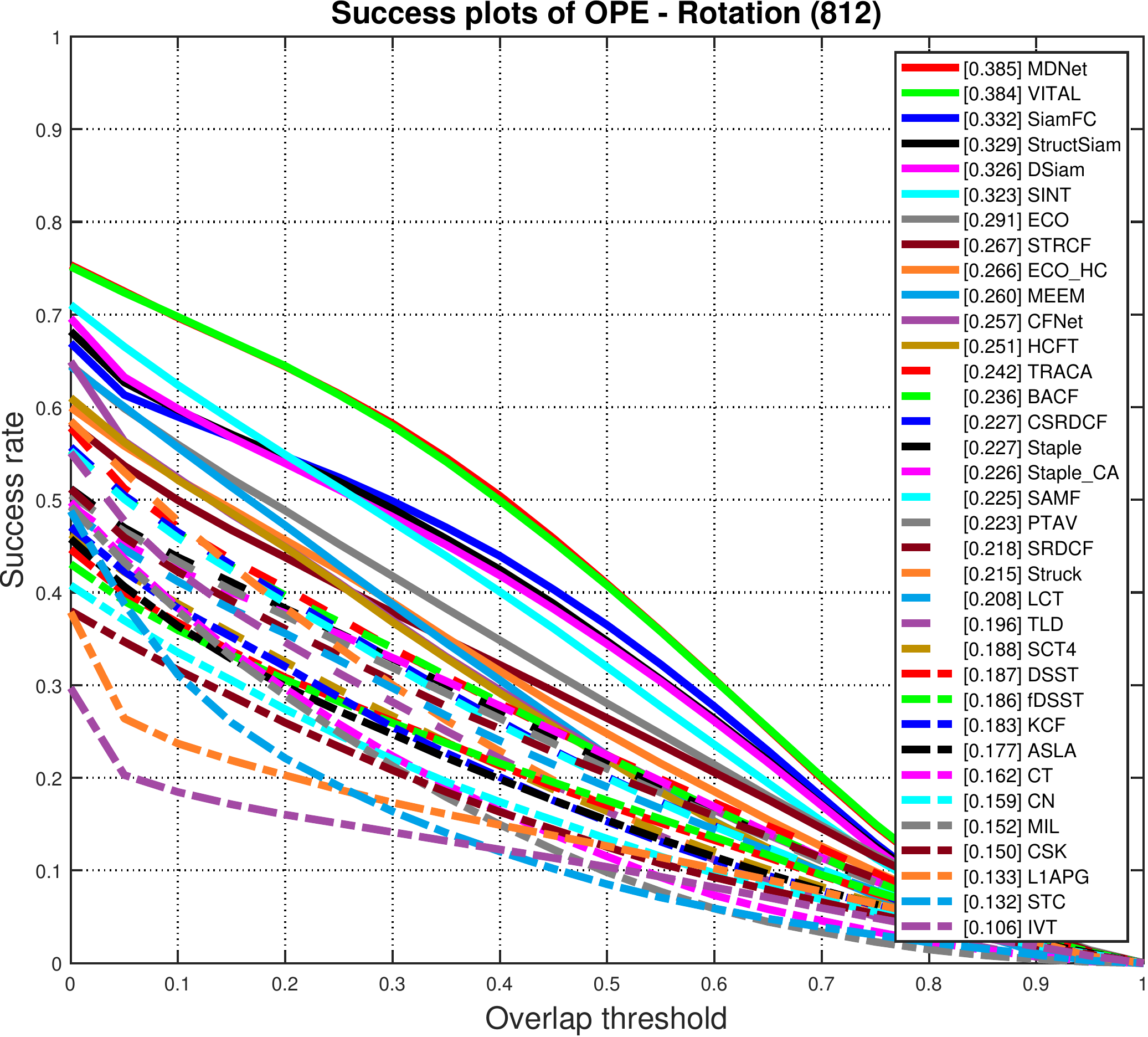}\\
	\includegraphics[width=4.55cm]{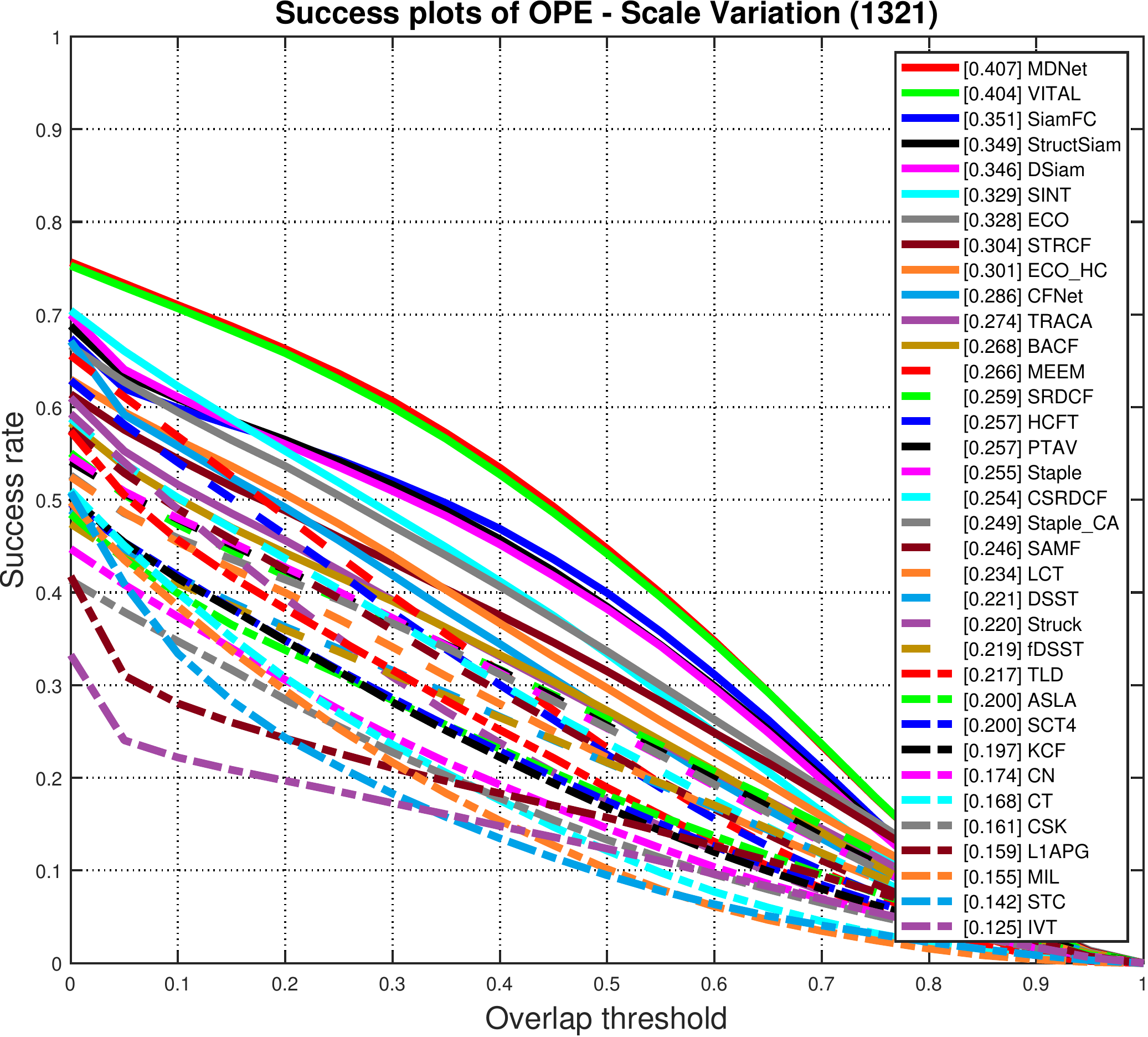}
	\includegraphics[width=4.55cm]{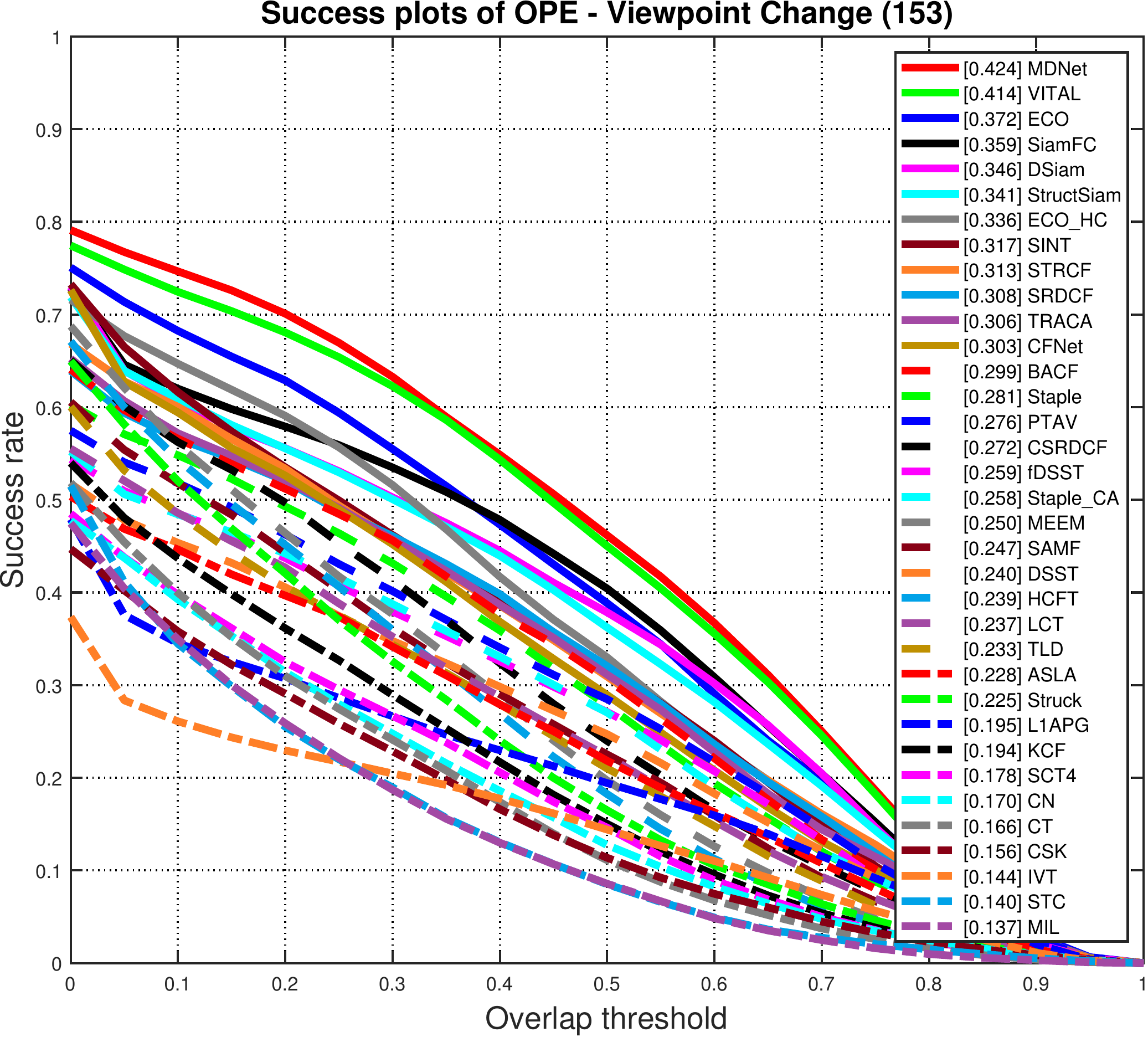}\\
	\caption{Performance of trackers on each attribute using success under protocol \uppercase\expandafter{\romannumeral1}. Best viewed in color.}
	\label{fig:protocol_1_all_att_res_success}
\end{figure*}

\newpage
\section{Detailed Attribute-based Performance under Protocol \uppercase\expandafter{\romannumeral2}}

Fig.~\ref{fig:protocol_2_all_att_res_precision} shows the performance of trackers on each attribute using precision under protocol \uppercase\expandafter{\romannumeral2}.
\begin{figure*}[!hbpt]
	\centering
	\includegraphics[width=4.55cm]{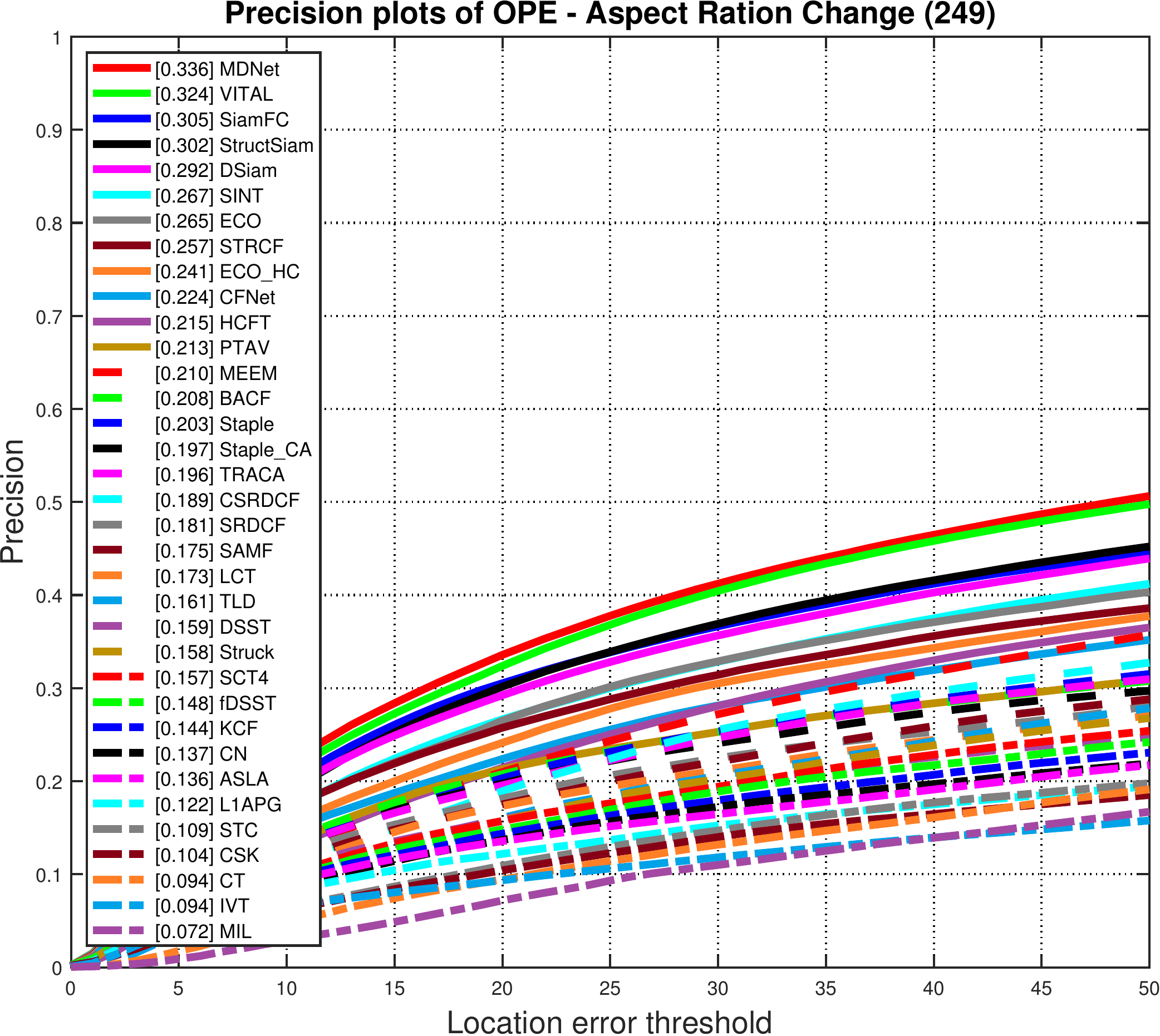}
	\includegraphics[width=4.55cm]{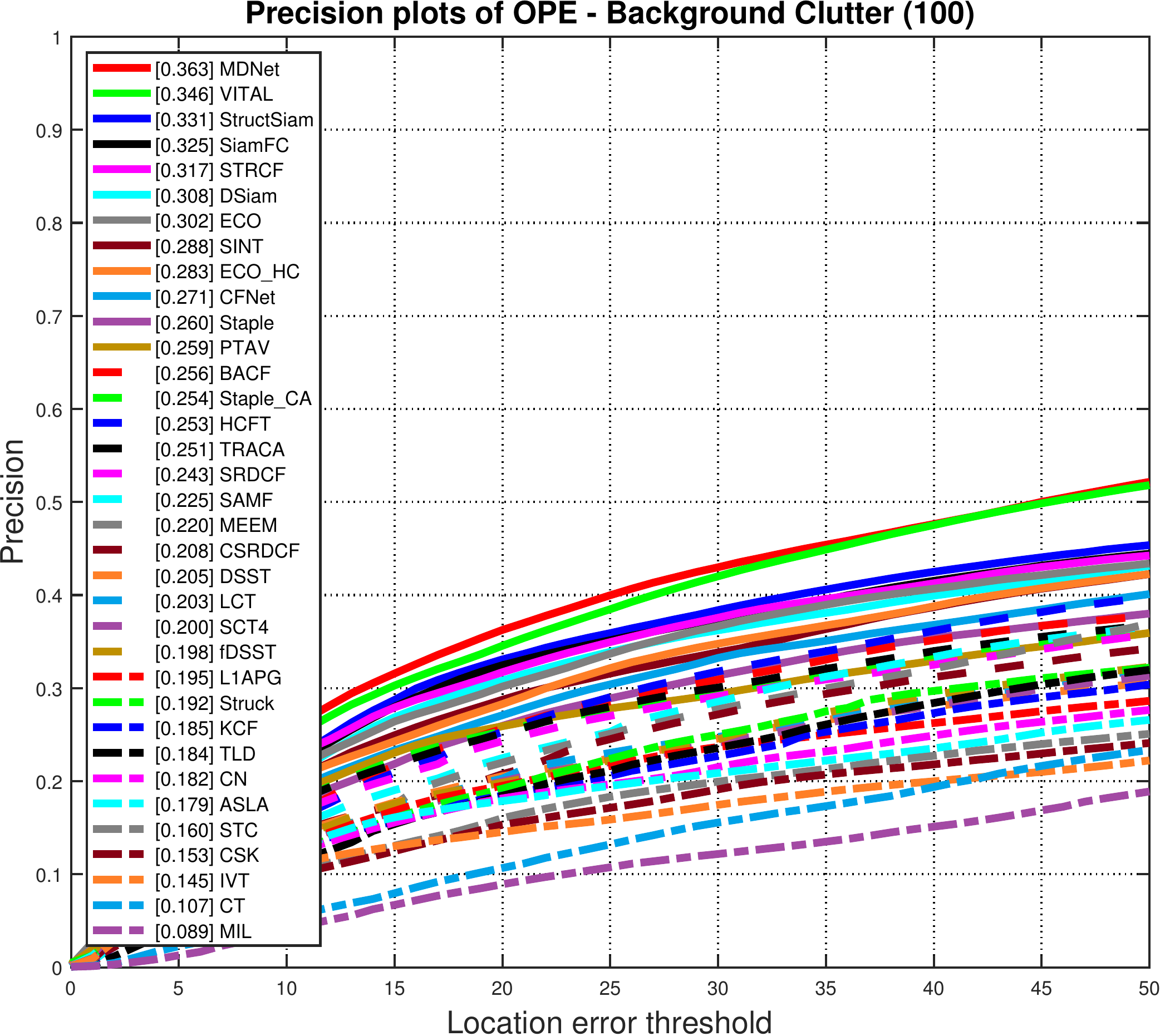}
	\includegraphics[width=4.55cm]{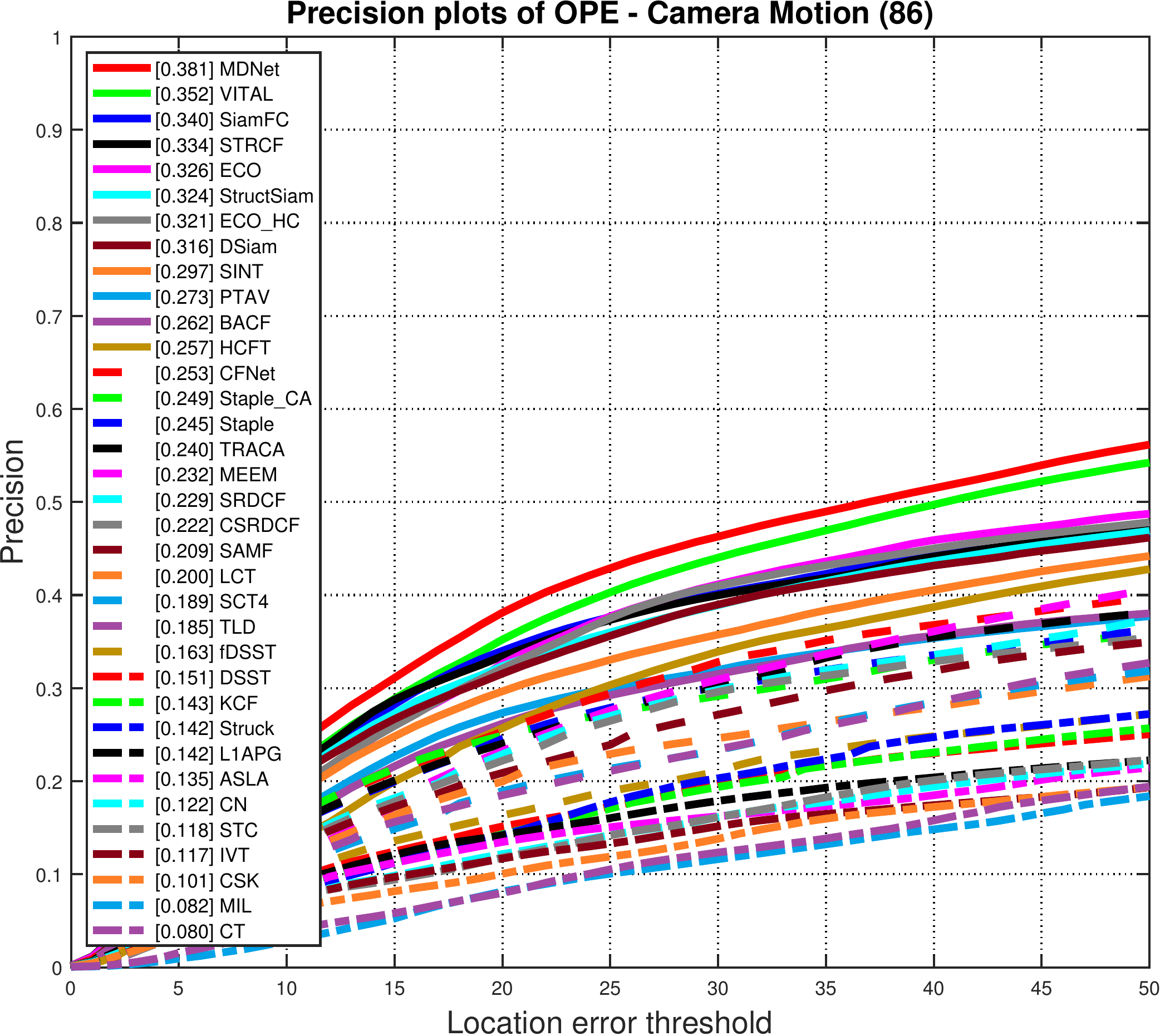}\\
	\includegraphics[width=4.55cm]{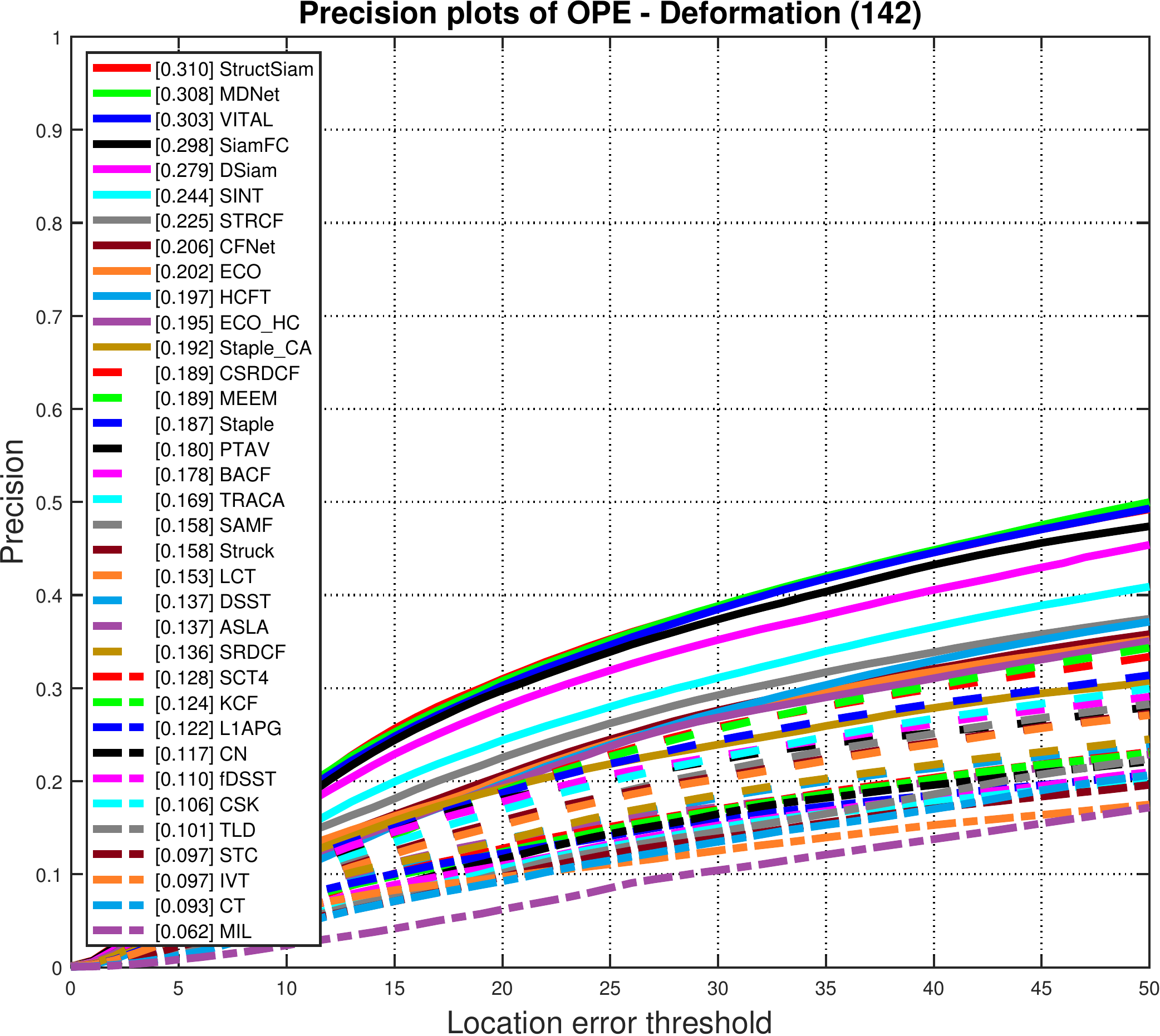}
	\includegraphics[width=4.55cm]{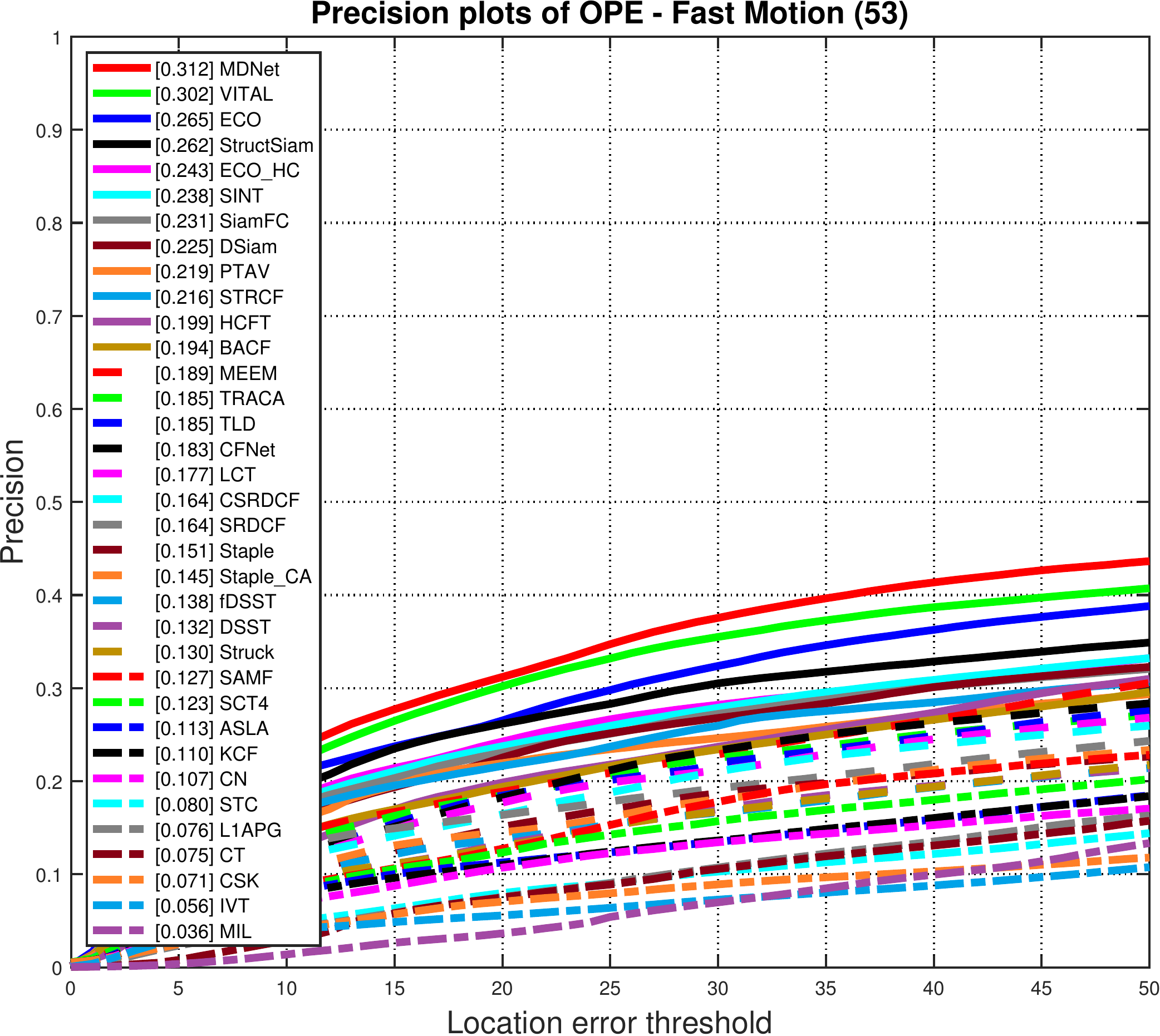}
	\includegraphics[width=4.55cm]{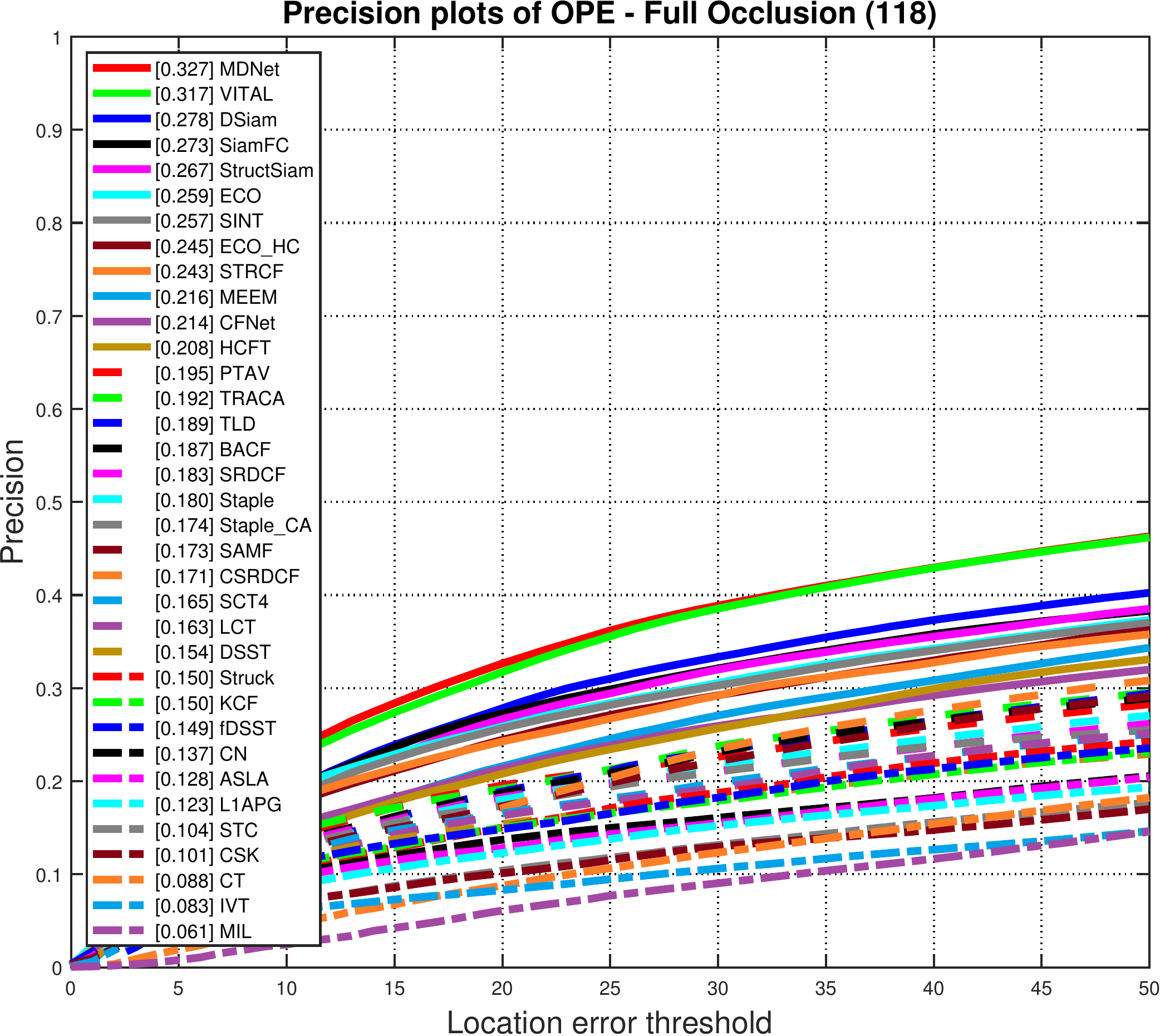}\\
	\includegraphics[width=4.55cm]{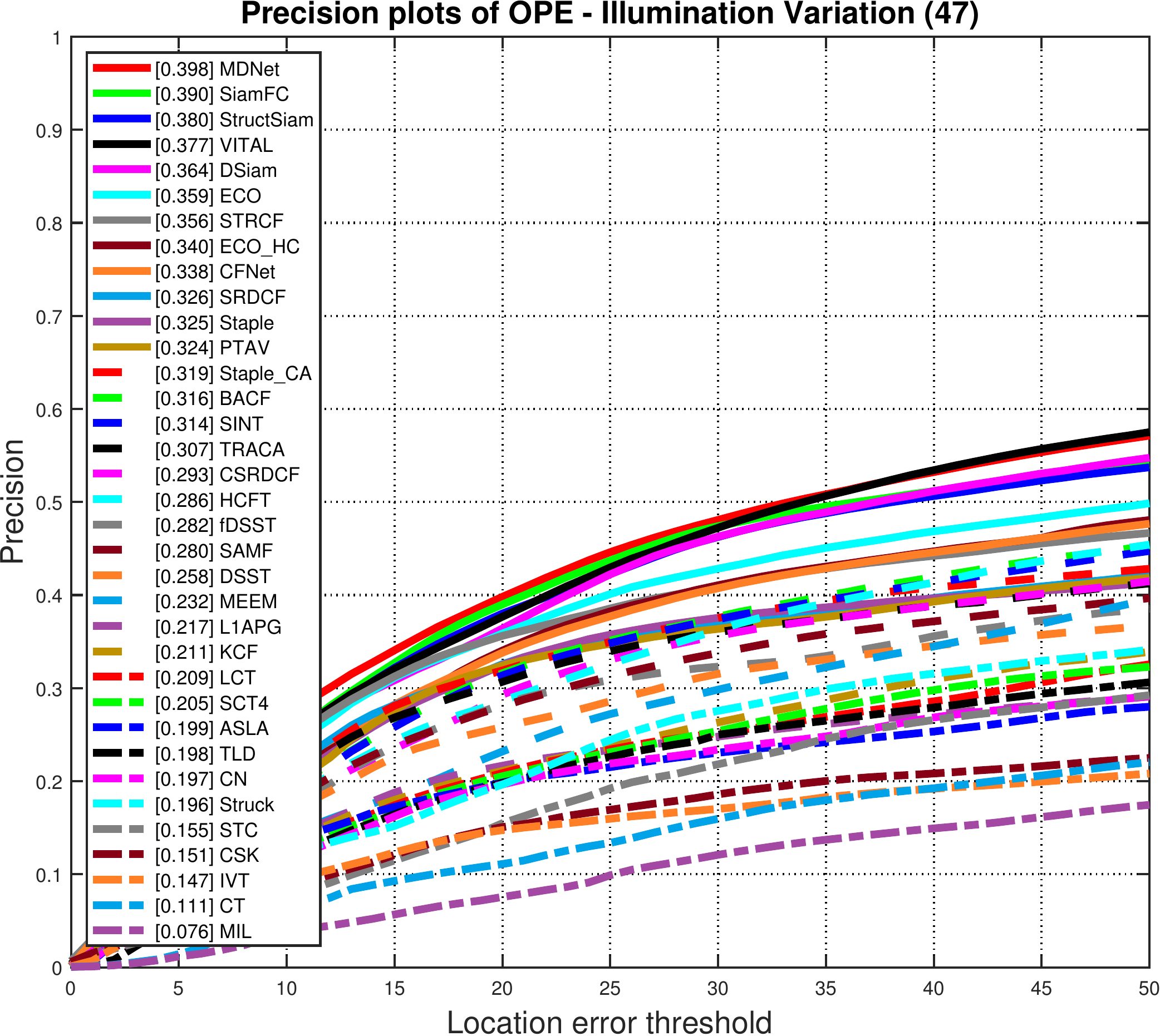}
	\includegraphics[width=4.55cm]{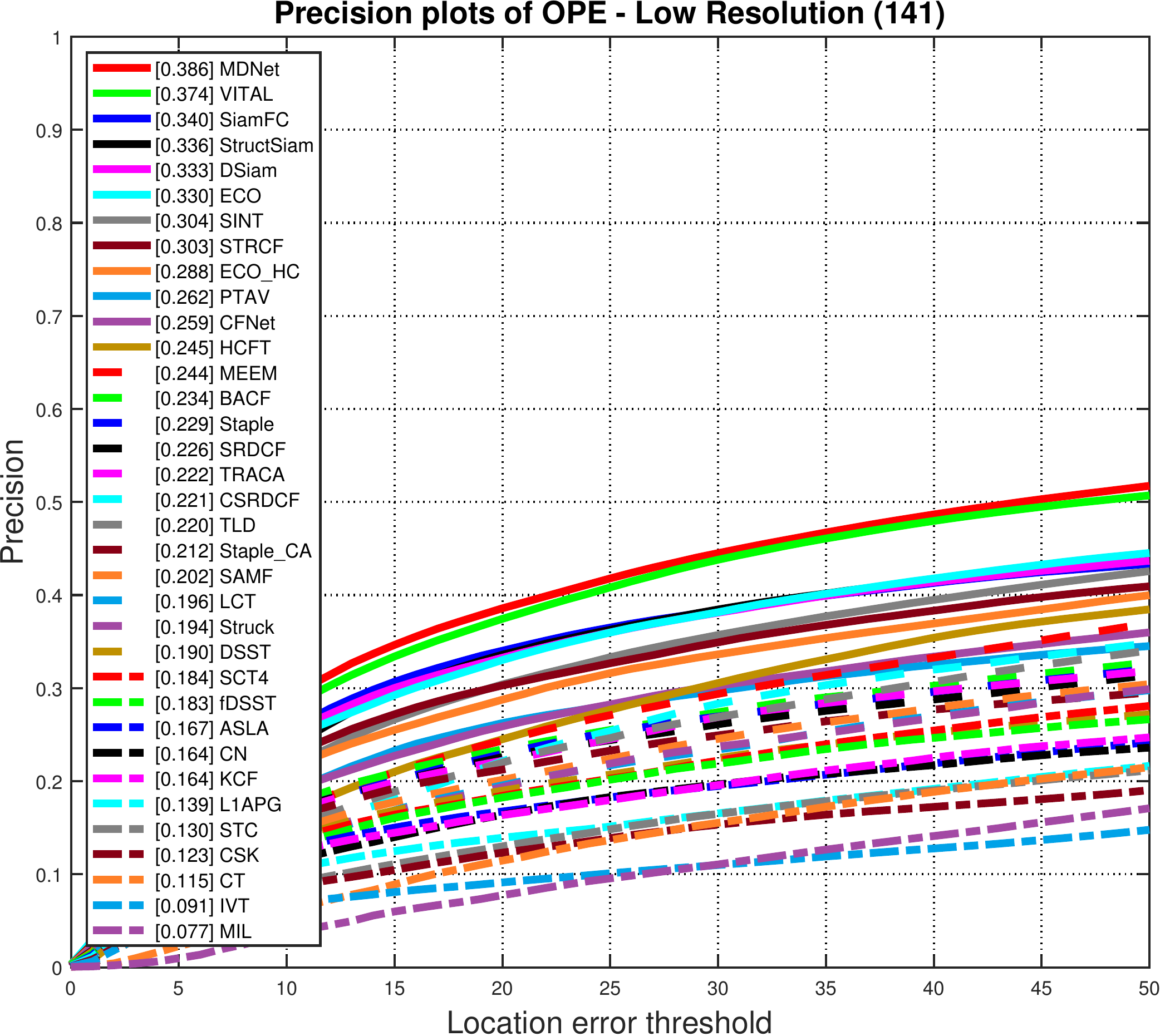}
	\includegraphics[width=4.55cm]{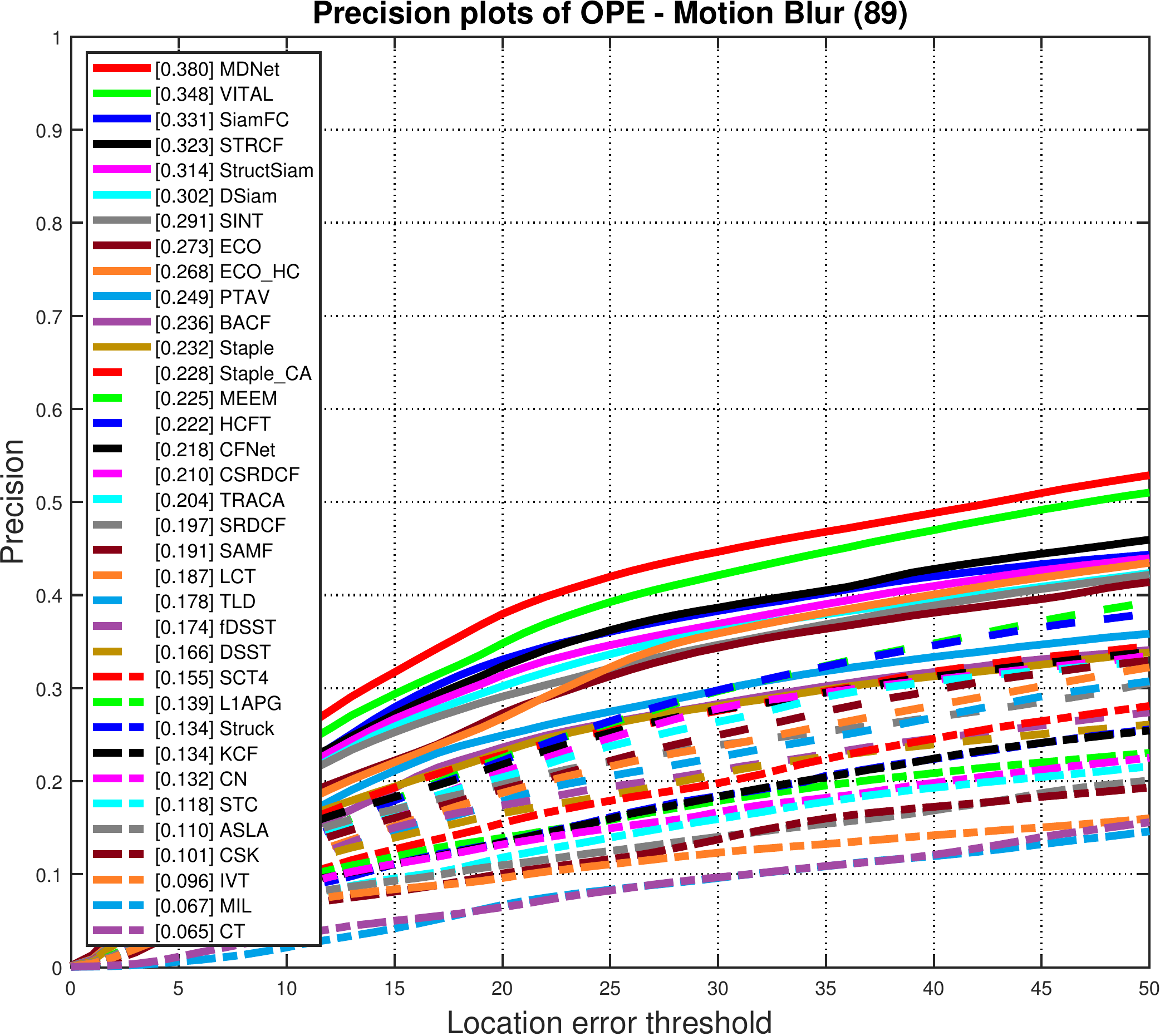}\\
	\includegraphics[width=4.55cm]{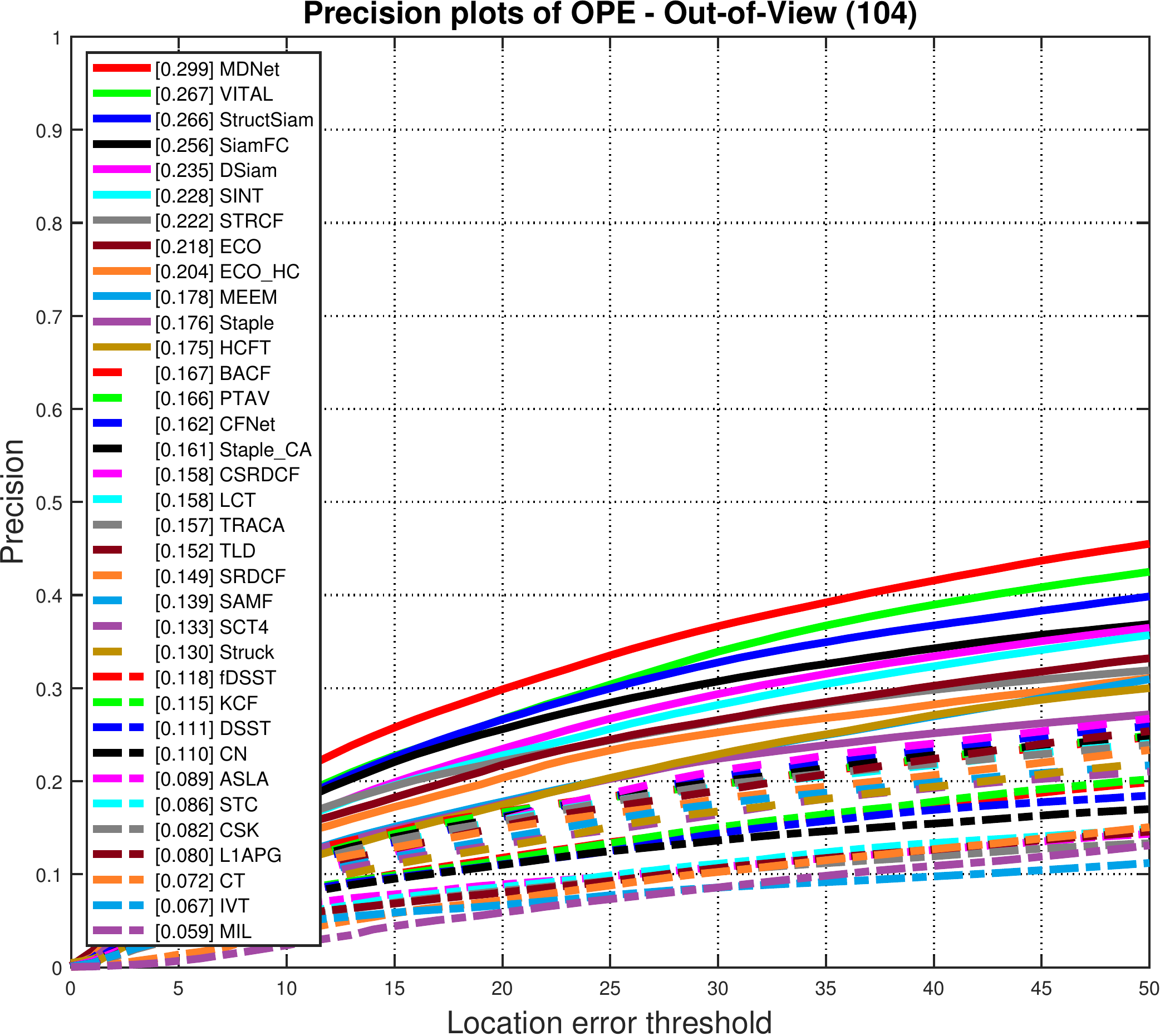}
	\includegraphics[width=4.55cm]{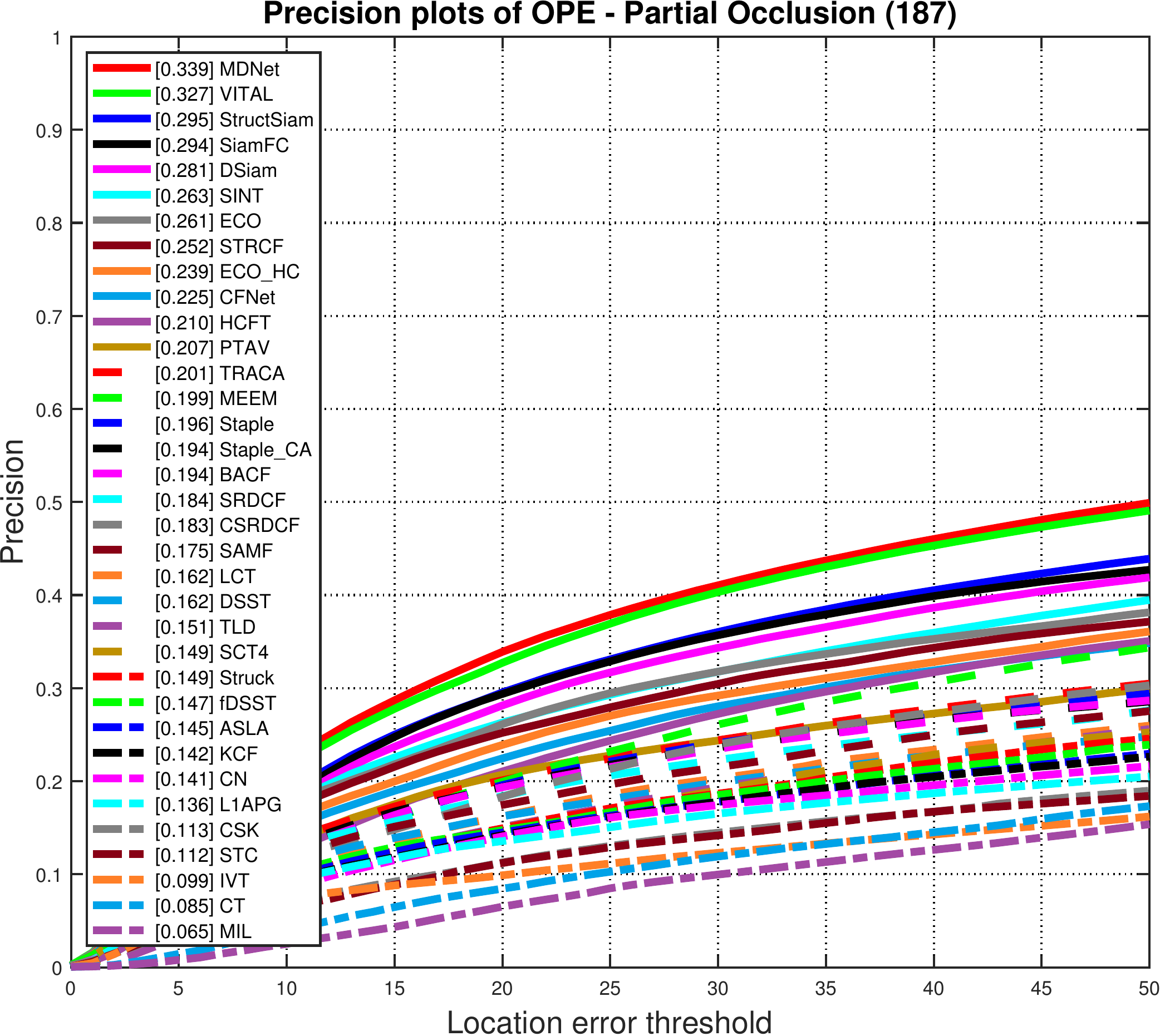}
	\includegraphics[width=4.55cm]{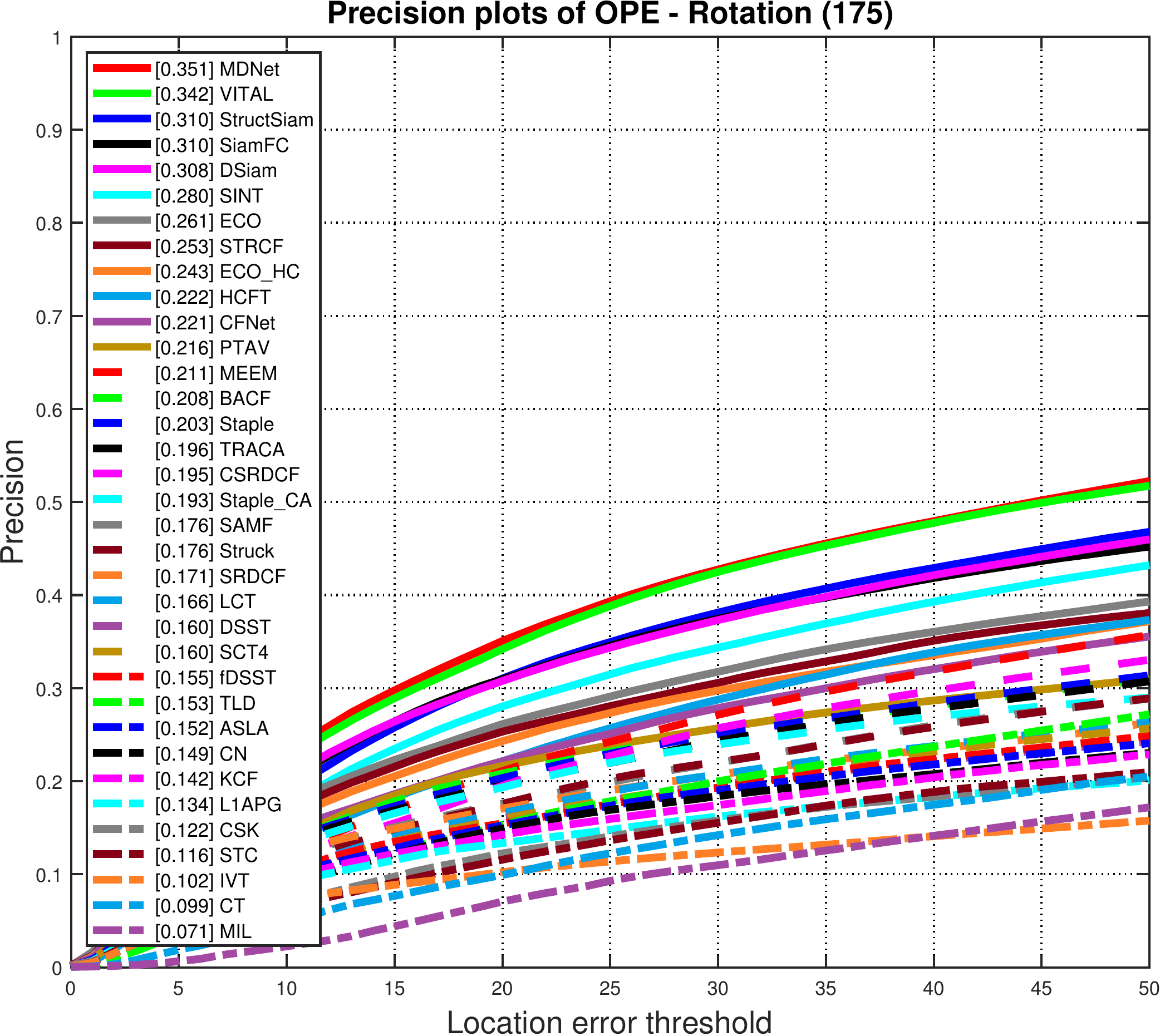}\\
	\includegraphics[width=4.55cm]{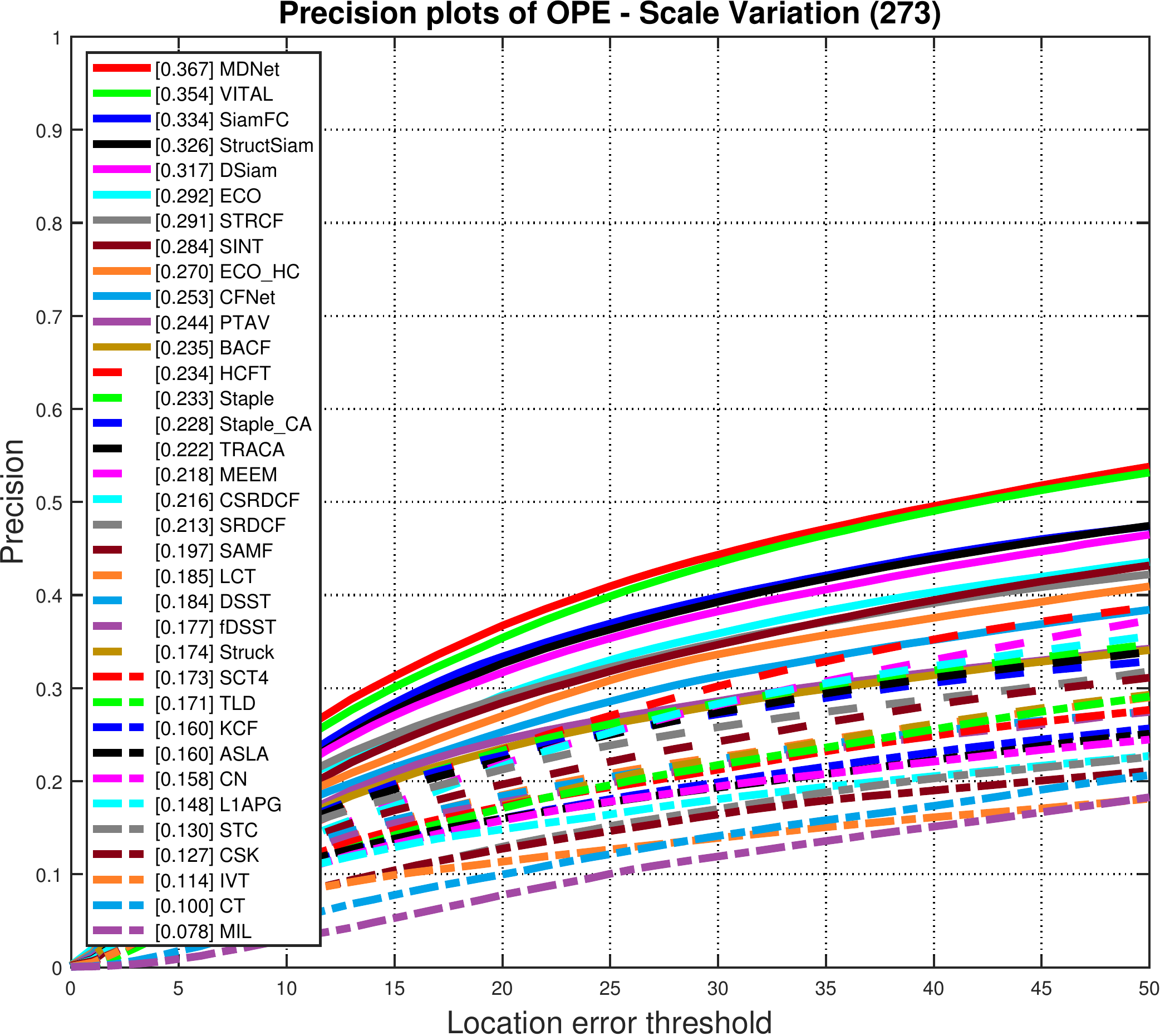}
	\includegraphics[width=4.55cm]{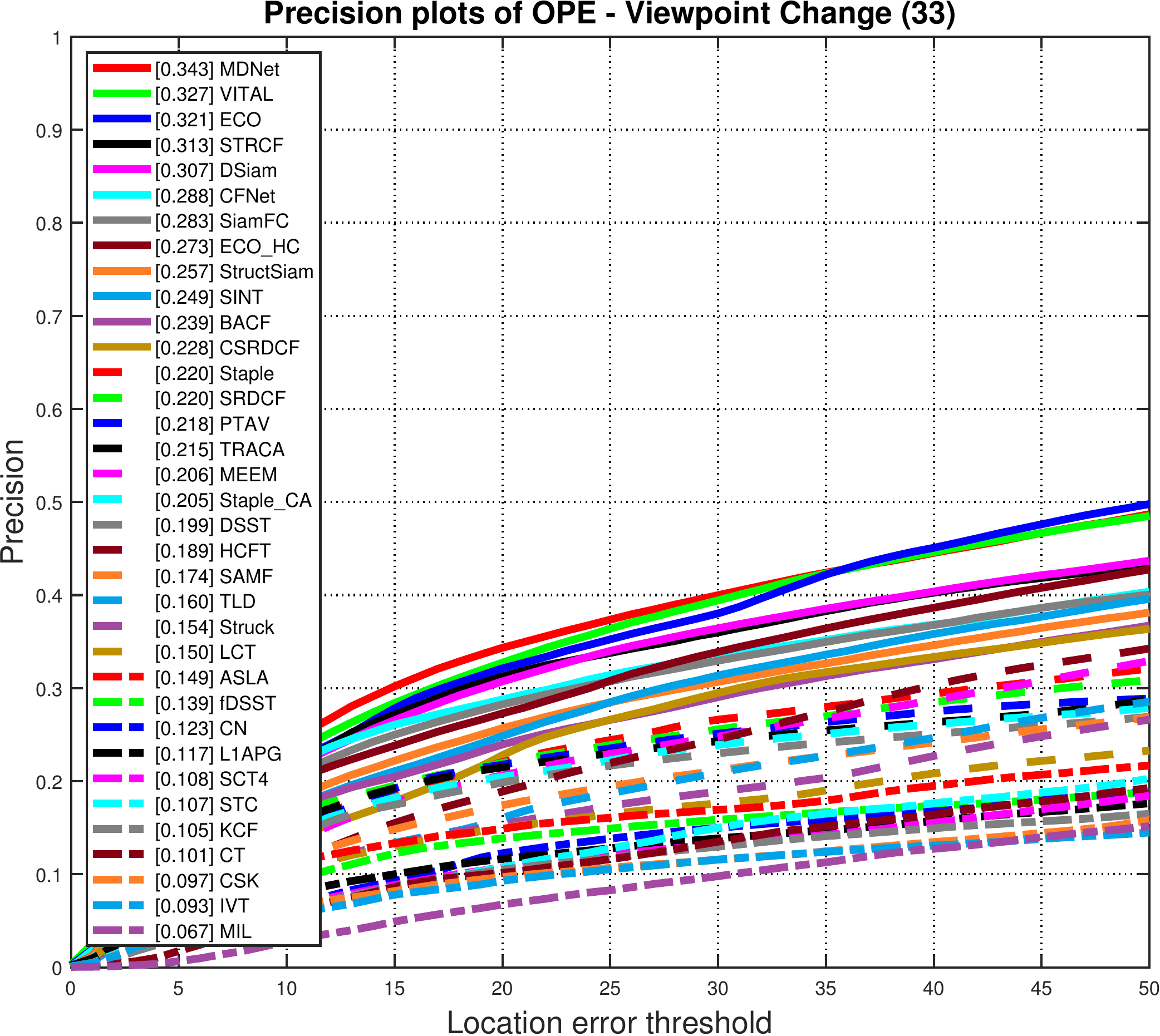}\\
	\caption{Performance of trackers on each attribute using precision under protocol \uppercase\expandafter{\romannumeral2}. Best viewed in color.}
	\label{fig:protocol_2_all_att_res_precision}
\end{figure*}

\newpage
Fig.~\ref{fig:protocol_2_all_att_res_norm_precision} shows the performance of trackers on each attribute using normalized precision under protocol \uppercase\expandafter{\romannumeral2}.
\begin{figure*}[!hbpt]
	\centering
	\includegraphics[width=4.55cm]{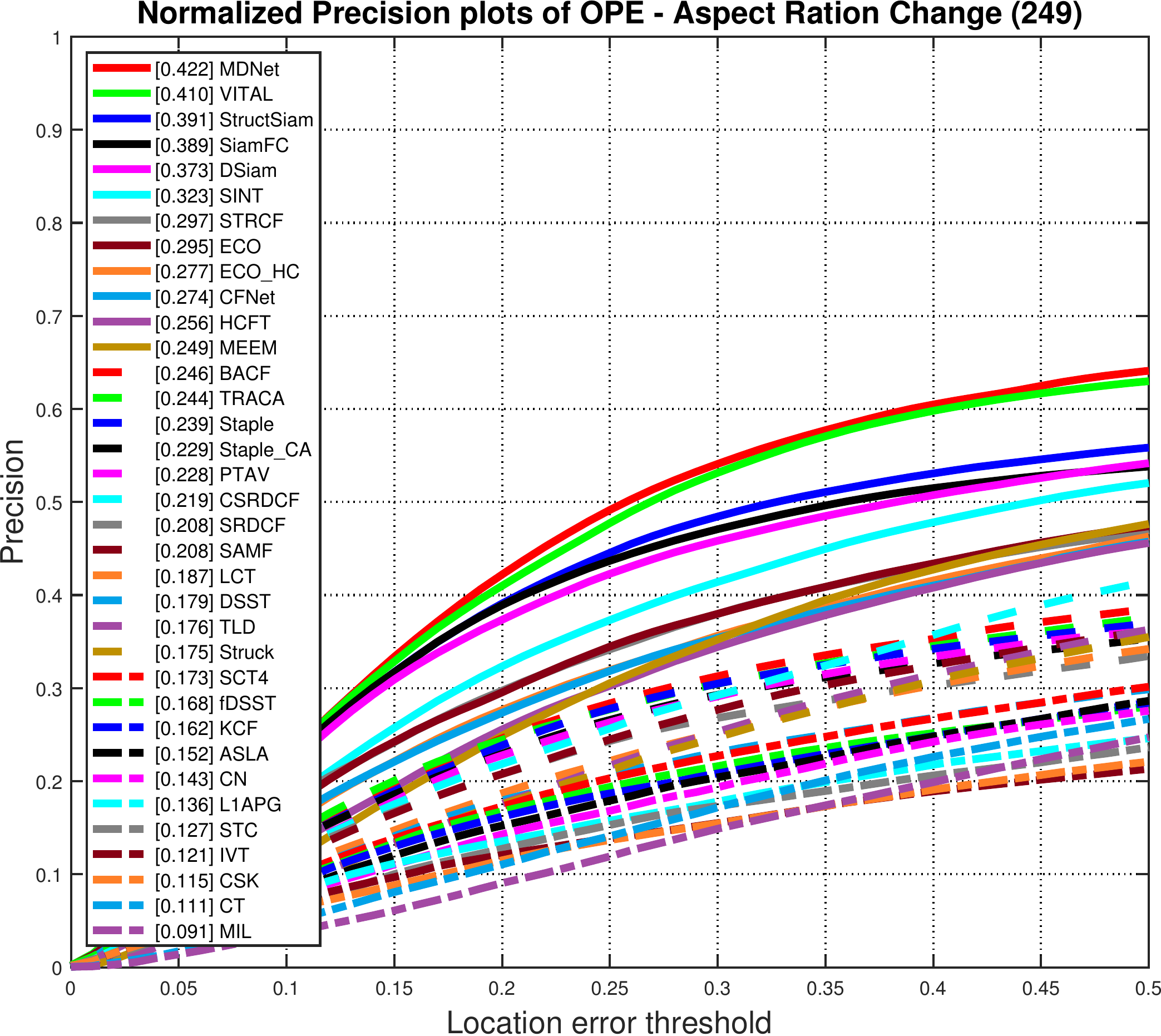}
	\includegraphics[width=4.55cm]{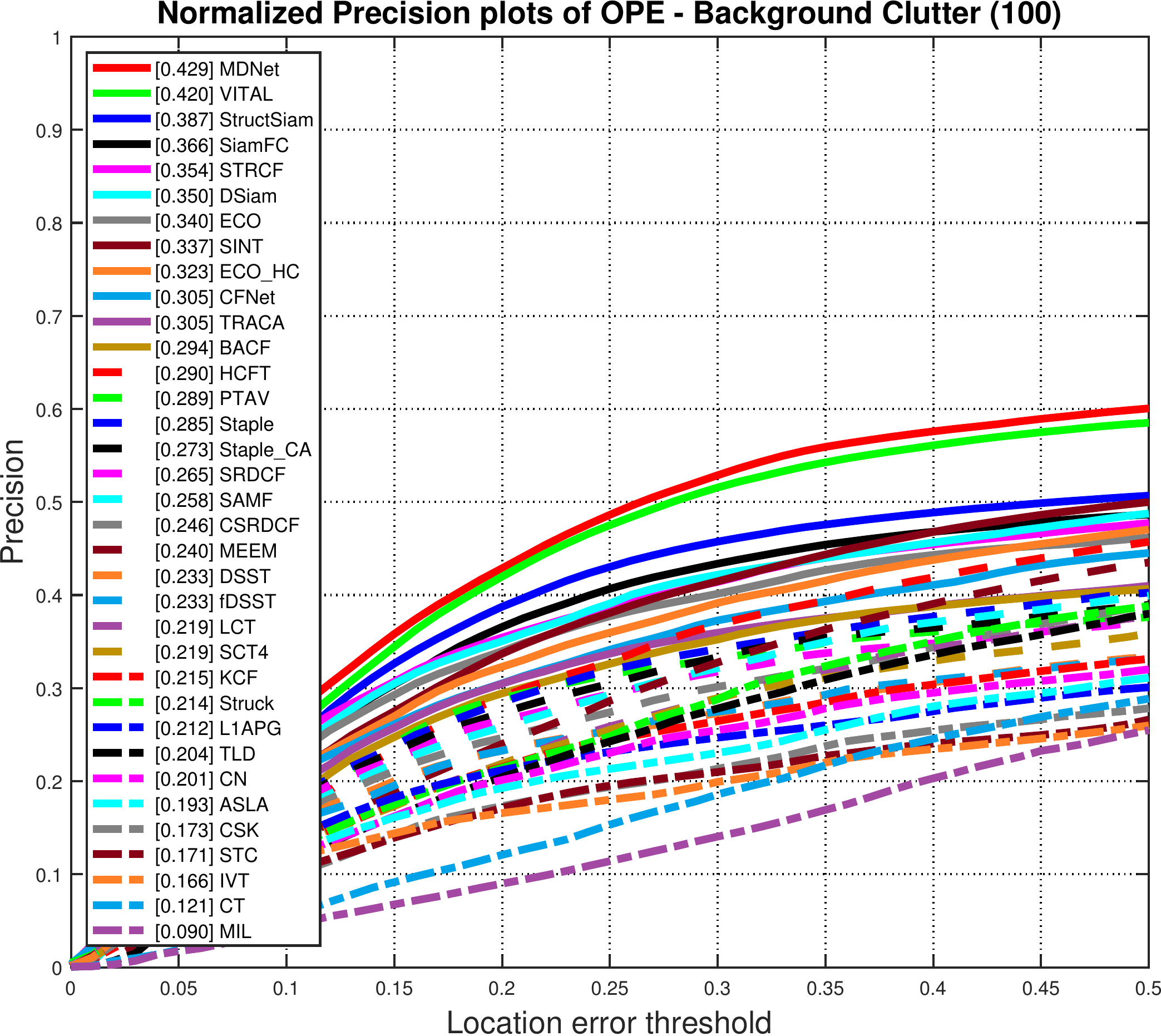}
	\includegraphics[width=4.55cm]{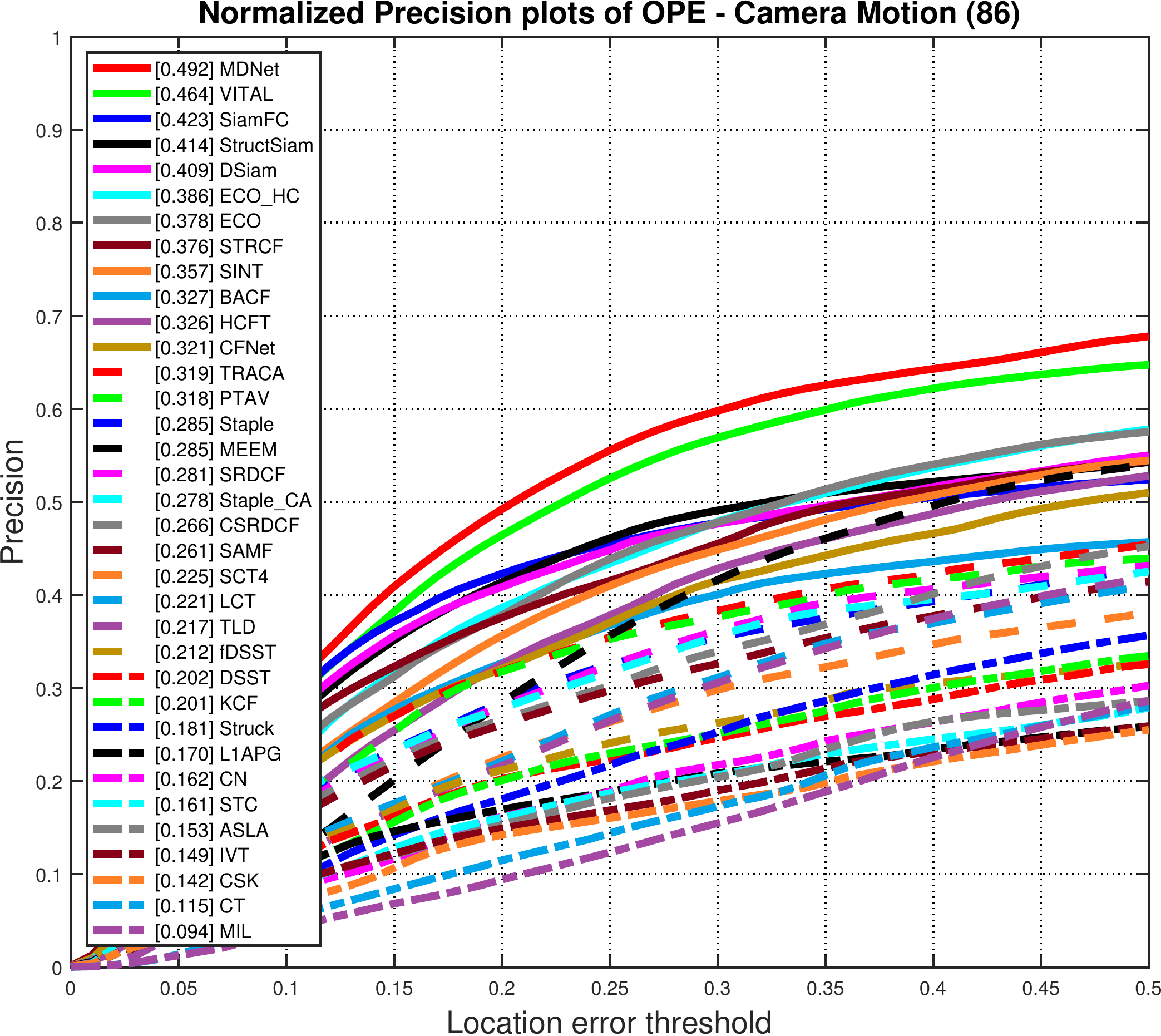}\\
	\includegraphics[width=4.55cm]{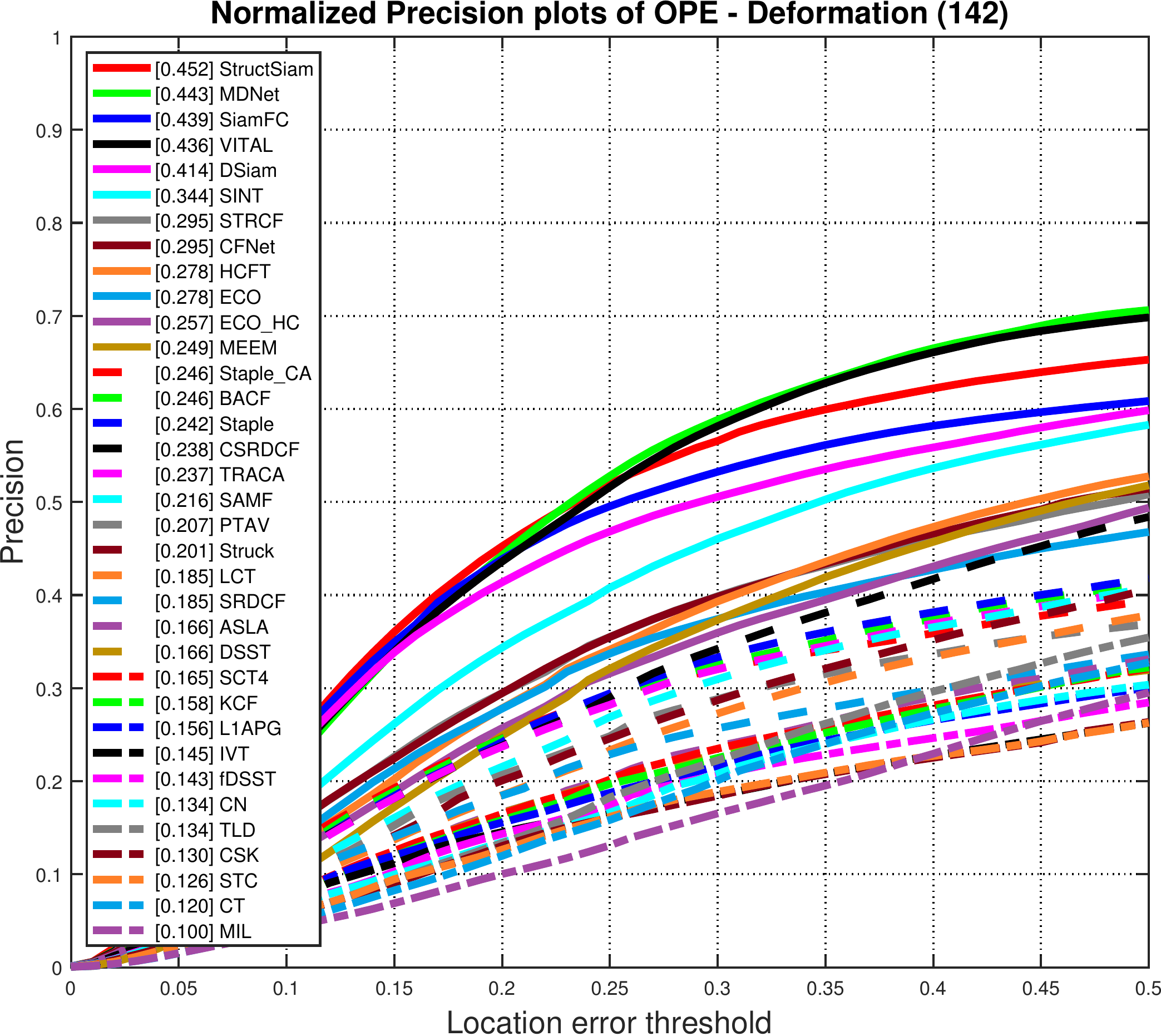}
	\includegraphics[width=4.55cm]{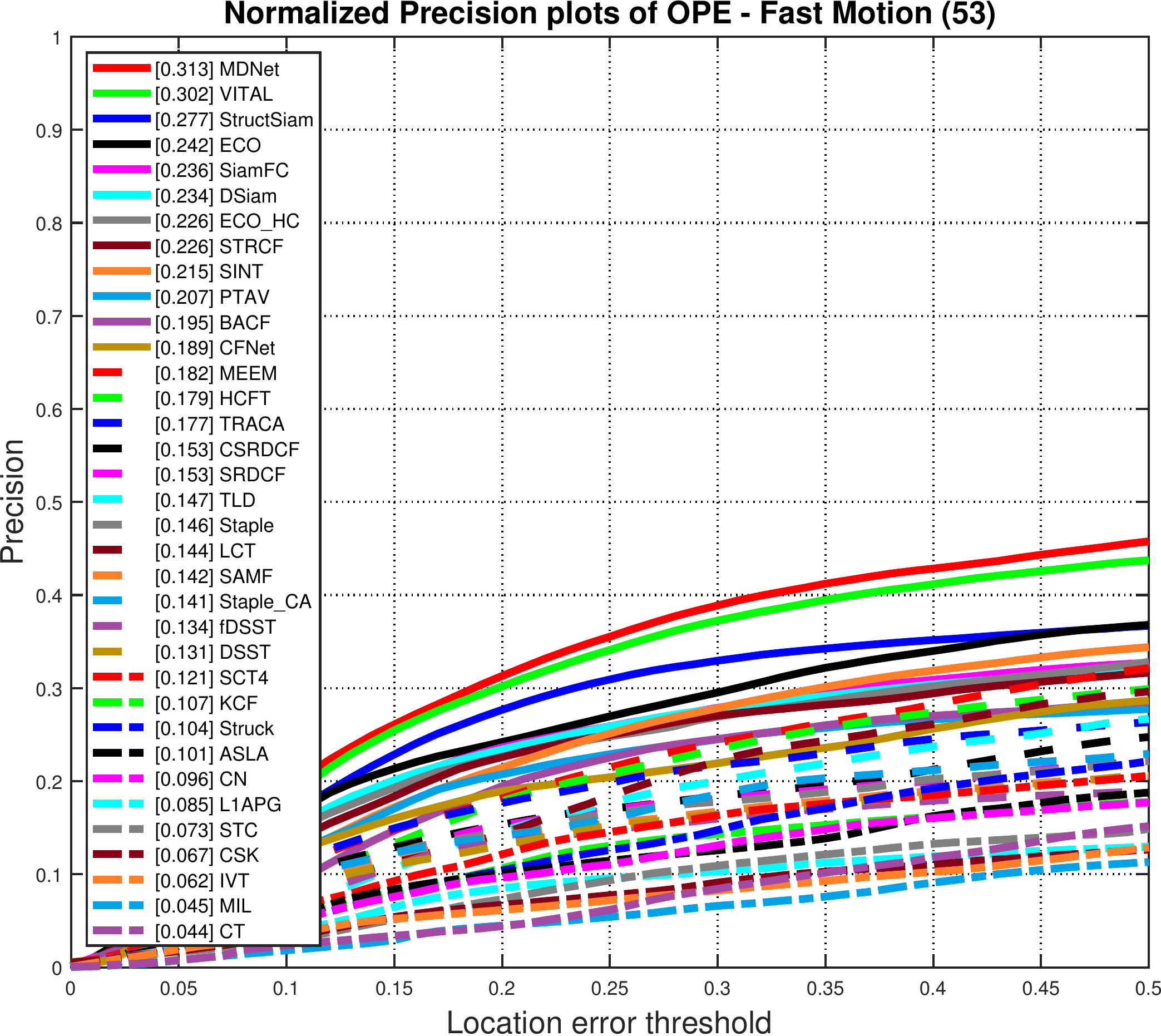}
	\includegraphics[width=4.55cm]{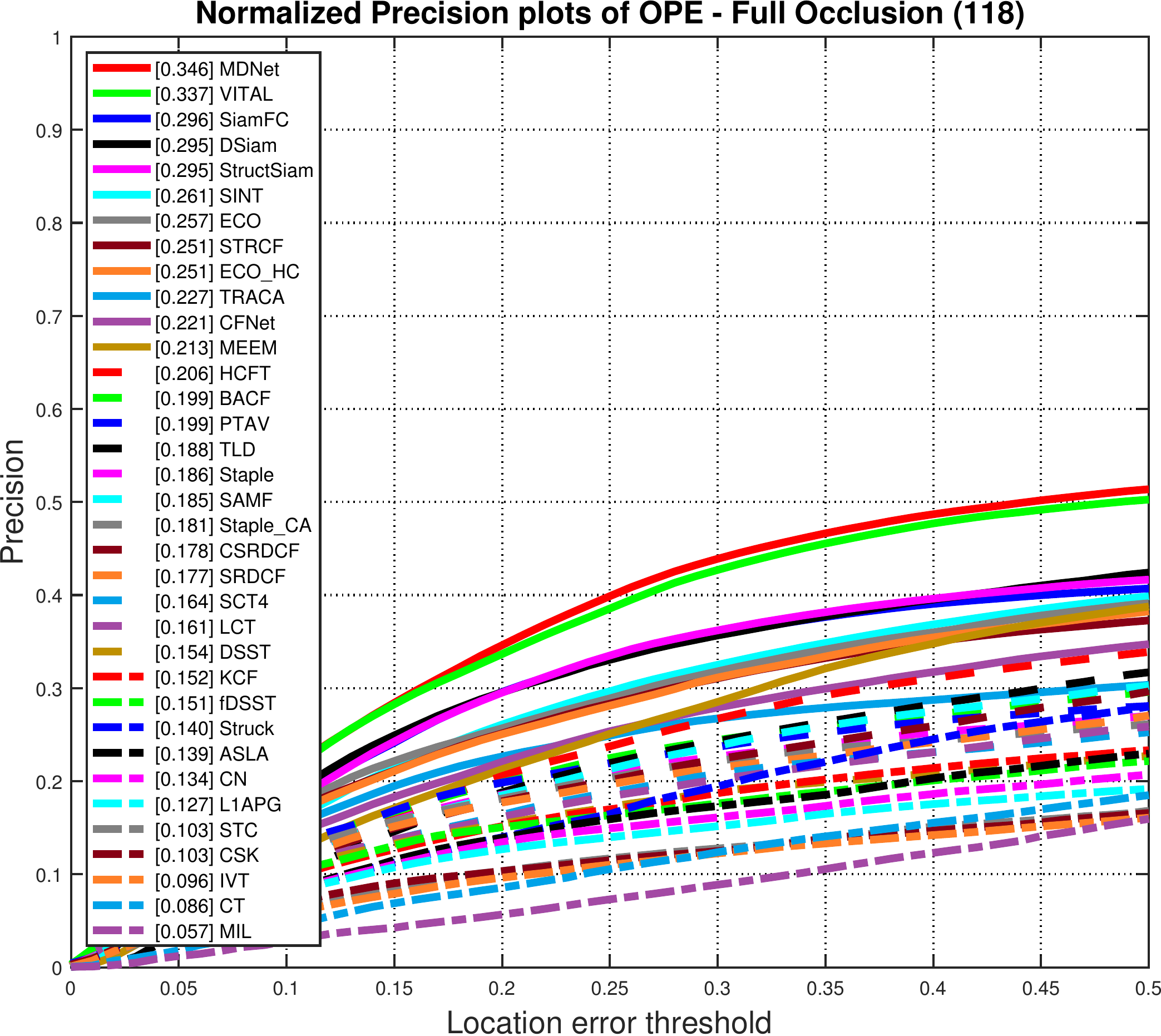}\\
	\includegraphics[width=4.55cm]{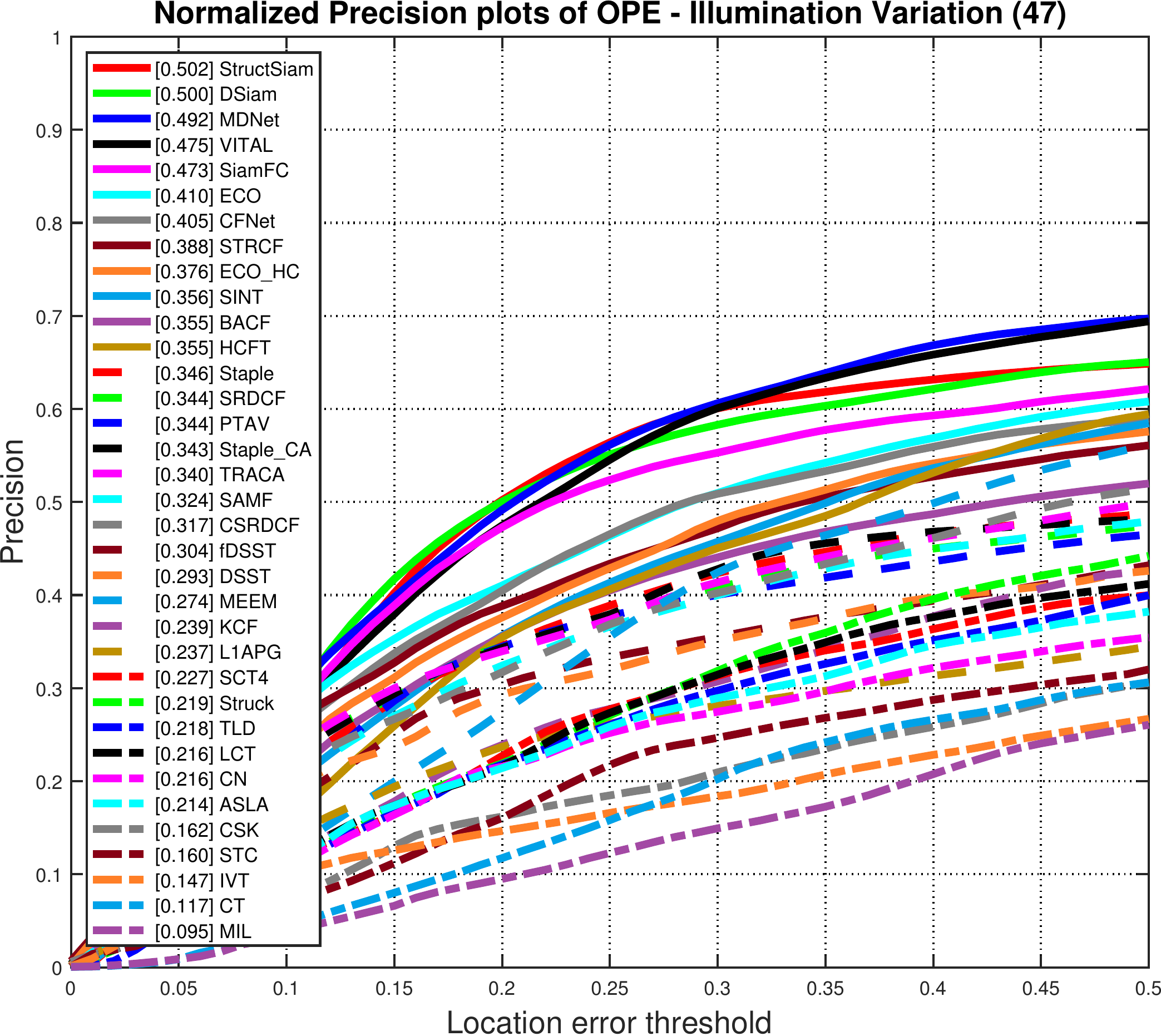}
	\includegraphics[width=4.55cm]{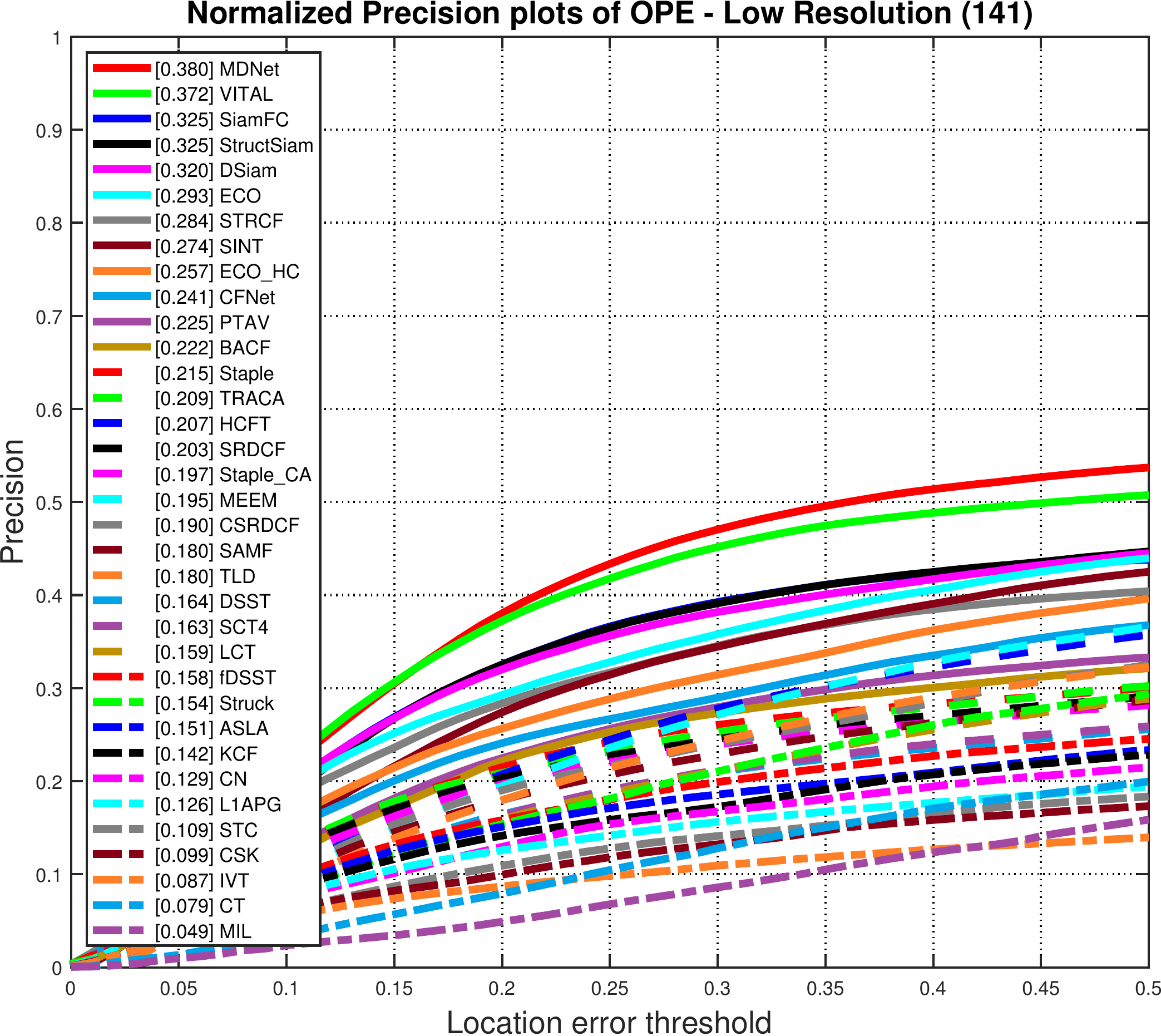}
	\includegraphics[width=4.55cm]{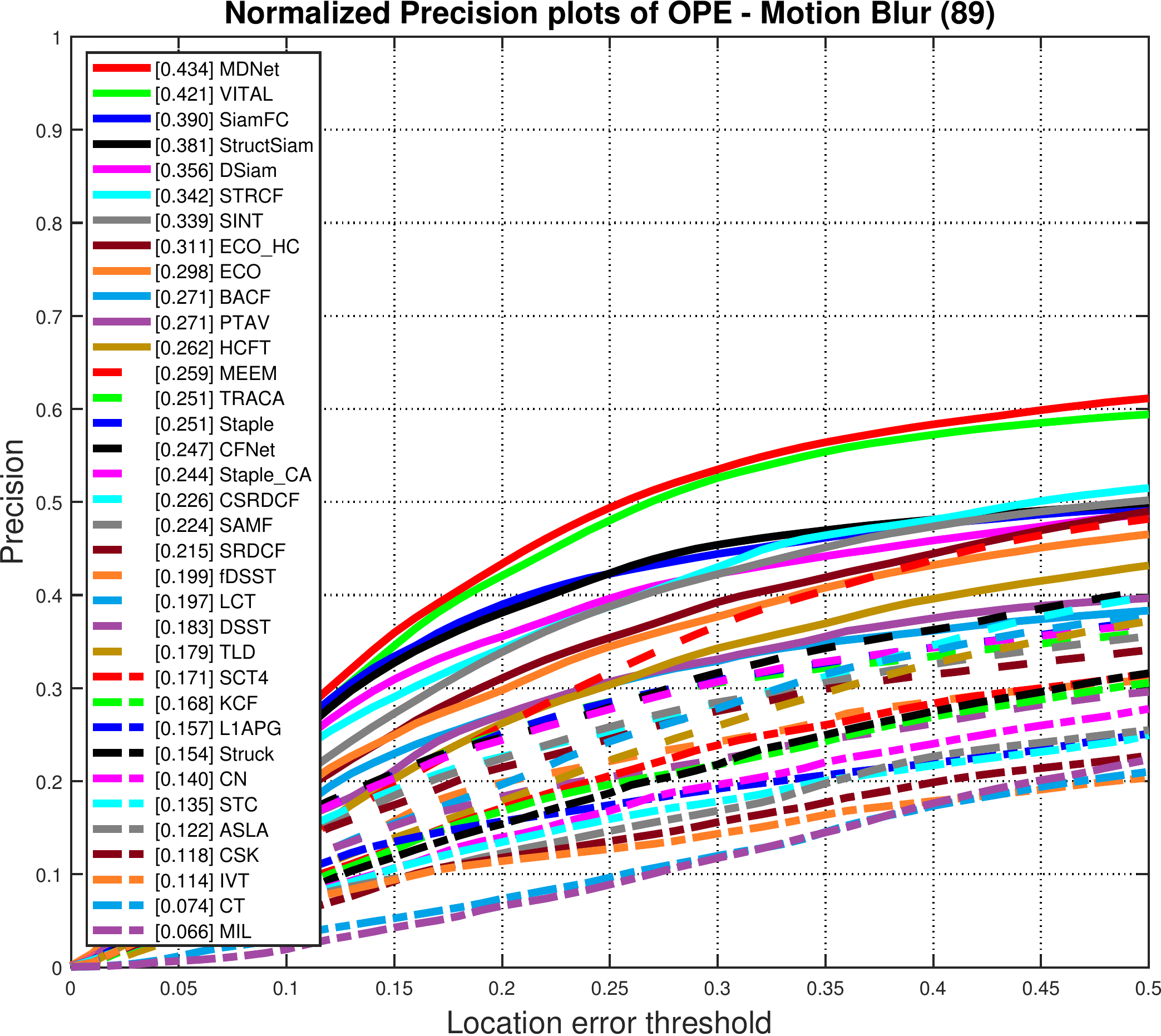}\\
	\includegraphics[width=4.55cm]{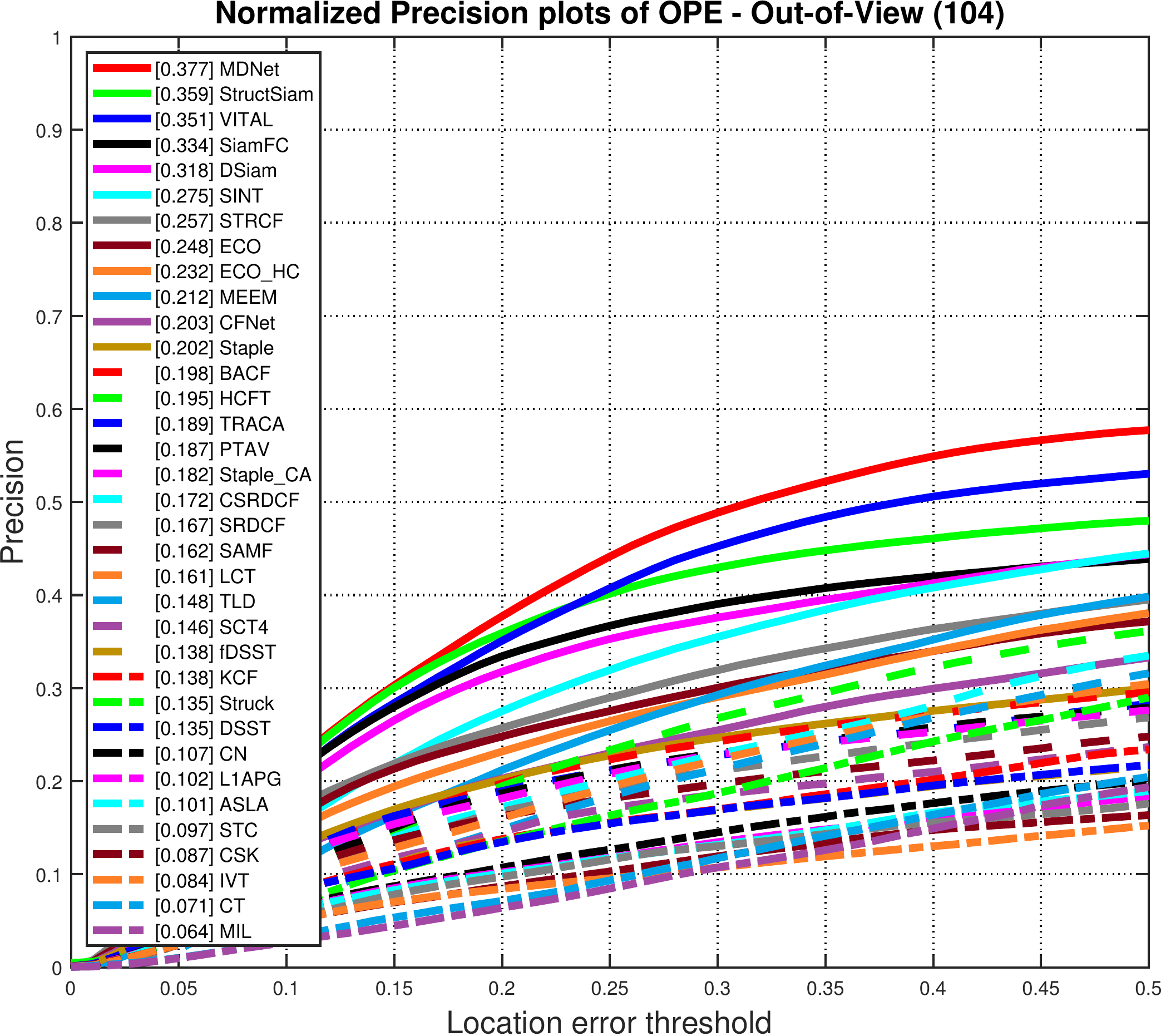}
	\includegraphics[width=4.55cm]{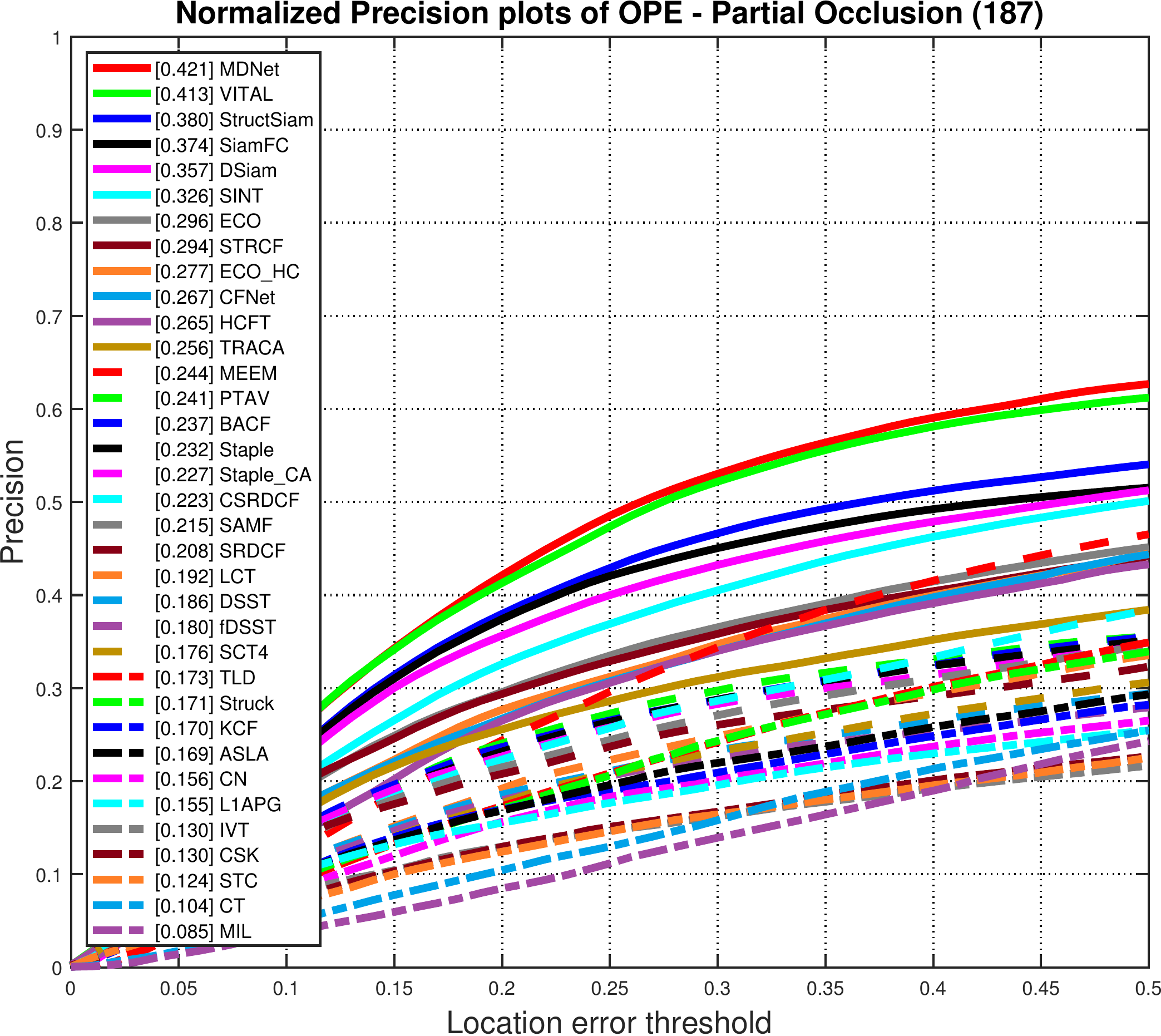}
	\includegraphics[width=4.55cm]{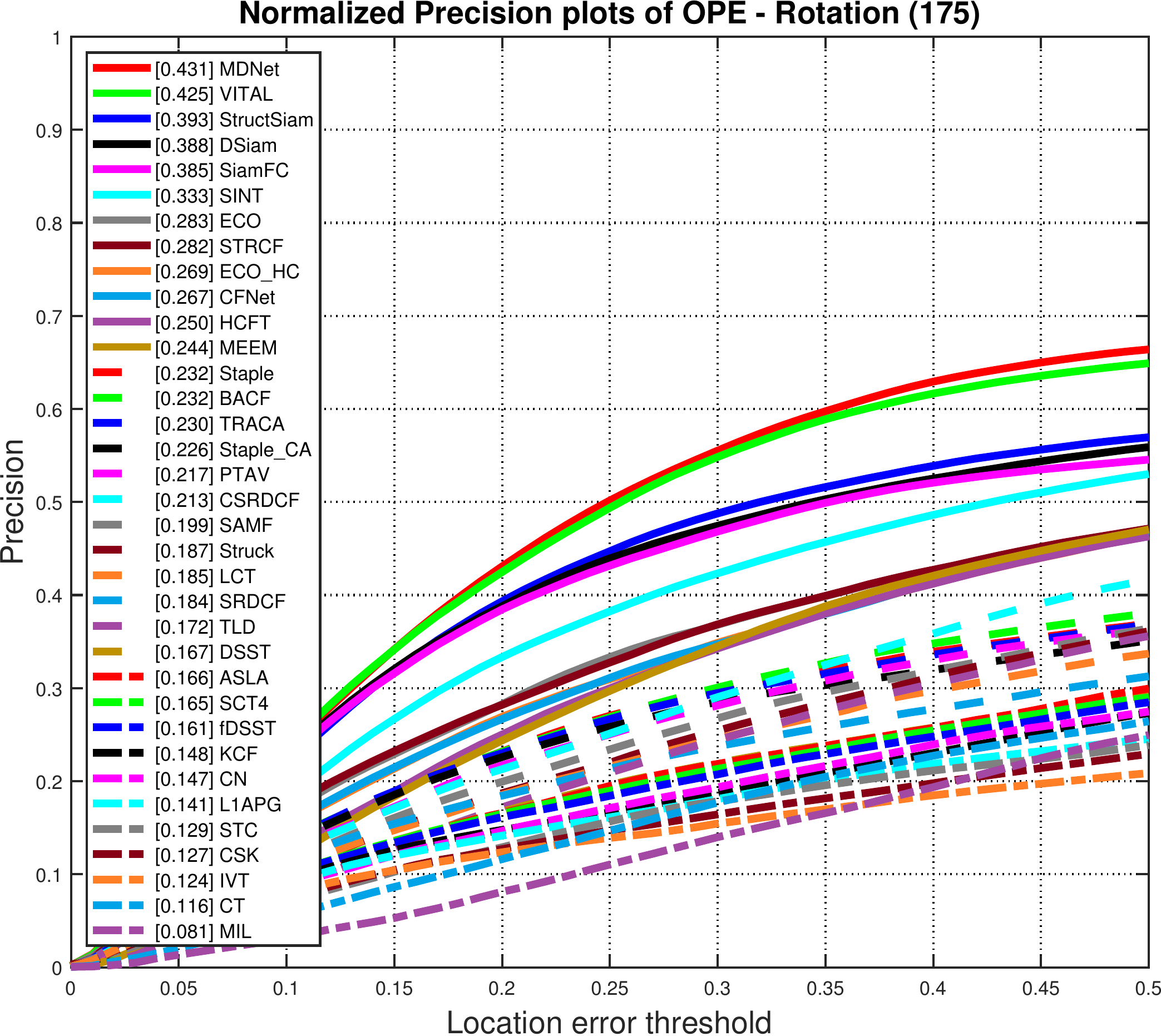}\\
	\includegraphics[width=4.55cm]{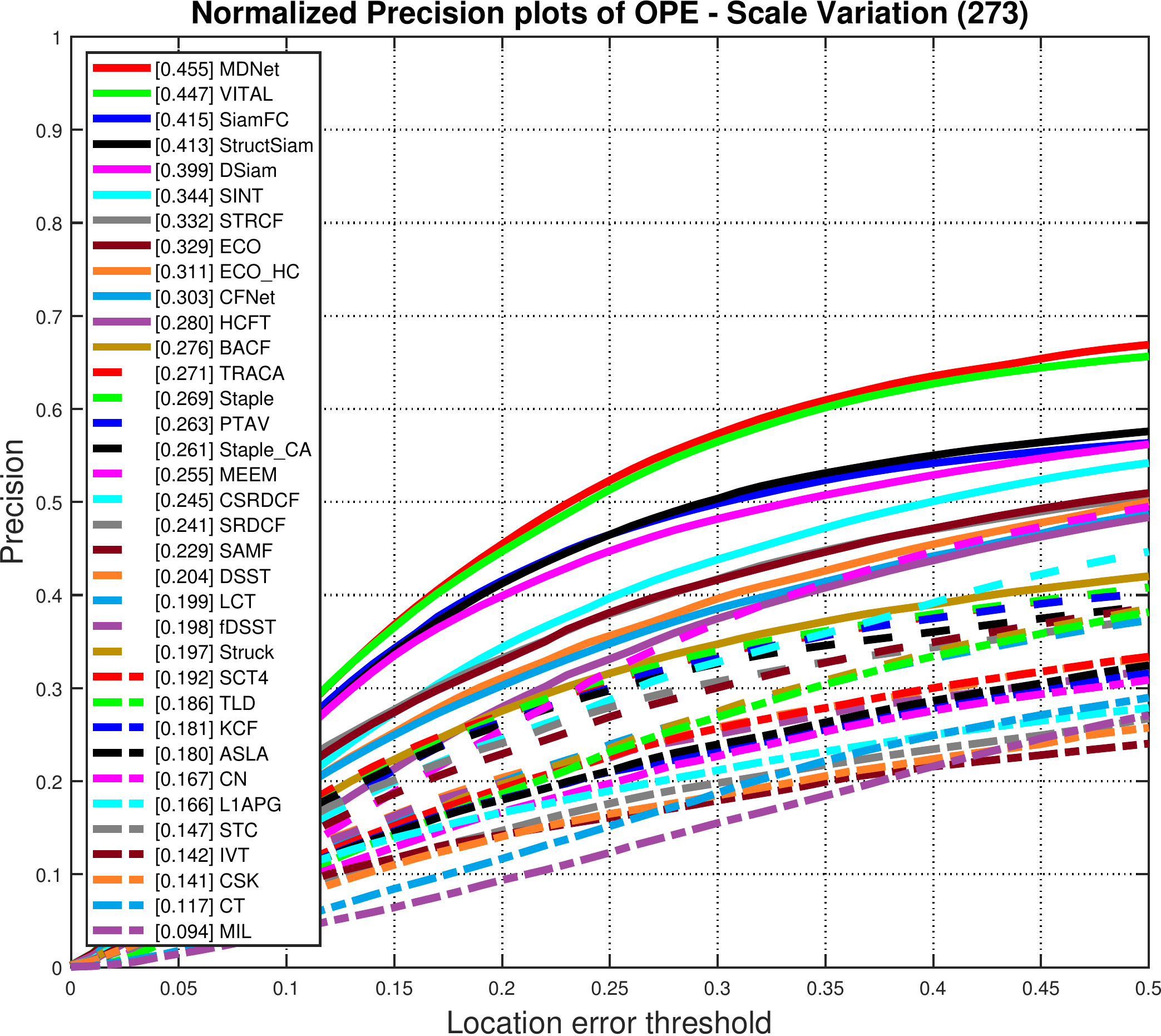}
	\includegraphics[width=4.55cm]{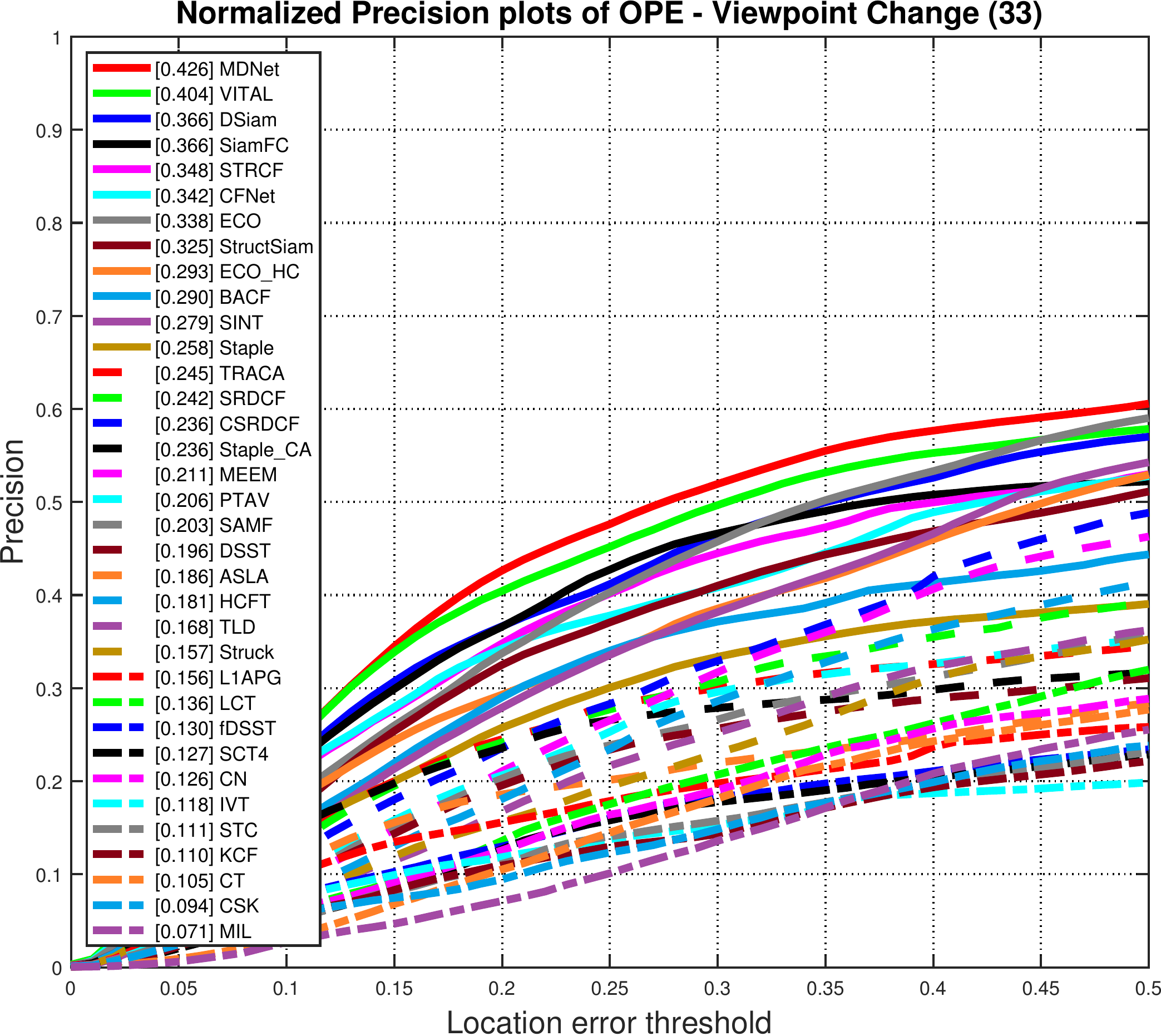}\\
	\caption{Performance of trackers on each attribute using precision under protocol \uppercase\expandafter{\romannumeral2}. Best viewed in color.}
	\label{fig:protocol_2_all_att_res_norm_precision}
\end{figure*}

\newpage
Fig.~\ref{fig:protocol_2_all_att_res_success} shows the performance of trackers on each attribute using success under protocol \uppercase\expandafter{\romannumeral2}.
\begin{figure*}[!hbpt]
	\centering
	\includegraphics[width=4.55cm]{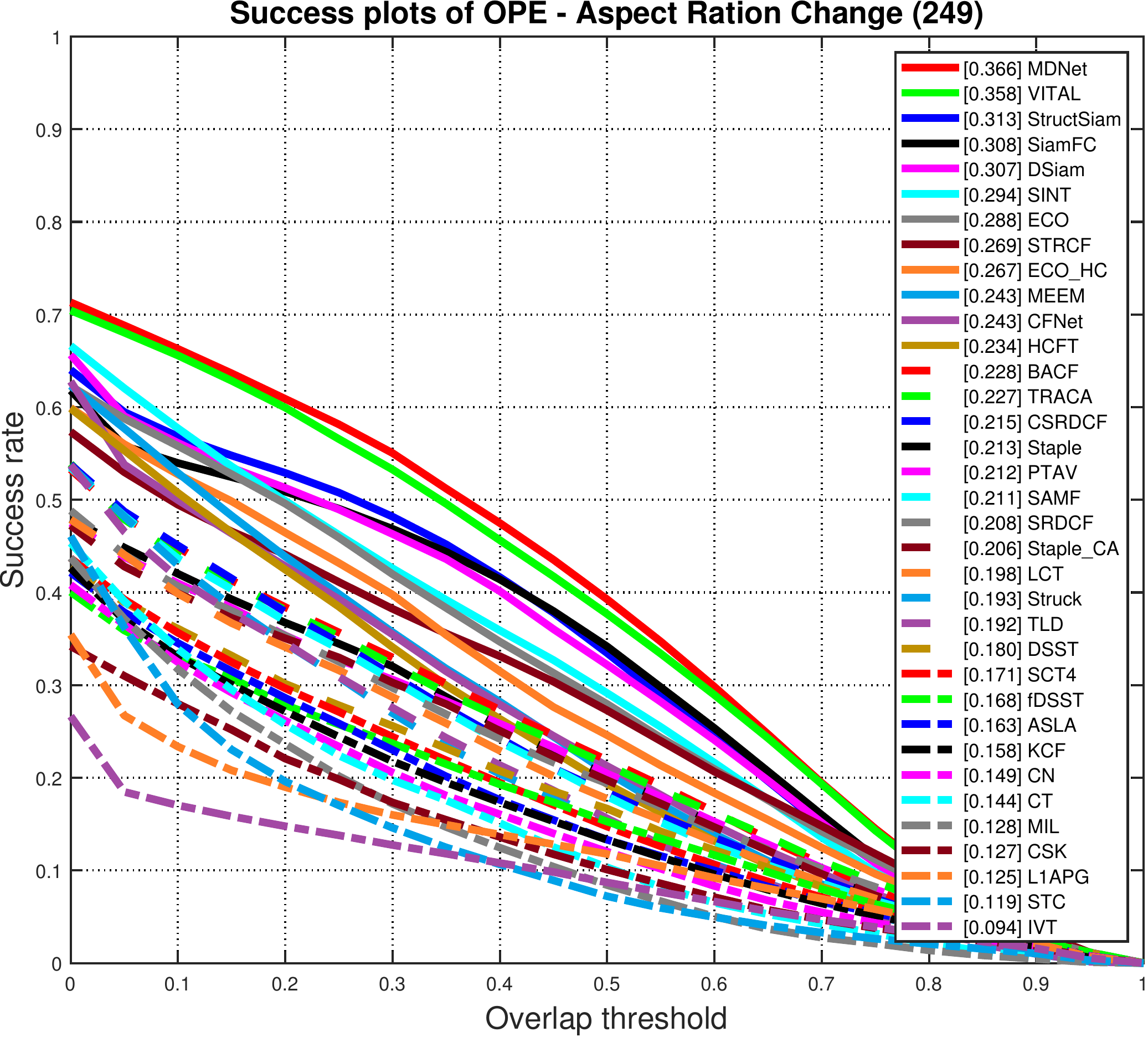}
	\includegraphics[width=4.55cm]{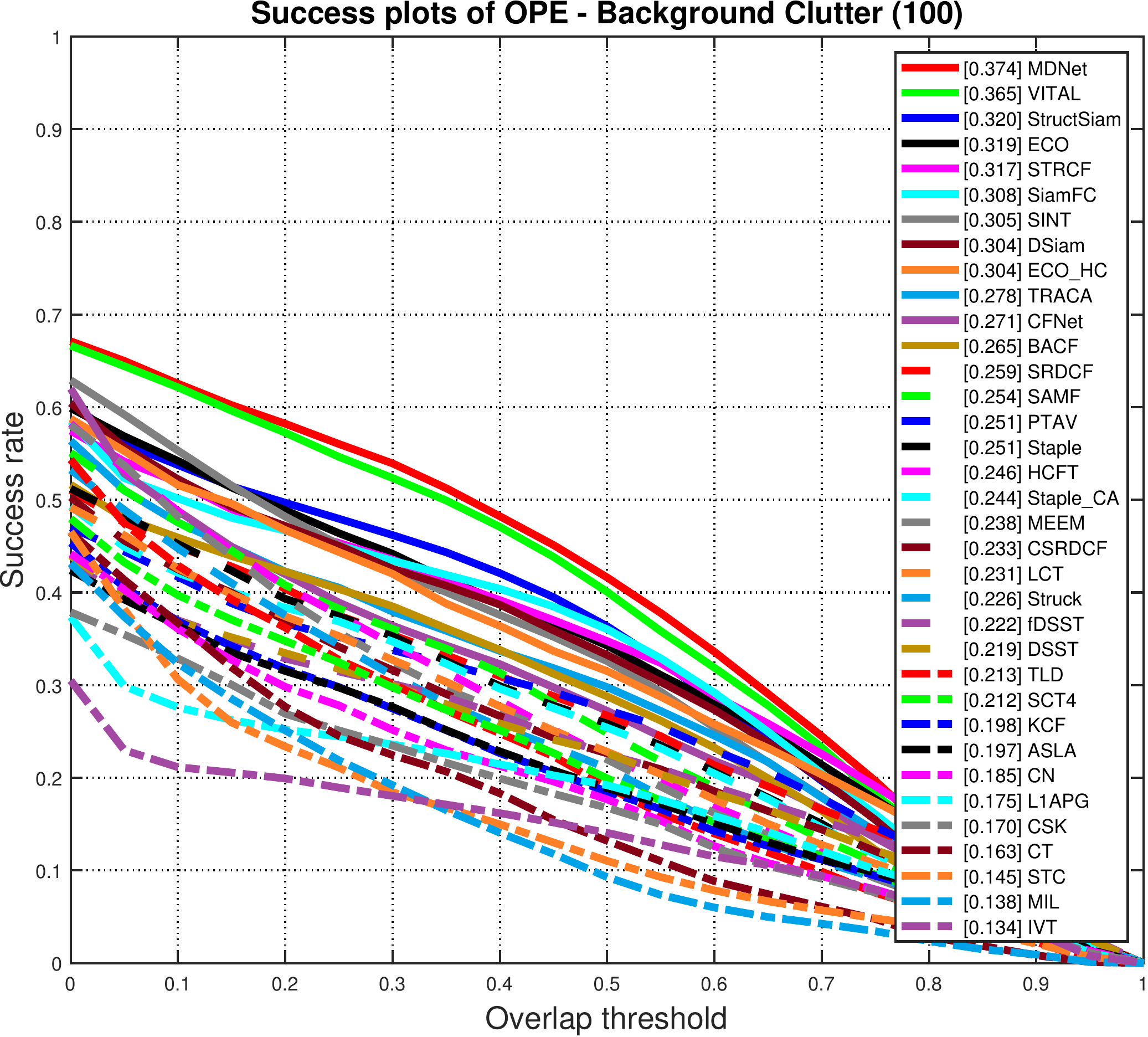}
	\includegraphics[width=4.55cm]{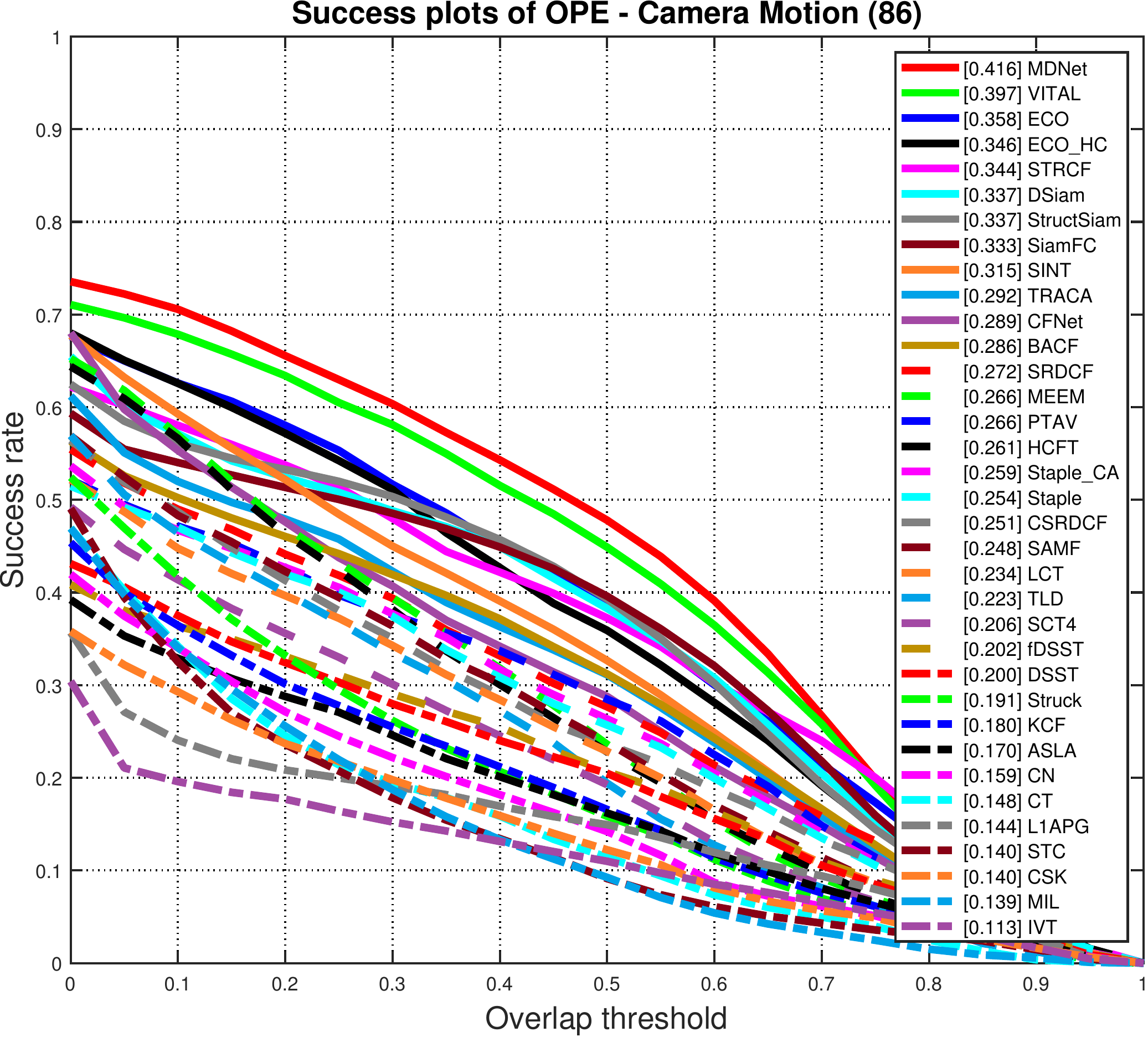}\\
	\includegraphics[width=4.55cm]{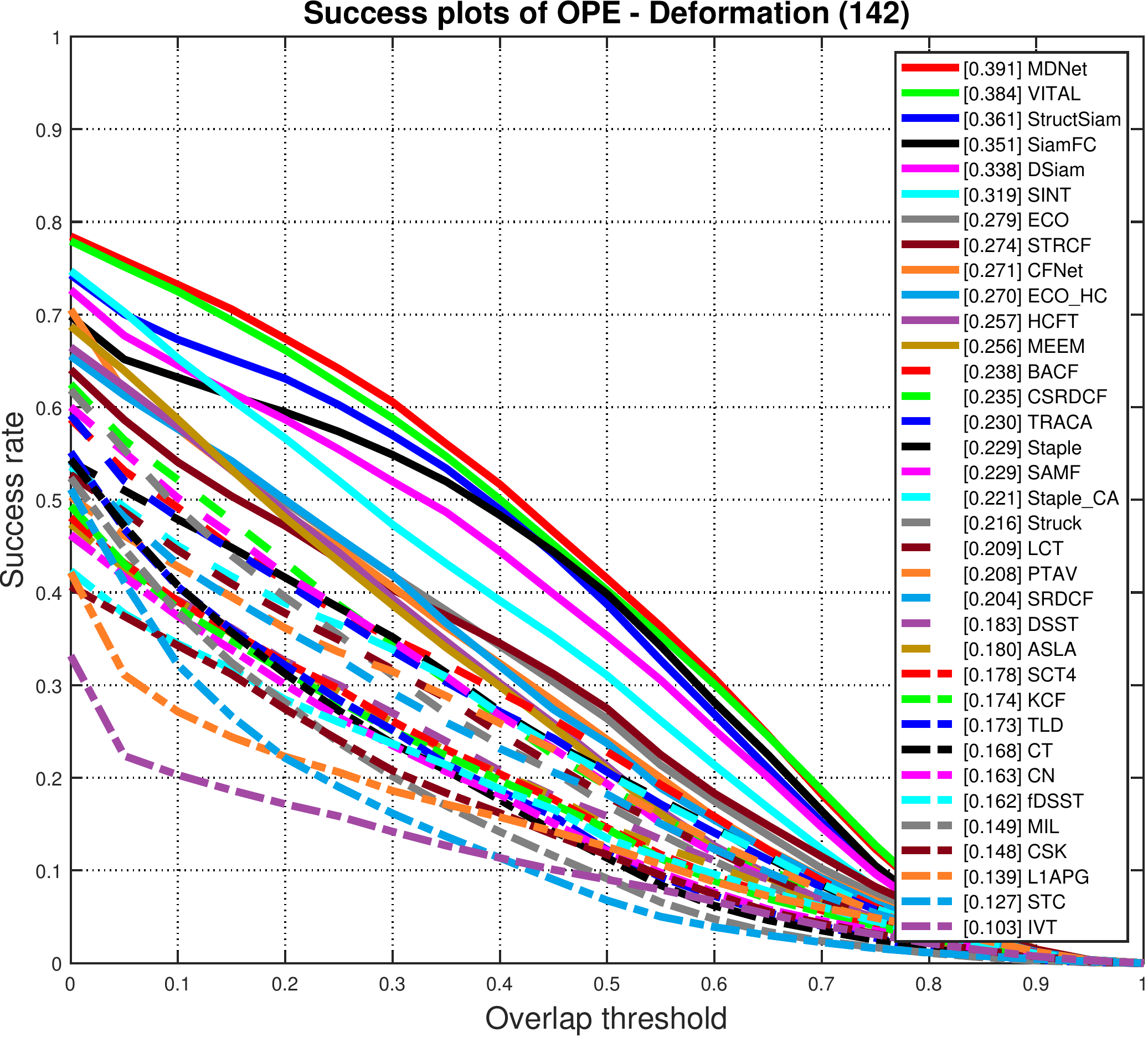}
	\includegraphics[width=4.55cm]{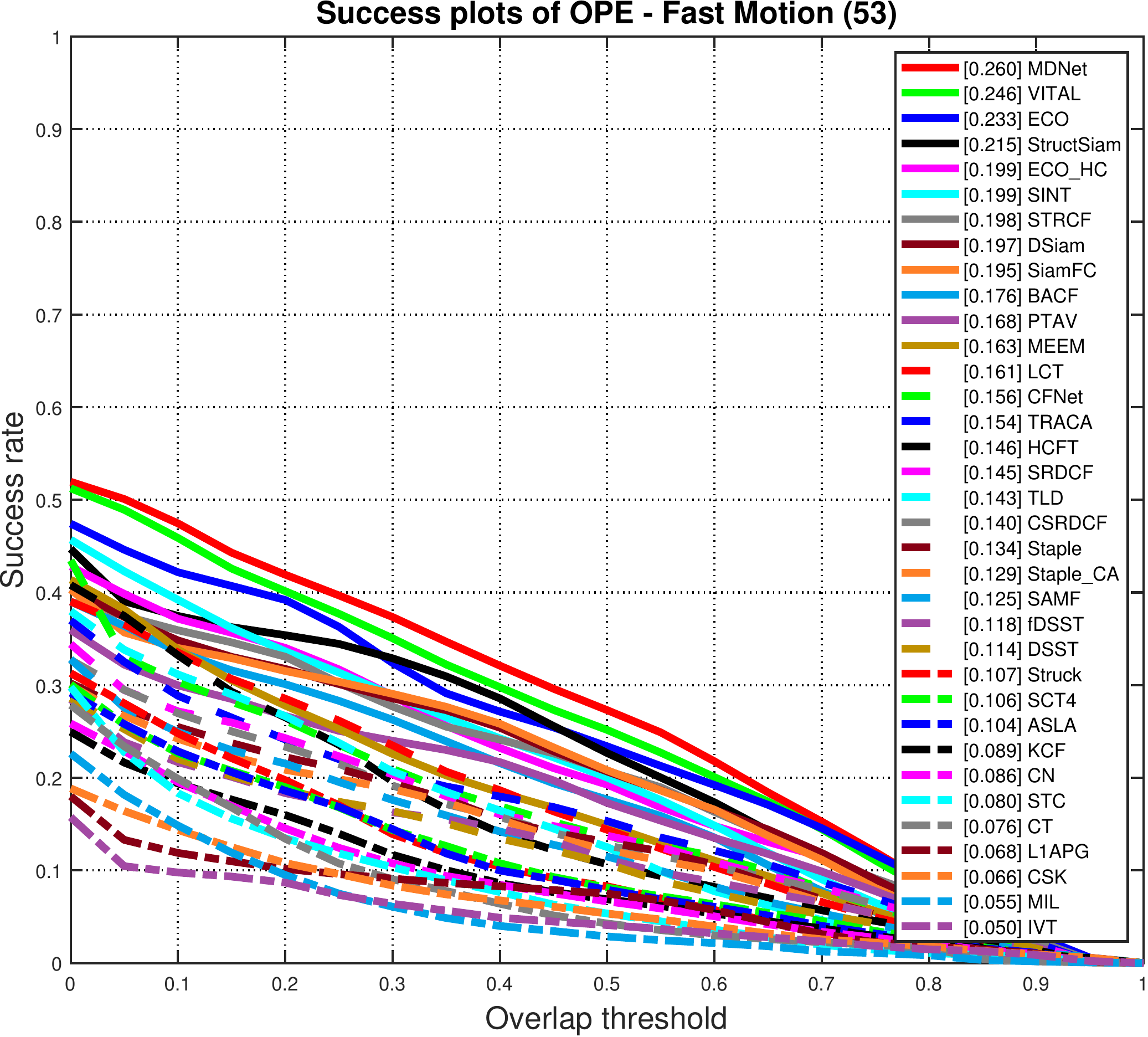}
	\includegraphics[width=4.55cm]{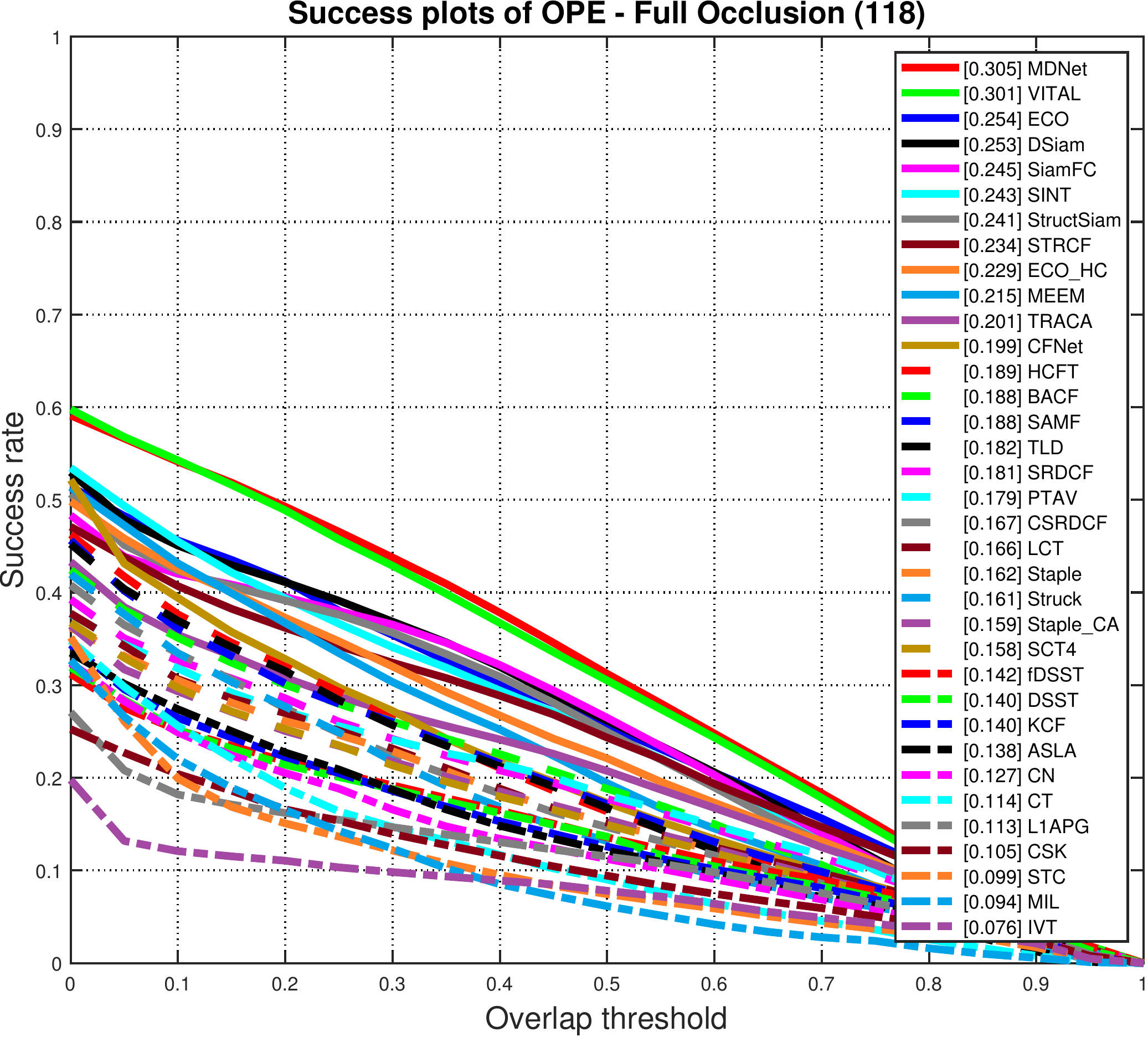}\\
	\includegraphics[width=4.55cm]{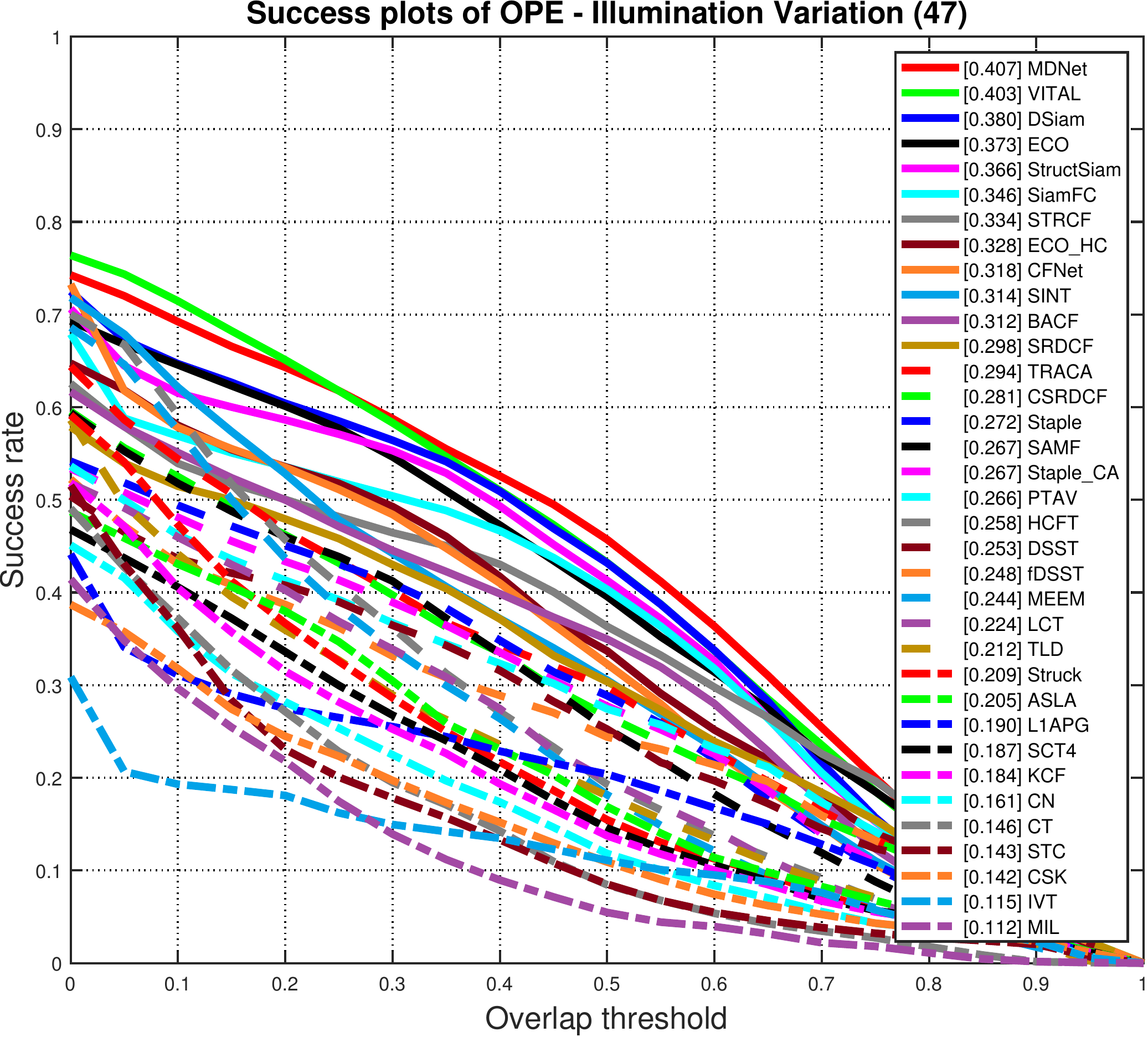}
	\includegraphics[width=4.55cm]{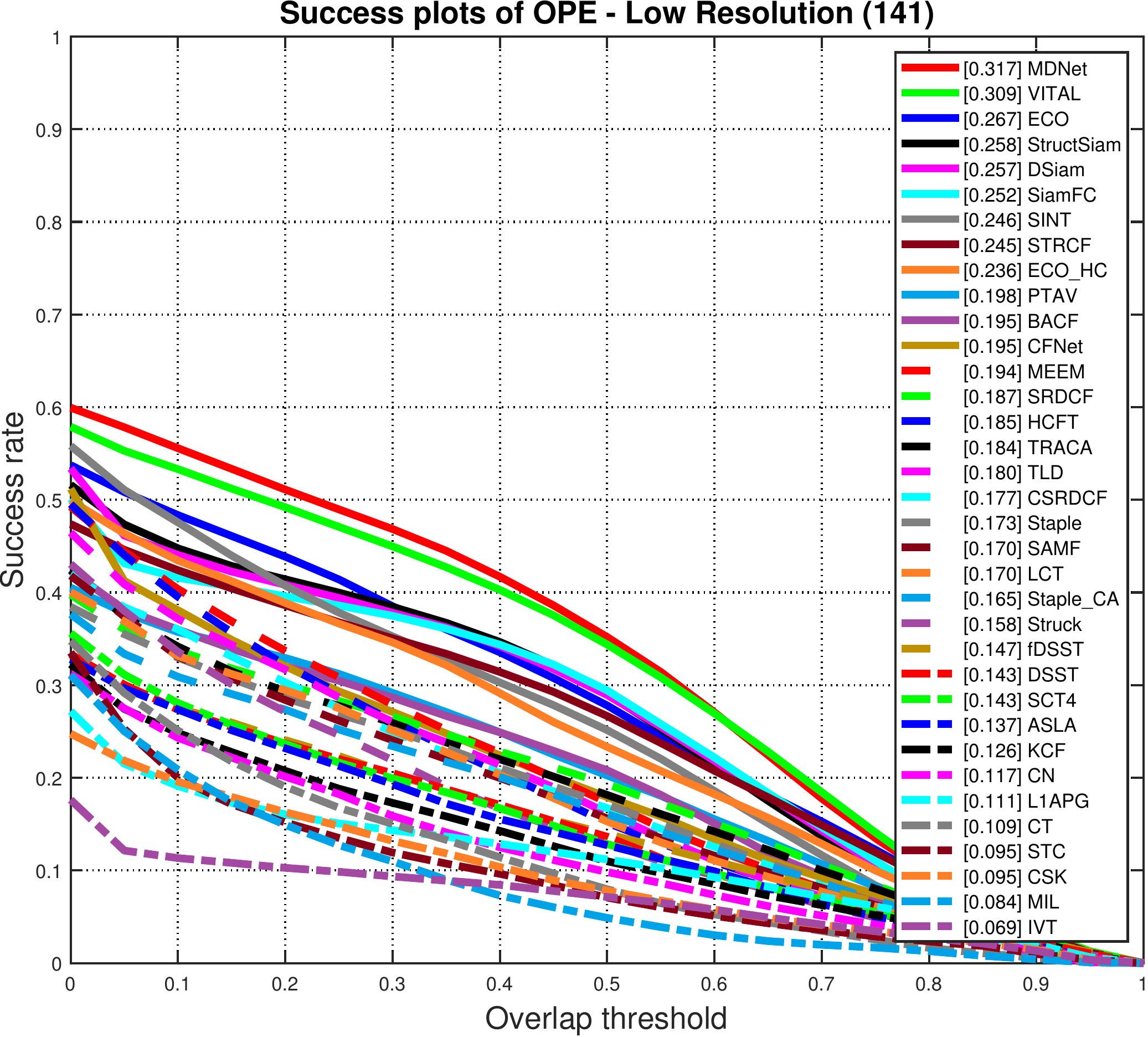}
	\includegraphics[width=4.55cm]{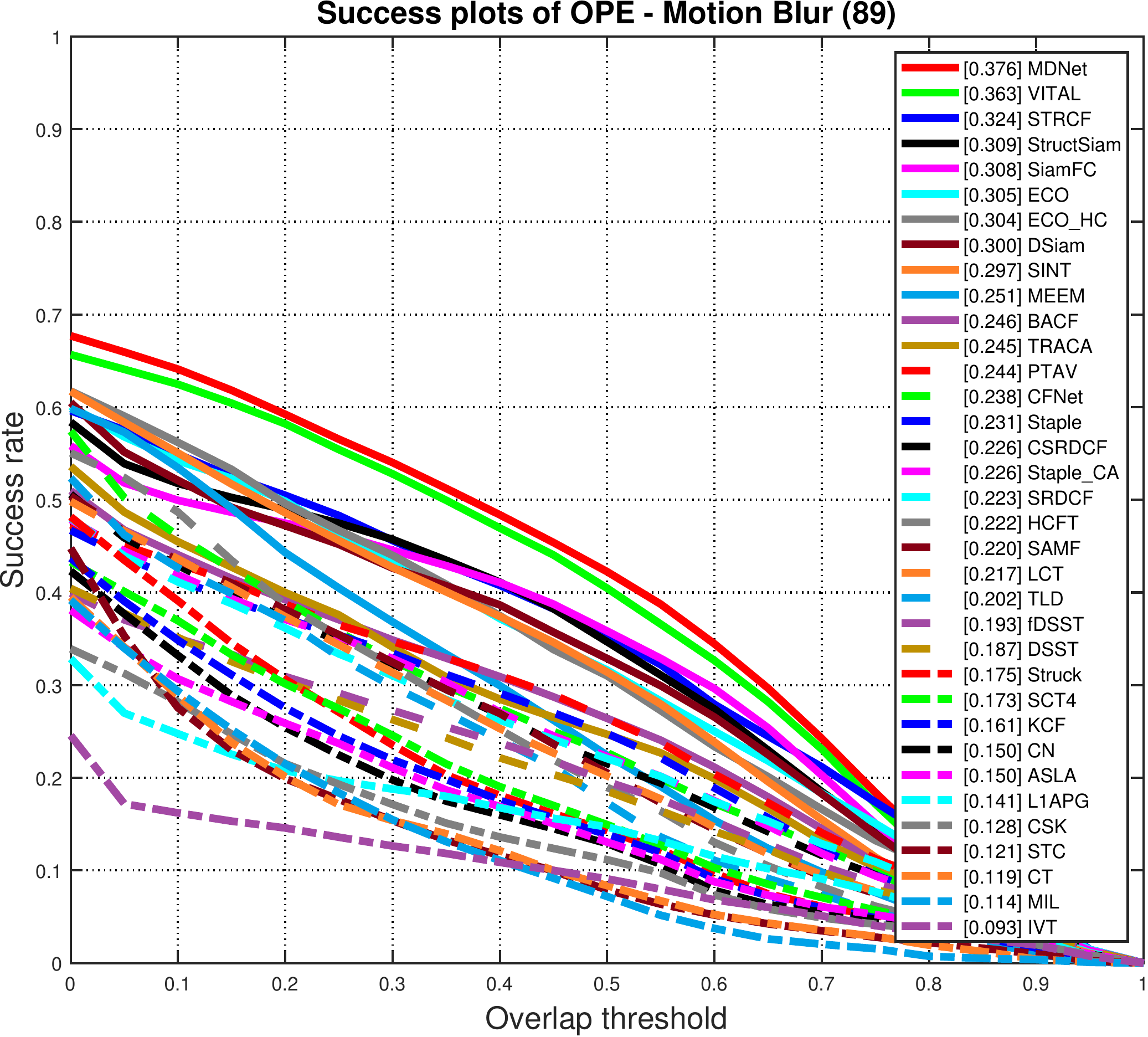}\\
	\includegraphics[width=4.55cm]{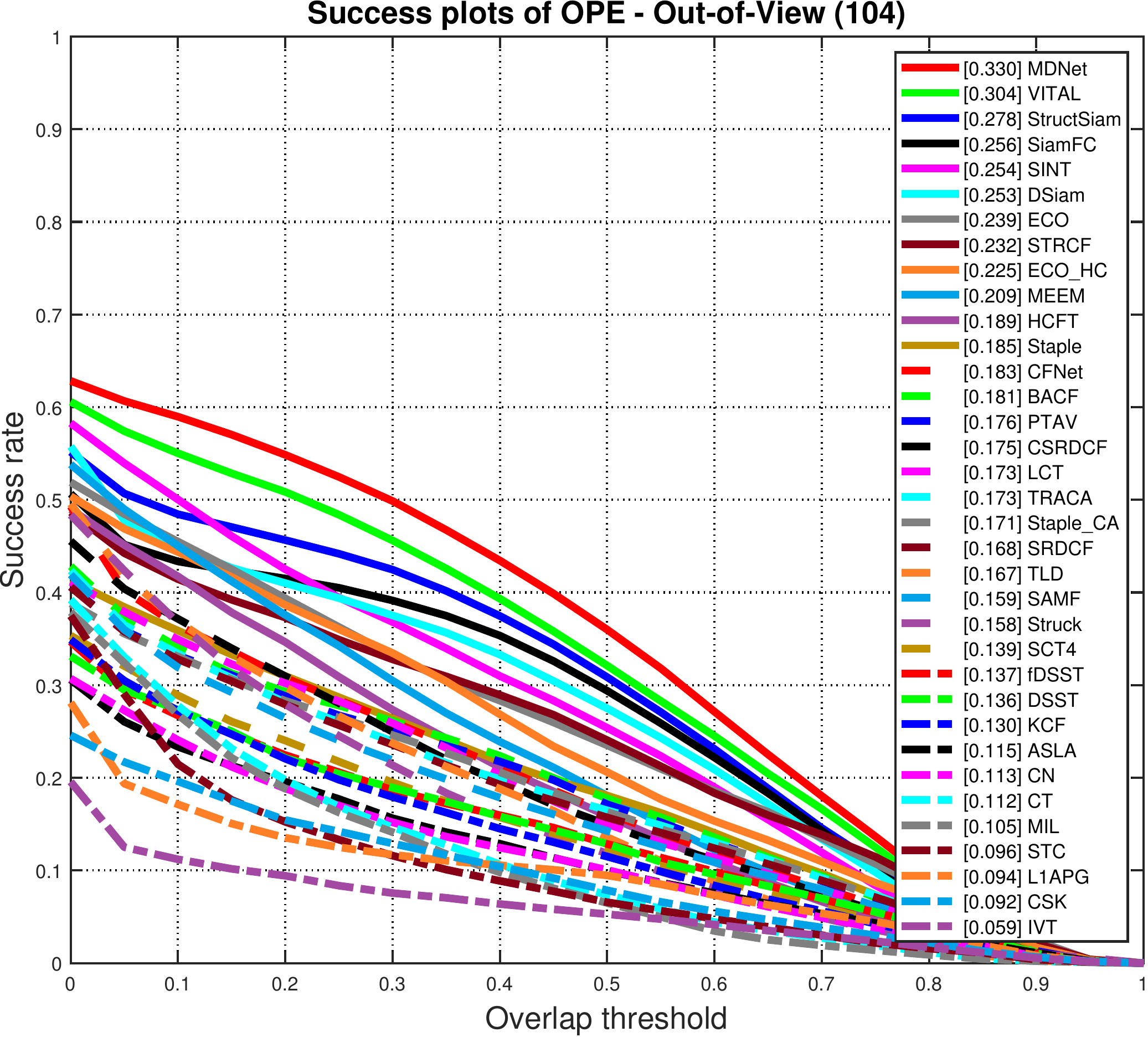}
	\includegraphics[width=4.55cm]{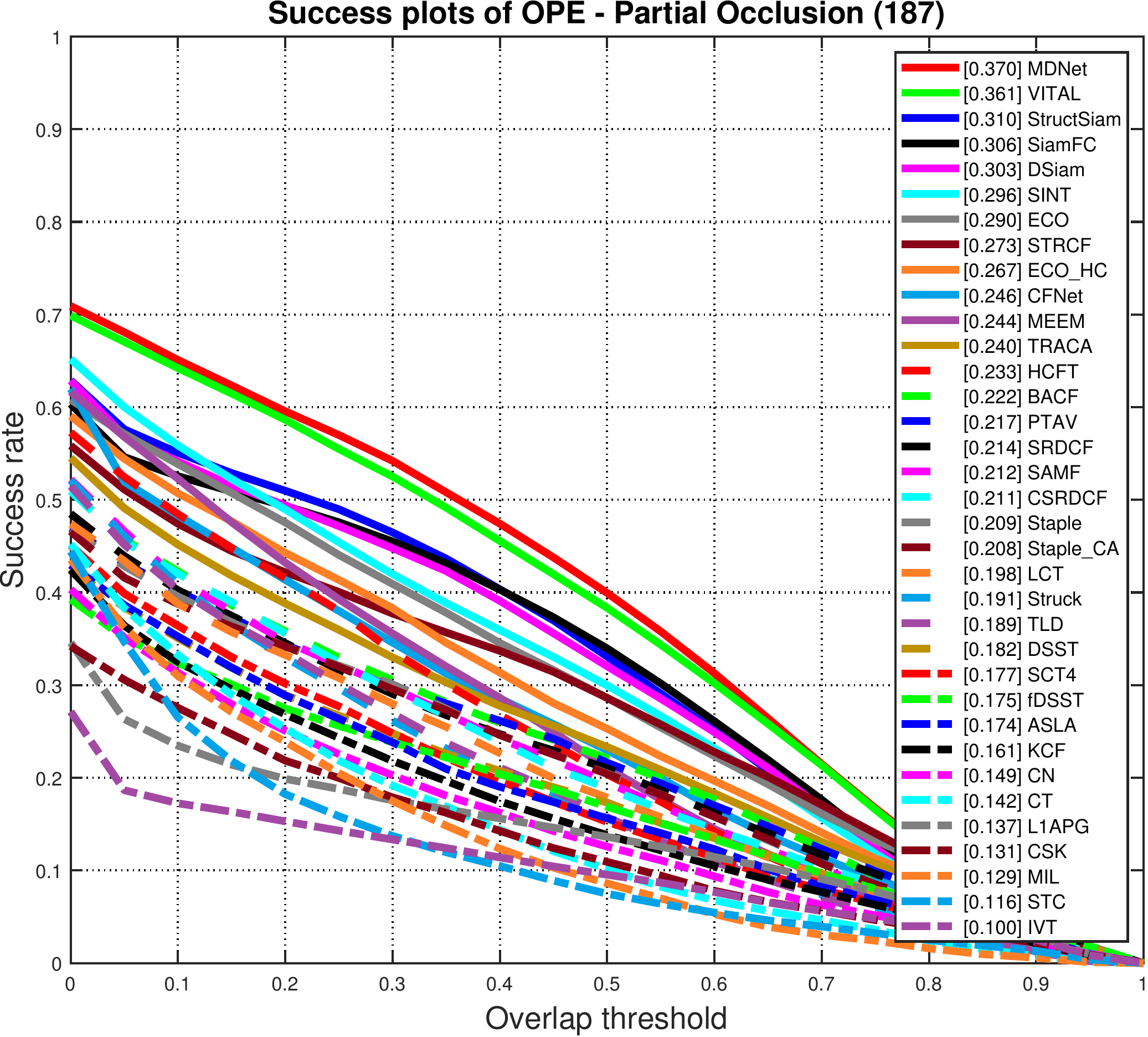}
	\includegraphics[width=4.55cm]{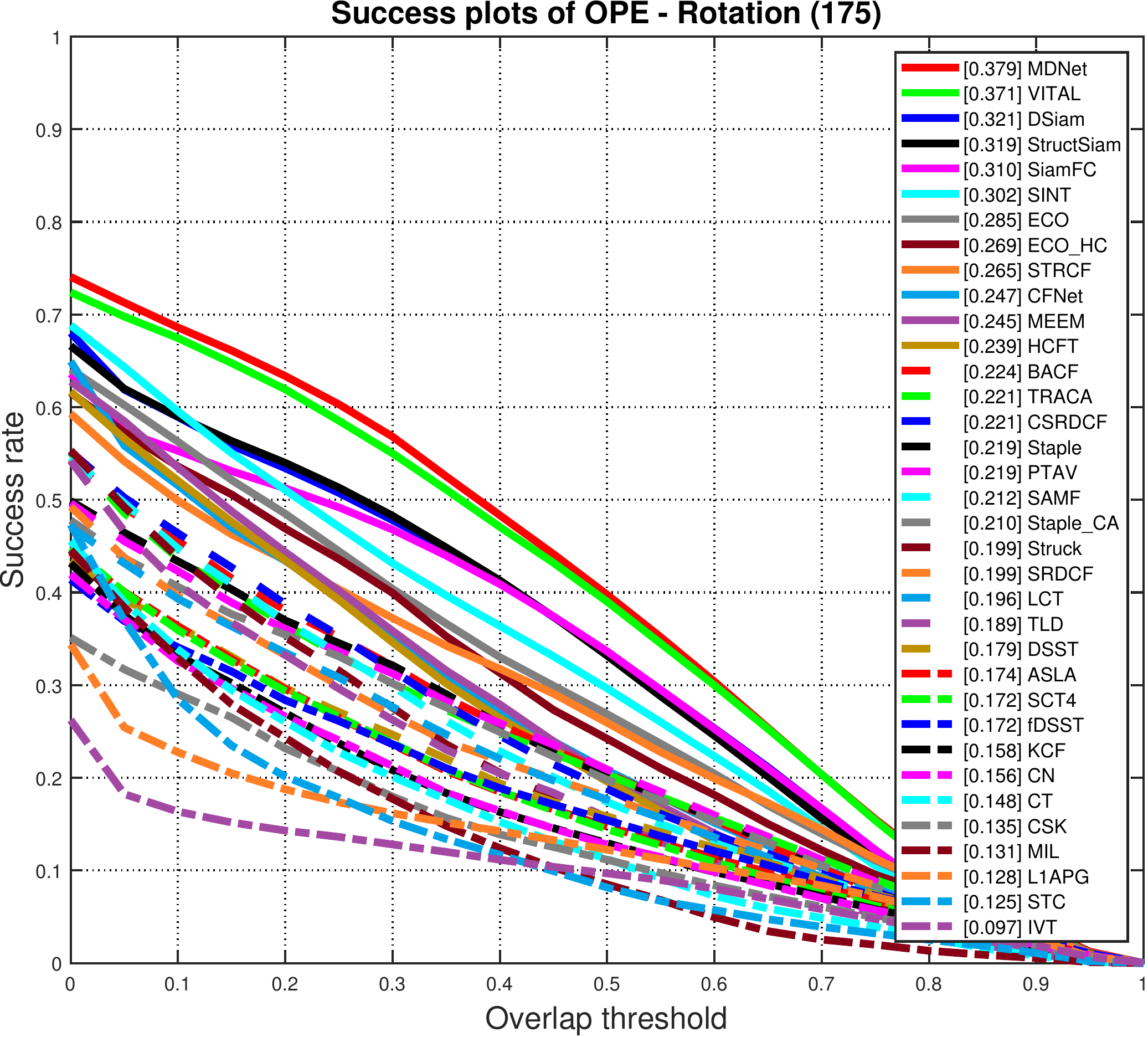}\\
	\includegraphics[width=4.55cm]{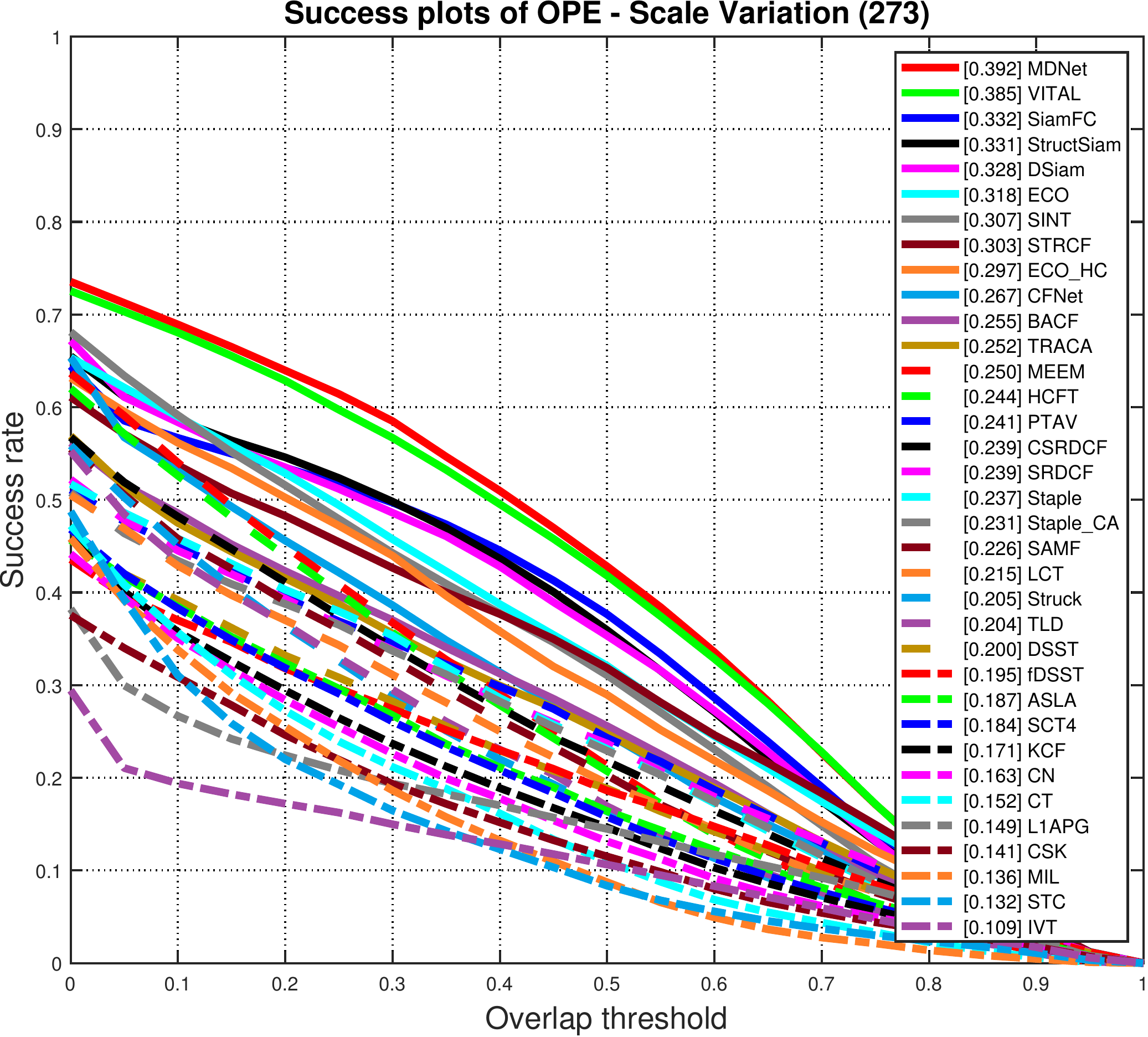}
	\includegraphics[width=4.55cm]{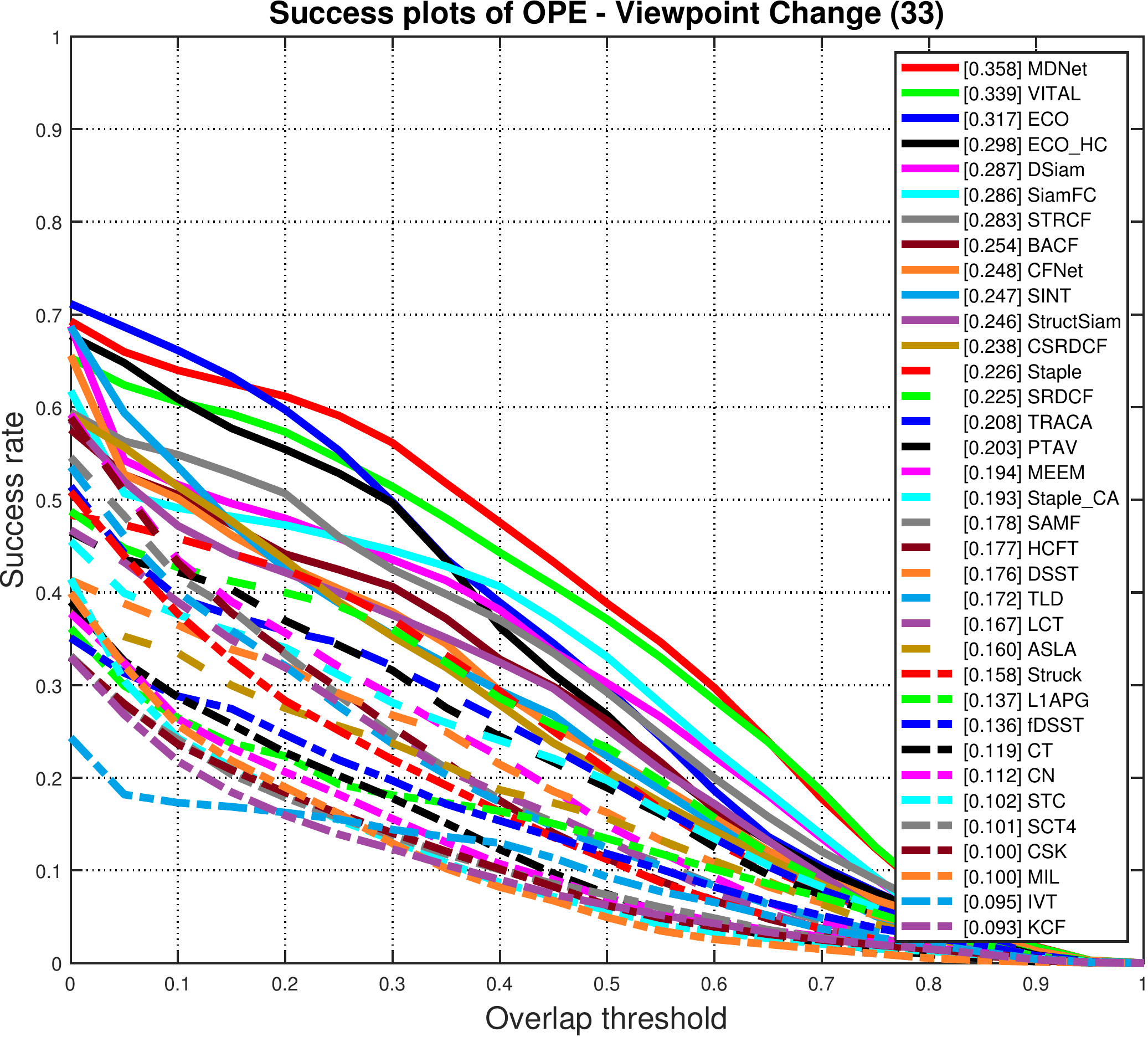}\\
	\caption{Performance of trackers on each attribute using success under protocol \uppercase\expandafter{\romannumeral2}. Best viewed in color.}
	\label{fig:protocol_2_all_att_res_success}
\end{figure*}
	
\end{onecolumn}

\end{document}